\documentclass[preprint,12pt]{elsarticle}

\usepackage{amssymb}
\usepackage{amsmath}

\usepackage[switch]{lineno}

\usepackage{subcaption}
\usepackage[export]{adjustbox}
\usepackage{url}
\usepackage{graphicx}
\usepackage{multirow}
\usepackage[table,x11names,dvipsnames]{xcolor}
\newtheorem{hypothesis}{Hypothesis}

\newtheorem{diff}{Diff}
\usepackage{booktabs} 
\usepackage{array,multirow}
\usepackage{makecell}
\usepackage{threeparttable} 
\usepackage{tabularx}
\usepackage{lscape}
\usepackage{longtable}
\usepackage{arydshln}
\usepackage{tabularray}
\usepackage{xltabular}
\usepackage{array}
\usepackage{url}

\usepackage{booktabs, makecell, rotating}
\usepackage{siunitx}

\usepackage{tikz}
\usetikzlibrary{arrows.meta, positioning, shapes, calc, fit}
\tikzset{
  >={Latex[length=3mm]},
  stage/.style={rectangle, rounded corners, minimum width=27mm, minimum height=10mm, fill=blue!6, draw=blue!50!black, very thick, font=\footnotesize, align=center, inner sep=3pt},
  decision/.style={diamond, aspect=2, fill=orange!15, draw=orange!60!black, very thick, font=\footnotesize, align=center, inner sep=1.5pt},
  small/.style={font=\scriptsize},
  link/.style={->, thick, draw=gray!70}
}

\usepackage{xltabular} 

\newcommand\highlightReference[1]{%
  \expandafter\newcommand\csname highlightReference-#1\endcsname{}%
}
\usepackage{xpatch}
\makeatletter
\xapptocmd\@lbibitem{\hlbibitem{#2}}{}{}
\makeatother
\def\hlbibitem#1 #2\par{%
  \expandafter\ifx\csname highlightReference-#1\endcsname\relax
    #2\par
  \else
    \highlight{#2}\par
  \fi
}
\usepackage{color}
\newcommand\highlight[1]{\textcolor{blue}{#1}}

\usepackage{soul}
\usepackage[normalem]{ulem}
\makeatletter
\soulregister\cite7
\soulregister\citet7
\soulregister\citep7
\soulregister\citealt7
\soulregister\citealp7
\soulregister\citeauthor7
\soulregister\citeyear7
\soulregister\citeyearpar7
\soulregister\ref7
\soulregister\eqref7
\soulregister\url7 
\makeatother
\newcommand{\eqnref}[1]{Eq.~(\ref{#1})}
\newcommand{\lamreturn}{$\lambda$-return}

\usepackage{pifont}
\newcommand{\cmark}{\ding{51}}%
\newcommand{\xmark}{\ding{55}}%
\usepackage{xcolor}
\usepackage{bbding}
\usepackage{hyperref} 

\journal{Neural Networks}

\begin{document}
\newif\ifreview 

\reviewfalse 
\ifreview
    \definecolor{origColor}{RGB}{0,0,0}
    \definecolor{addColor}{RGB}{0,0,180}
    \definecolor{removedColor}{RGB}{255,182,193}
    \definecolor{placeholderColor}{RGB}{255,120,0}
    \newcommand{\orig}[1]{\textcolor{origColor}{#1}}
    \newcommand{\add}[1]{\textcolor{addColor}{#1}}
    \providecommand{\removed}{}
    \renewcommand{\removed}[1]{\textcolor{removedColor}{#1}}
    \newcommand{\removedstrike}[1]{\textcolor{removedColor}{\st{#1}}}
    \newcommand{\removedshort}[1]{\textcolor{removedColor}{\begingroup\setlength{\fboxsep}{0pt}\ifcsname st\endcsname\st{#1}\else\sout{#1}\fi\endgroup}}
    \color{origColor}
    \linenumbers
\else
    \definecolor{origColor}{RGB}{0,0,0}
    \definecolor{addColor}{RGB}{0,0,0}
    \definecolor{removedColor}{RGB}{0,0,0}
    \definecolor{placeholderColor}{RGB}{0,0,0}
    \newcommand{\orig}[1]{#1}
    \newcommand{\add}[1]{#1}
    \newcommand{\removed}[1]{}
    \newcommand{\removedstrike}[1]{}
    \newcommand{\removedshort}[1]{}
\fi
\ifreview
\else
    \renewcommand\highlight[1]{#1}
\fi
\newcommand{\metric}[1]{\textcolor{placeholderColor}{\textbf{#1}}}
\newcommand{\secref}[1]{Section~\ref{#1}}
\newcommand{\figref}[1]{Fig.~\ref{#1}}
\newcommand{\tabref}[1]{Table~\ref{#1}}
\newcommand{\eqrefp}[1]{Eq.~(\ref{#1})}
\makeatletter
\@ifundefined{lineref}{\newcommand{\lineref}[1]{line~\ref{#1}}}{} 
\makeatother
\newcommand{\pagelineref}[1]{p.~\pageref{#1}, line~\ref{#1}}
\newcommand{\pagelinerange}[2]{p.~\pageref{#1}, lines~\ref{#1}--\ref{#2}}

\begin{frontmatter}



\title{Multi-Step First: A Lightweight Deep Reinforcement Learning Strategy for Robust Continuous Control with Partial Observability\tnoteref{t1}}

\tnotetext[t1]{Published in \textit{Neural Networks}, Vol.~199, 2026, 
Article 108521. \href{https://doi.org/10.1016/j.neunet.2025.108521}
{https://doi.org/10.1016/j.neunet.2025.108521}}

\author[label1,label2,label4]{Lingheng Meng}
\affiliation[label1]{organization={Electrical and Computer Engineering, University of Waterloo},
            addressline={200 University Avenue West},
            city={Waterloo},
            postcode={N2L 3G1},
            state={ON},
            country={Canada}}

\affiliation[label2]{organization={Electrical and Computer Systems Engineering, Monash University},
            addressline={18 Alliance Lane},
            city={Clayton},
            postcode={3800},
            state={VIC},
            country={Australia}}

\author[label1,label3]{Rob Gorbet} 

\affiliation[label3]{organization={Knowledge Integration, University of Waterloo},
            addressline={200 University Avenue West},
            city={Waterloo},
            postcode={N2L 3G1},
            state={ON},
            country={Canada}}

\author[label2]{Michael Burke}
\author[label2,label4]{Dana Kuli\'c}

\affiliation[label4]{organization={Data61, CSIRO},
            addressline={Research Way},
            city={Calyton},
            postcode={3168},
            state={VIC},
            country={Australia}}

\begin{abstract}
Deep Reinforcement Learning (DRL) has made considerable advances in simulated and physical robot control tasks, especially when problems admit a fully observed Markov Decision Process (MDP) formulation. When observations only partially capture the underlying state, the problem becomes a Partially Observable MDP (POMDP), and performance rankings between algorithms can change. We empirically compare Proximal Policy Optimization (PPO), Twin Delayed Deep Deterministic Policy Gradient (TD3), and Soft Actor-Critic (SAC) on representative POMDP variants of continuous-control benchmarks. Contrary to widely reported MDP results where TD3 and SAC typically outperform PPO, we observe an inversion: PPO attains higher robustness under partial observability. We attribute this to the stabilizing effect of multi-step bootstrapping. Furthermore, incorporating multi-step targets into TD3 (MTD3) and SAC (MSAC) improves their robustness. These findings provide practical guidance for selecting and adapting DRL algorithms in partially observable settings without requiring new theoretical machinery.

\end{abstract}


\end{frontmatter}



\section{Introduction}
\label{Introduction}

Deep Reinforcement Learning (DRL) has propelled progress in both discrete and continuous control domains \cite{mnih2013playing,mnih2015human,lillicrap2015continuous,duan2016benchmarking,mahmood2018benchmarking}, yet deploying these gains on real robotic systems remains difficult \cite{henderson2018deep,dulac2021challenges}. Among practical obstacles, \textit{partial observability} (missing or noisy state information) quietly invalidates the common Markov Decision Process (MDP) \cite{sutton2018reinforcement} assumption, turning problems into Partially Observable MDPs (POMDPs) \cite{cassandra1998survey,shani2013survey} and reshuffling expected algorithm performance.

Standard benchmarking pipelines implicitly assume fully observed dynamics; practitioners often begin with TD3 \cite{fujimoto2018addressing} or SAC \cite{haarnoja2018soft} given published superiority on canonical MDP suites \cite{duan2016benchmarking}. However, realistic sensing limitations, latency, dropout, or intentional observation pruning (e.g., removing velocity channels) regularly induce hidden state. In novel interactive systems \cite{meng2023learning}, confidently classifying task observability upfront is rarely feasible.

Recurrent augmentations proposed for POMDPs \cite{hausknecht2015deep,lample2017playing,igl2018deep} do not guarantee retained performance on the original MDP versions; recurrence can degrade learning via instability or redundant memory \cite{kapturowski2018recurrent,ni2021recurrent,meng2021memory}. Two gaps follow: (1) lack of a lightweight empirical \emph{signal} to suspect partial observability; (2) limited availability of algorithms whose rank ordering is stable across MDP/POMDP variants.

We surface and generalize an \textbf{Unexpected Result}: in multiple continuous-control benchmarks converted to POMDPs (noise, flickering, sensor removal, velocity stripping) PPO \cite{schulman2017proximal} \emph{outperforms} TD3 \cite{fujimoto2018addressing} and SAC \cite{haarnoja2018soft}—despite underperforming them on the original MDP versions \cite{duan2016benchmarking}. This inversion recurs across Ant, HalfCheetah, Hopper, and Walker2D variants.
These four tasks are standard MuJoCo continuous-control benchmarks provided by Gymnasium \cite{todorov2012mujoco,towers_gymnasium_2023}. 
Their diversity (morphology, contact patterns, degrees of freedom) makes them a representative substrate for robustness comparisons under induced partial observability.

We hypothesize multi-step bootstrapping ($\lambda$-returns and Monte Carlo baselines) \cite{schulman2015high,de2018multi,van2016effective} provides temporal smoothing that implicitly recovers missing latent state features, while 1-step targets used by TD3/SAC are more brittle under partial observability. Related empirical indications that longer returns mitigate estimation issues appear in \cite{hessel2018rainbow,meng2021effect}. Exploration style (conservative PPO clips \cite{schulman2017proximal} vs. entropy-driven SAC \cite{haarnoja2018soft} or Gaussian TD3 noise \cite{fujimoto2018addressing}) is examined as a secondary factor. Multi-step extensions MTD3 and MSAC substantially restore robustness \cite{tang2020self,bai2021empirical}, supporting the temporal propagation view.
Concurrently, advances in efficient long-horizon sequence modeling---retention mechanisms (RetNet) \cite{shi2023retnet}, selective state-space (Mamba) \cite{gu2023mamba}, and optimized attention kernels (FlashAttention-2) \cite{dao2023flashattention2}---lower the computational barrier to processing extended context. Our findings indicate that even without adopting these newer architectures, simple multi-step return horizons (n=2--5) yield a substantial fraction of the robustness gains such models might target, suggesting a pragmatic progression: tune multi-step horizons first, then consider specialized long-context architectures if inversion persists.
Complementary perspectives from control of delayed and fault-prone competitive neural networks \cite{cao2024exponential} and synchronization of memristive inertial competitive neural systems under time-varying delay \cite{subhashri2025robust} emphasize rigorous state estimation and stability mechanisms when dynamics, delays, and actuator failures impair direct observation. These works motivate seeking lightweight temporal aggregation (multi-step returns) as a first response before introducing heavier model-based or sliding-mode style controllers in DRL settings with aliasing.

\paragraph{Contributions} This paper provides:
\begin{enumerate}
    \item \textbf{Benchmarked inversion breadth:} Reproducible PPO \emph{$>$ }TD3/SAC inversions across four morphologically diverse MuJoCo control tasks (Ant, HalfCheetah, Hopper, Walker2D) and four observability perturbation types (velocity removal, sensor flickering, additive random noise, random sensor missing).
    \item \textbf{Lightweight theoretical lens:} An observability-gap based framework (gap definition, variance bound, bias--variance horizon trade-off, inversion threshold) explaining why multi-step bootstrapping plus conservative updates attenuate performance degradation under partial observability.
    \item \textbf{Temporal bootstrapping mechanism:} Empirical and conceptual evidence that moderate multi-step return horizons (n=2--5) recover latent state information more effectively than 1-step targets without requiring recurrent networks.
    \item \textbf{Adaptation pathway (MTD3/MSAC):} Simple n-step extensions of TD3 and SAC that substantially restore robustness, isolating horizon depth as the principal controllable factor.
    \item \textbf{Diagnostic heuristic:} Performance inversion (PPO surpassing TD3 and SAC where the opposite holds on fully observed benchmarks) proposed and justified as a practical signal of hidden state/observation insufficiency.
    \item \textbf{Efficiency characterization:} Symbolic computational and memory cost analysis contrasting added multi-step overhead with recurrence or long-context sequence models, clarifying when horizon extension is a cost-effective first intervention.
    \item \textbf{Cross-disciplinary linkage \& positioning:} Synthesis connecting delay/fault-tolerant neural control results and emerging efficient long-sequence architectures (RetNet, Mamba, FlashAttention-2) to a staged DRL robustness strategy: (i) detect inversion, (ii) extend horizon, (iii) tune exploration, (iv) escalate to memory or specialized sequence models if needed.
    \item \textbf{Actionable guidelines:} Consolidated recommendations prioritizing horizon tuning and multi-step variants before introducing recurrence or architectural complexity when observability is uncertain.
\end{enumerate}

\textbf{Paper structure and roadmap.} The remainder of this paper is organized as follows: Section \ref{sec:Exemplar_Robot_Control_Problem} presents a motivating Walker2D example demonstrating how observation noise causes PPO to outperform TD3 and SAC, contrary to established MDP rankings. Section \ref{sec:Generalization_of_the_Unexpected_Result_on_Other_Tasks} formulates our central hypothesis and describes experimental methodology for testing generalization across multiple benchmarks and observation perturbation types, with actual validation results presented in Section \ref{sec:experimental_results}. Section \ref{sec:Analysing_and_Improving_Robustness_to_Partial_Observability} analyzes key algorithmic differences explaining PPO's superior robustness, focusing on multi-step bootstrapping versus one-step updates and exploration strategies, leading to hypotheses about multi-step extensions (MTD3, MSAC). Section \ref{subsec:theory_multistep_bootstrapping} provides theoretical justification for multi-step bootstrapping's temporal smoothing effects under partial observability. Section \ref{sec:experimental_setup} details the experimental methodology, while Section \ref{sec:experimental_results} presents comprehensive experimental validation, performance comparisons, and efficiency analysis. Section \ref{sec:Discussion} synthesizes findings into practical guidelines and discusses limitations.

\section{Related Work}
MDP solutions obtained using DRL have enabled tremendous advancements. For MDPs with discrete action spaces, Deep Q-Network (DQN) \cite{mnih2013playing}, Double DQN \cite{van2016deep}, Dueling DQN \cite{wang2016dueling}, Actor Critic with Experience Replay (ACER)\cite{wang2016sample} have been proposed and shown to achieve or exceed human performance. For continuous action spaces, Trust Region Policy Optimization (TRPO)\cite{schulman2015trust}, PPO \cite{schulman2017proximal}, Deep Deterministic Policy Gradient (DDPG)\cite{lillicrap2015continuous}, TD3 \cite{fujimoto2018addressing}, and SAC \cite{haarnoja2018soft} have been proposed. In addition, there are also DRL algorithms designed to work for both discrete and continuous action spaces, e.g., Advantage Actor Critic (A2C)\cite{mnih2016asynchronous}. There are also works applying DRL to a multi-agent system by modeling the system as a Decentralized MDP (Dec-MDP) \cite{beynier2013dec}. However, all these algorithms are designed for and tested with MDPs that are well engineered with fully-observable states for most cases, and it is unclear whether these algorithms will work well or to what extent they can maintain their relative performance in POMDPs. In this paper, we focus on continuous control tasks and show that PPO, SAC, TD3 do not maintain their relative performance in POMDPs, with PPO underperforming SAC and TD3 in MDPs but outperforming SAC and TD3 in POMDPs.

POMDPs have gained extensive attention in the community and been investigated within both model-free \cite{hausknecht2015deep,lample2017playing,meng2021memory} and model-based \cite{igl2018deep,singh2021structured,shani2013survey} DRL, including POMDPs with discrete\cite{baisero2021unbiased} or continuous\cite{ni2021recurrent} action spaces. \cite{ni2021recurrent} empirically shows that recurrent model-free RL can be a strong baseline for many POMDPs. \cite{subramanian2022approximate} proposes a theoretical framework for approximate planning and learning in POMDPs, where an approximate information state can be used to learn an approximately optimal policy. In addition, multi-agent systems can also be modeled as a Decentralized POMDP (Dec-POMDP) and solved by DRL algorithms \cite{beynier2013dec,oliehoek2016concise}, which demonstrates the broad application of DRL in POMDPs. \cite{arcieri2024pomdp} introduced a method to jointly infer POMDP transition and observation models via MCMC, then train robust domain-randomized policies using DRL. Hierarchical RL  \cite{ishida2024soap} is also used to solve POMDPs and even outperforms memory-based RL on some tasks. \cite{lu2024rethinking} shows that adding Deep Linear Recurrent Units (LRUs) boosts Transformers ability to encode temporal history and provides improved results in POMDPs.

Multi-step methods (also called n-step methods/bootstrapping) refer to RL algorithms that utilize multi-step immediate rewards to estimate the bootstrapped value of taking an action in a state. These have been investigated in the literature for improving reward signal propagation \cite{van2016effective,mnih2016asynchronous,de2018multi,hessel2018rainbow} leading to faster learning speed.  Multi-step bootstrapping has also been shown to help to alleviate the over-estimation problem in \cite{meng2021effect}. However, to the best of our knowledge, there is no work connecting multi-step methods to their potential ability to capture temporal information when solving POMDPs. In this work, we show that PPO with multi-step bootstrapping outperforms TD3 and SAC that only using 1-step bootstrapping in POMDPs. We also empirically show that multi-step bootstrapping helps TD3 and SAC to perform better in POMDPs.

Recent linear-time or retention-based sequence models \cite{shi2023retnet,gu2023mamba} and efficiency-focused attention kernels \cite{dao2023flashattention2} provide alternative paths to encode longer temporal dependencies under partial observability. Integrating such architectures with bias-reducing multi-step targets remains largely unexplored; our empirical horizon analysis supplies baseline variance/performance trade-offs that future hybrid (multi-step + long-context model) methods should surpass.

Outside mainstream DRL, fault-tolerant sampled-data estimation with Markov jump parameters \cite{cao2024exponential} and dissipative sliding mode synchronization of delayed competitive neural networks \cite{subhashri2025robust} address structurally similar uncertainty (delays, actuator failure, inertial coupling). Their emphasis on compensating incomplete or perturbed state channels reinforces the premise that temporal integration can partially bridge observability gaps prior to adopting specialized control-theoretic constructs within learning frameworks.

From the related work on DRL for MDPs and POMDPs, we can see that MDPs and POMDPs are normally tackled separately in DRL and little work is done on bridging MDPs and POMDPs and their solvers. Even though there are some efforts on solving POMDPs by using MDP solvers in conjunction with imitation learning \cite{yoon2008probabilistic,littman1995learning}, \cite{arora2018hindsight} shows that multi-resolution information gathering cannot be addressed using MDP based POMDP solvers by only focusing on one specific task. In addition to inventing algorithms that work for both MDPs and POMDPs, there is also effort studying how to reducing PODMPs to MDPs. For example, \cite{sandikci2010reduction} studies the reduction of a POMDP to a MDP by exploiting grid approximation, but this assumption on a finite discrete belief state space limits applications to continuous state spaces. 

Successfully applying DRL to real applications involves many design choices, with state representation particularly important. For example, \cite{reda2020learning} studies the environment design, including the state representations, initial state distributions, reward structure, control frequency, episode termination procedures, curriculum usage, the action space, and the torque limits, that matter when applying DRL. They empirically show these design choices can affect the final performance significantly. In addition, \cite{ibarz2021train} focuses on investigating the challenges of training real robots with DRL, compared to simulation. As another work related to state representation, \cite{mandlekar2021matters} studies the factors that matter in learning from offline human demonstrations, where observation space design is highlighted as a particularly prominent aspect. These works aim to comprehensively cover broader topics in applying DRL, but in this work we mainly focus on the partial observability problem during DRL for robot control. Moreover, we try to reproduce the problem encountered when applying DRL to novel robots using benchmark tasks where the problem can be more easily reproduced and studied.

As this paper conducts experimental study across both MDPs and POMDPs, it is worth noting bench-marking efforts for DRL algorithms. Particularly, we note that DRL algorithms are often separately bench-marked for MDPs and POMDPs \cite{duan2016benchmarking,ni2021recurrent}. This introduces uncertainty around the performance of DRL algorithms benchmarked on MDPs in POMDPs and 
vice versa. We speculate that this phenomena is partly due to the unbalanced availablity of testbeds for MDPs and POMDPs. There are many MDP testbeds, e.g., Gymnasium \cite{towers_gymnasium_2023}, DeepMind Control Suite \cite{tunyasuvunakool2020dm_control}, and many more \url{https://github.com/clvrai/awesome-rl-envs} than POMDP testbeds. As an example testbed for discrete action spaces, \cite{shao2022mask} proposes Mask Atari for deep reinforcement learning as a POMDP benchmark, where the observation of Atari 2600 games are manipulated by masks in order to create POMDPs. For continuous action spaces, \cite{morad2023popgym} provides Popgym to benchmark partially observable reinforcement learning. In addition to the availability of MDP and POMDP testbeds, MDP testbeds are commonly benchmarked in \cite{duan2016benchmarking,tianshou,stable-baselines}, but because the POMDP bestbeds are not so popular, there are not many popular benchmarks for them neither. Even though Popgym includes a POMDP-version of Cartpole and Pendulum, the action space of these tasks is small compared to the continuous control tasks provided in Gymnasium. Therefore, this paper uses POMDP testbeds modified from MDP testbeds in Gymnasium, exploiting similar techniques to those proposed by \cite{meng2021memory}. 

\section{Background}
\label{sec:background}

A \emph{Markov Decision Process (MDP)}: is a sequential decision process defined as a 4-tuple $S, A, P, R$, where $S$ is the state space, $A$ is the action space, $P(s'| s, a) = p(s_{t+1}=s'|s_t=s, a_t=a)$ is the transition probability that action $a$ in state $s$ at time $t$ will lead to a new state $s'$ at time $t+1$, and $R(s, a, s')\in\mathbb{R}$ provides the immediate reward $r$ indicating how good taking action $a$ is in state $s$ after transitioning to a new state $s'$. In a MDP, it is assumed that the state transitions defined in $P$ satisfy the Markov property, i.e. the next state $s'$ only depends on the current state $s$ and the action $a$. Normally, the state $s$ is not accessible to an agent, but its representation $o$ is given. When the observation $o$ fully captures the current state $s$, we call this a fully observable MDP. However, when the observation $o$ cannot fully represent the current state $s$, we call the decision process a Partially Observable Markov Decision Process (POMDP). In some cases, using a history of past observations and/or actions and/or rewards up to time $t$ as a new observation can reduce a POMDP to MDP. For example, for tasks where the velocity is part of the system state, using a history of past positions of a robot as the observation can make the POMDP, where only position is included in its observation, a MDP. For these cases, history aids with dealing with partial observability.

\emph{Partial Observability in DRL}: A fundamental challenge in POMDPs is the degradation of state information available to the agent. This manifests in several forms \cite{kaelbling1998planning,cassandra1998survey}: (1) \textbf{State aliasing}, where multiple distinct latent states $s_1, s_2, \ldots, s_k$ produce identical observations $o$ (formally, $h(s_i) = h(s_j) = o$ for $s_i \neq s_j$ given observation function $h: S \rightarrow O$); (2) \textbf{Noisy observations}, where the observed state $o_t = s_t + \epsilon_t$ is corrupted by noise $\epsilon_t$ (e.g., sensor noise, quantization errors, communication disturbances); (3) \textbf{Missing observations (flickering)}, where observations are intermittently unavailable due to sensor failures, communication dropouts, or processing delays, requiring agents to act based on stale or interpolated information; and (4) \textbf{Incomplete observations}, where only a subset of state components are observable. All forms create uncertainty about the true underlying state, leading to suboptimal decision-making and increased learning difficulty \cite{shani2013survey}. This work investigates how multi-step bootstrapping provides robustness across these partial observability scenarios commonly encountered in robotic control \cite{meng2021memory}.

\emph{Reinforcement Learning (RL)}: \cite{sutton2018reinforcement} studies how to solve MDPs or POMDPs, without requiring the transition dynamics $P$ to be known.  Specifically, an agent observes $o_t$ at time $t$, then decides to take action $a_t$ according to its current policy  $a_t \sim \pi(a_t|o_t)$. Once the action $a_t$ is taken, the agent observes a new observation $o_{t+1}$ and receives a reward signal $r_t$ from the environment. By continuously interacting with the environment, the agent learns an optimal policy ${\pi}^{*}$ to maximize the expectation of the discounted return $\mathbb{E}_{{\pi}^{*},o_0\sim \rho_0}\left [ \sum_{
t=0}^{T} \gamma^{t}r_t | o_0 \right ]$ starting from initial observation $o_0\sim\rho_0$, where the discount factor $\gamma$ is used to balance the short-term vs. long-term return.

There are three functions that are commonly used in RL algorithms. A state-value function $V^{\pi}(s)$ of a state $s$ under a policy $\pi$ is the expected return when starting in $s$ and following $\pi$ thereafter, which can be formally defined by $V^{\pi}(s)=\mathbb{E}_{\pi}\left [ \sum_{i=0}^{T-t} \gamma^{i}r_{t+i} | s_t=s \right ]$. An action-value function $Q(s,a)$ of taking action $a$ in state $s$ and following $\pi$ afterwards can be defined as $Q^{\pi}(s,a)=\mathbb{E}_{\pi}\left [ \sum_{i=0}^{T-t} \gamma^{i}r_{t+i} | s_t=s, a_t=a \right ]$. The advantage $A^{\pi}(s,a)$ of taking action $a$ in state $s$ is defined as $Q^{\pi}(s,a) - V^{\pi}(s)$ when following a policy $\pi$. For these functions, if the state $s$ is not directly observable, its representation $o$ will be used. To simplify the notation, for the rest of this paper, we will use $o$ to represent $s$.

\subsection{Deep Reinforcement Learning}
Deep Reinforcement Learning (DRL) employs Deep Neural Networks to represent these value functions and/or policy. In this paper, we will use three of the most popular DRL algorithms which will be briefly introduced in the following.

Both \textbf{\textit{Soft Actor-Critic (SAC)}} \cite{haarnoja2018soft} and \textbf{\textit{Twin Delayed Deep Deterministic Policy Gradient (TD3)}} \cite{fujimoto2018addressing} are off-policy actor-critic DRL approaches. Both employ two neural networks to approximate two versions of the state-action value function (named \emph{critics}) $Q_{i=1,2}$ parameterized by $\theta_{i=1,2}$.  Learning two versions of the critic is used to address the function approximation error by taking the minimum over the bootstrapped Q-values in the next observation $o_{t+1}$. Unlike TD3, SAC gives a bonus reward to an agent at each time step, proportional to the entropy of the policy at that timestep. In addition, TD3 learns a deterministic policy $\mu$ parameterized by $\phi$, whereas SAC learns a stochastic policy $\pi_{\psi}$ parameterized by $\psi$. Given a mini-batch of experiences $(o_t,a_t,r_t,o_{t+1})$ uniformly sampled from the replay buffer $D$, the target bootstrapped Q-value $\hat{Q}(o_t,a_t)$ of taking action $a_t$ in observation $o_t$ can be defined as follows:
\begin{equation}
    \hat{Q}(o_t,a_t) = r_t + \gamma \left [ \min_{i=1,2}  Q_{\theta_{i}^{-}}\left ( o_{t+1},  a^{-} \right ) + \alpha H(\pi(\cdot | o_{t+1})) \right ]
    \label{eq:target_Q_TD3_and_SAC}
\end{equation}
where the target Q-value functions are parameterized by $\theta_{i}^{-}$, $a^{-}=\mu_{\phi^{-}}\left ( o_{t+1} \right )$ for TD3 with target policy $\mu_{\phi^{-}}$ and $a^{-}\sim\pi_{\psi^{-}}\left ( a| o_{t+1} \right )$ for SAC with target policy $\pi_{\psi^{-}}$, and $\alpha \geq 0$ balances the maximization of the accumulated reward and entropy.  For TD3 $\alpha=0$. Then, $Q_i$ can be optimized by minimizing the expected difference between the prediction and the bootstrapped value with respect to parameters $\theta_{i}$, following
\begin{equation}
    \min_{\theta_{i}} \mathbb{E}_{(o_t,a_t,r_t,o_{t+1})\sim D}^{}\left [ Q_{\theta_{i} }(o_t,a_t) - \hat{Q}(o_t,a_t) \right ]^{2}.
\end{equation}
For TD3, the policy is optimized by maximizing the expected Q-value over the mini-batch of $o_t$ with respect to the policy parameter $\phi$, following 
\begin{equation}
    \max_{\phi} \mathbb{E}_{o_t\sim D}Q_{\theta_{1} }(o_t,\mu_{\phi}(o_t)),
    \label{eq:td3_policy_loss}
\end{equation}
and for SAC the policy is updated to maximize the expected Q-value on $(o_t, a)$ where $a$ is sampled from policy $\pi_{\psi}(\cdot| o_t)$ and the expected entropy of $\pi$ in observation $o_t$ as 
\begin{equation}
    \max_{\psi } \mathbb{E}_{o_t\sim D}\left [ \min_{i=1,2}Q_{\omega _{i} }(o_t,a)\mid_{a\sim \pi_{\psi}(a|o_t)} + \alpha H(\pi_{\psi}(\cdot, o_t))  \right ].
    \label{eq:sac_policy_loss}
\end{equation}

\textbf{Proximal Policy Optimization (PPO)} \cite{schulman2017proximal} optimizes a policy by taking the biggest possible improvement step using the data collected by the current policy, but at the same time limiting the step size to avoid performance collapse. A common way to achieve this is to attenuate policy adaptation. Formally, for a set of observation and action pairs $(o_t,a_t)$ collected from the environment based on the current policy $\pi_{\varphi_k}$, the new policy $\pi_{\varphi}$ is obtained by maximizing the expectation over the loss function $L(o_t, a_t, \varphi_k, \varphi)$ with respect to the policy parameter $\varphi$ as $\max_{\varphi} \mathbb{E}_{(o_t,a_t)\sim \pi_{\varphi_k}} L(o_t, a_t, \varphi_k, \varphi)$. The loss function $L$ is defined as which is calculated based on the $T$ experiences collected by the current policy. In \eqnref{eq:ppo_monte_carlo_return}, $d_T$ indicates after $T$ steps if the agent reached the terminal state or not, and $o_{T+1}$ is the next observation after $T$ steps.
\begin{equation}
    \begin{aligned}
    L(o_t, a_t, \varphi_k, \varphi) = \min \left (c \times A^{\pi_{\varphi_k}}(o_t,a_t), \text{clip}(c, 1-\epsilon, 1+\epsilon)A^{\pi_{\varphi_k}}(o_t,a_t)  \right ), 
    \end{aligned}
    \label{eq:ppo_policy_loss}
\end{equation}
where $c = \frac{\pi_\varphi(a_t|o_t) }{\pi_{\varphi_k}(a_t|o_t)}$, $\epsilon$ is a small hyperparameter that roughly says how far away the new policy is allowed to go from the current one, and $A^{\pi_{\varphi_k}}(o_t,a_t)$ is the advantage value. A common way to estimate the advantage is called generalized advantage estimator GAE($\lambda$) \cite{schulman2015high}, based on the \lamreturn{} and the estimated state-value $V(o_t)$ in observation $o_t$ as $A^{\pi_{\varphi_k}}(o_t,a_t) = G_t(\lambda)-V_\upsilon(o_t)$. The $G_t(\lambda)$ is defined by 
\begin{equation}
    G_t(\lambda) = (1-\lambda)\sum_{n=1}^{T-t-1}\lambda^{n-1}R_{t}^{(n)} + \lambda^{T-t-1}R_{t}^{(T-t)}
    \label{eq:ppo_lambda_return}
\end{equation}
where $\lambda\in[0,1]$ balances the weights of different multi-step returns and the summation of the coefficients satisfies $1 = (1-\lambda)\sum_{n=1}^{T-t-1}\lambda^{n-1}+\lambda^{T-t-1}$.  Particularly, $R_{t}^{(n)}$ is defined as $ R_t^{(n)} = \sum_{i=t}^{t+n-1} \gamma^{i-t}r_t+\gamma^{n}V_\upsilon(o_{t+n})$ that is bootstrapped by state-value $V_\upsilon(o_{t+n})$ in observation $o_{t+n}$. The state-value function $V_{\upsilon}$ is optimized to minimize the mean-square-error between the predicted state value $V_\upsilon(o_t)$ and the Monte-Carlo return as $\min_{\upsilon } \mathbb{E}_{o_t}\left [ V_\upsilon (o_t) - R_t \right ]^{2}$, where the Monte-Carlo return $R_t$ is defined as
\begin{equation}
    R_t=\sum_{i=t}^{T}\gamma^{i-t}r_t + (1-d_T)\gamma^{T}V_\upsilon (o_{T+1}),
    \label{eq:ppo_monte_carlo_return}
\end{equation}
which is calculated based on the $T$ experiences collected by the current policy. In Eq. \ref{eq:ppo_monte_carlo_return}, $d_T$ indicates after $T$ steps if the agent reached the terminal state or not, and $o_{T+1}$ is the next observation after $T$ steps.

\textbf{Multi-step Methods} (also called $n$-step methods) \cite{sutton2018reinforcement} refer to RL algorithms utilizing multi-step bootstrapping. Formally, for a multi-step bootstrapping where $n$ is the step size after which a bootstrapped value will be used, the $n$-step bootstrapping of $Q(o_t, a_t)$ can be defined as 
\begin{equation}
    Q^{(n)}(o_t, a_t) = \sum_{i=t}^{t+n-1} \gamma^{i-t}r_t+\gamma^{n}Q(o_{t+n}, a),
\end{equation}
or $n$-step bootstrapping of $V(o_t)$
\begin{equation}
    V^{(n)}(o_t) = \sum_{i=t}^{t+n-1} \gamma^{i-t}r_t+\gamma^{n}V(o_{t+n}),
\end{equation}
where $a \sim \pi \left ( a|o_{t+n} \right )$ and $ \left\{ \left ( o_{t+i},a_{t+i},r_{t+i}, o_{t+i+1} \right )\right\}_{i=0}^{n-1}$ are a sequence of experiences following $\pi$. Note that when $n=1$, it reduces to 1-step bootstrapping, which is commonly used in Temporal Difference (TD) based RL.

\section{Exemplar Robot Control Problem}
\label{sec:Exemplar_Robot_Control_Problem}

\begin{figure*}[htp!]
    \centering
    \begin{minipage}[c]{.7\textwidth}
        \begin{subfigure}[b]{0.25\linewidth}
             \centering
             \includegraphics[width=\linewidth]{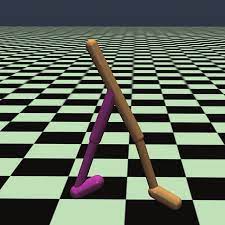}
             \vspace{.05cm}
             \caption{Walker2D}
             \label{fig:demo_walker2d_image}
        \end{subfigure}
        \begin{subfigure}[b]{.35\linewidth}
             \centering
             \includegraphics[width=\linewidth]{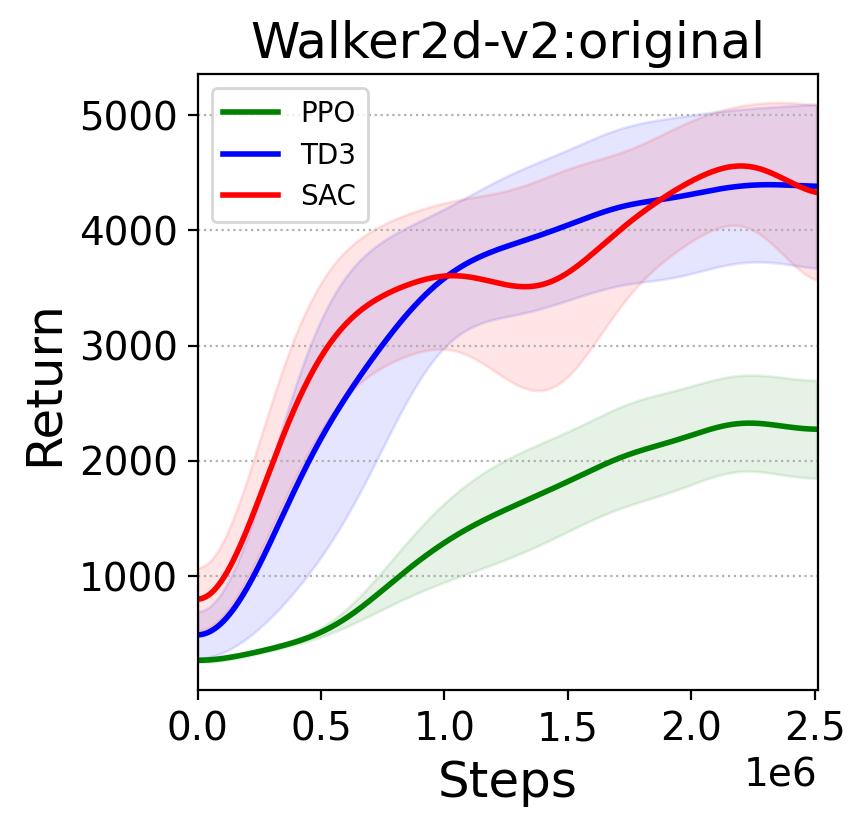}
             \caption{Original Task}
             \label{fig:demo_walker2d_original_task}
        \end{subfigure}
        \begin{subfigure}[b]{.35\linewidth}
             \centering
             \includegraphics[width=\linewidth]{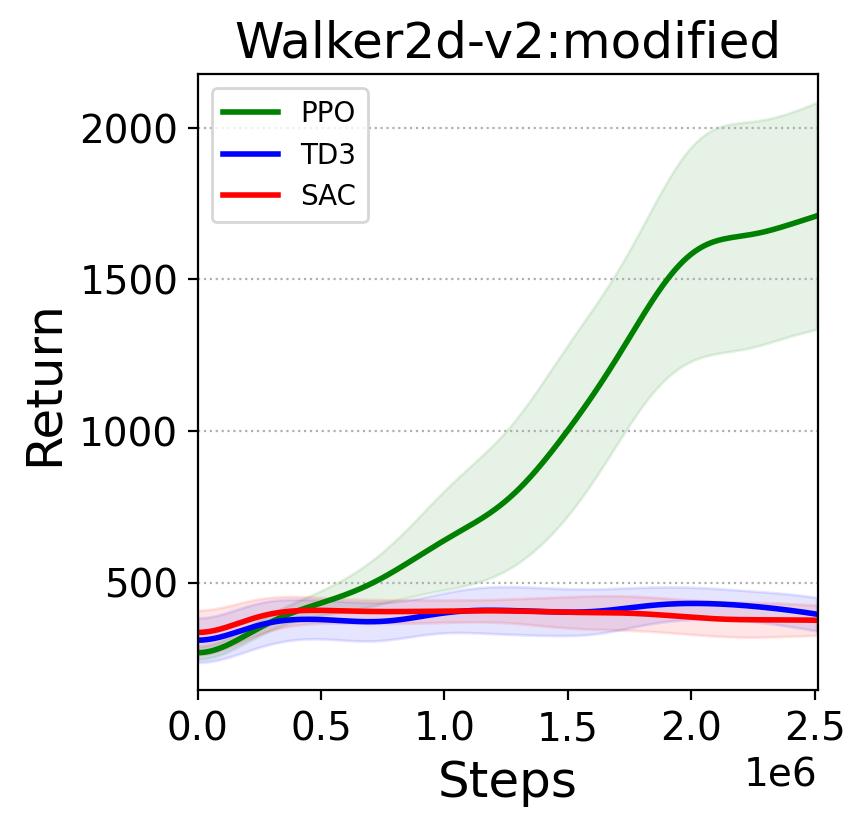}
             \caption{Modified Task}
             \label{fig:demo_walker2d_modified_task}
        \end{subfigure}
    \end{minipage}
    \begin{minipage}[c]{0.28\textwidth}
        \caption{Performance of PPO, TD3 and SAC on Walker2D, where (a) shows the Walker2D robot, (b) shows the performance of the three DRL algorithms on the original task, and (c) shows their performance on the modified task with random noise in the task observation.}
        \label{fig:demo_walker2d_task}
    \end{minipage}
\end{figure*}

We present a motivating example illustrating how induced observation corruption reshapes algorithm rankings. The Walker2D Mujoco task (Fig. \ref{fig:demo_walker2d_image}) in Gymnasium (\url{https://gymnasium.farama.org}) is used in two forms: an \textbf{original} fully observed MDP including joint positions and velocities, and a \textbf{modified} partially observed variant with i.i.d. Gaussian noise $\epsilon \sim \mathbb{N}(0,\,0.1)$ added per observation component ($o' = o + \epsilon$). This simple transformation preserves reward structure while degrading state fidelity, creating controlled partial observability through noisy observations.
Such noise emulates real robotic sensing imperfections, e.g., encoder jitter, analog quantization, transient EMI, etc. The resulting gap between latent physical state and received observation elevates PPO relative to TD3/SAC, foreshadowing the broader inversions we later generalize.

Baseline expectation (per prior MDP benchmarks \cite{fujimoto2018addressing,haarnoja2018soft}) is TD3/SAC $>$ PPO. The observed inversion (Fig. \ref{fig:demo_walker2d_modified_task})—PPO retaining stability while TD3/SAC degrade—initially appears anomalous. A parallel observation in interactive system control \cite{meng2023learning} suggests the effect is not task‑idiosyncratic. We treat this inversion as a candidate \textit{diagnostic}: unexpected PPO dominance may signal partial observability or mis-specified observation design. Table \ref{tab:Expected_VS_Unexpected_Result} formalizes the expectation vs. empirical inversion.

\begin{table}[htp!]
    \caption{Expected vs Unexpected Result}
    \label{tab:Expected_VS_Unexpected_Result}
    \small
    \centering
    \begin{tabular}{r|l}
        \hline
        \textbf{Expected}\label{expected_result} & \makecell[l]{TD3 and SAC outperform PPO on MDP tasks\\ modified to be POMDP.} \\\hline
        \textbf{Unexpected}\label{unexpected_result} & \makecell[l]{TD3 and SAC underperform PPO on MDP tasks\\ modified to be POMDP.}\\\hline
    \end{tabular}
\end{table}

Because the noisy sensor setting in the exemplar control task introduced in Section \ref{sec:Exemplar_Robot_Control_Problem} is common in practise, it is worth investigating to what extent the \textbf{performance inversion} will generalize to other benchmark tasks and other types of POMDPs studied in RL literature and understand its underlying causes.

\section{Generalization of the Performance Inversion on Other Tasks}
\label{sec:Generalization_of_the_Unexpected_Result_on_Other_Tasks}

The Walker2D example suggests that performance inversion under observation degradation may be a general phenomenon rather than task-specific. To test this systematically, we extend the evaluation across multiple MuJoCo continuous control benchmarks representing diverse morphologies and control challenges: Ant (quadruped locomotion), HalfCheetah (planar running), Hopper (single-leg hopping), and Walker2D (bipedal walking).

Beyond the simple additive noise used in the motivating example, we investigate four distinct mechanisms for inducing partial observability, each capturing different real-world sensor imperfections and information loss patterns. These variants systematically probe how different types of observation degradation affect algorithm robustness while maintaining identical reward structures for fair comparison. The detailed design and rationale for each variant are presented in Section \ref{sec:pomdp_variants}.

To experimentally validate the generalization, we formally make the \textbf{Hypothesis \ref{hypothesis_generalization}} as: 
\begin{hypothesis}
\label{hypothesis_generalization}
If the partial observability introduced by adding random noise caused the performance inversion relative to the expectation on MDP, then similar results should be reproducible on MDP vs. POMDP versions of other benchmark tasks. Concretely, TD3 and SAC are expected to outperform PPO on MDPs, but underperform PPO on POMDPs.
\end{hypothesis}

Confirmation of \textbf{Hypothesis \ref{hypothesis_generalization}} would have significant practical implications: (i) published MDP superiority of TD3/SAC should not be naively extrapolated to tasks with uncertain observability; (ii) unexpected PPO $>$ TD3/SAC performance can serve as a practical diagnostic signal prompting systematic audit of observation completeness and sensor reliability.

\section{Analysing and Improving Robustness to Partial Observability}
\label{sec:Analysing_and_Improving_Robustness_to_Partial_Observability}

Having reproduced the unexpected robustness inversion, we now provide theoretical analysis to understand the underlying mechanisms before presenting experimental validation. We systematically analyze the key algorithmic differences that plausibly drive robustness under partial observability: (i) temporal credit assignment depth (multi-step vs one-step bootstrapping); (ii) exploration strategy (conservative policy updates vs aggressive exploration). Through formal theoretical foundations (Section \ref{subsec:theory_multistep_bootstrapping}, Section \ref{subsec:theory_exploration_strategies}) followed by controlled experimental studies, we demonstrate how these factors interact with different forms of partial observability.

We consider other distinctions (on- vs off-policy data reuse, actor-critic vs policy-gradient class, replay buffer scale) less central. Prior work \cite{mnih2015human,zhang2017deeper} links buffer size to variance and correlation, but our observations show TD3/SAC fragility persists even with ample buffers, implying horizon and update style dominate robustness effects.

We therefore focus on two key algorithmic differences that most directly explain the performance inversion observed in this work:
\begin{diff}
\label{diff_multistep_bootstrapping}
PPO exploits multi-step bootstrapping for both advantage and state-value function estimation, while TD3 and SAC only use one-step bootstrapping for action-value function estimation.
\end{diff}
\begin{diff}
\label{diff_exploration}
PPO updates its policy conservatively leading to gentle exploration, TD3 explores by adding Gaussian action noise with fixed standard deviation to its deterministic policy, whereas SAC always encourages exploration.
\end{diff}

We analyze each difference in detail below, developing theoretical foundations and testable hypotheses. Experimental validation of these hypotheses is presented in Section \ref{sec:experimental_results}.

\subsubsection{Theoretical Foundation: Multi-step Bootstrapping and Information Recovery}
\label{subsec:theory_multistep_bootstrapping}

For \textbf{Diff \ref{diff_multistep_bootstrapping}}, PPO uses \lamreturn{} defined in \eqnref{eq:ppo_lambda_return}, which is a weighted average of $n$-step returns where $n\in[1,T-t-1]$, to calculate the advantage $A^{\pi_{\varphi_k}}(o_t,a_t)$ of taking action $a_t$ in observation $o_t$, and then uses the Monte-Carlo return defined in \eqnref{eq:ppo_monte_carlo_return} to update its state-value function. However, TD3 and SAC only use $1$-step bootstrapping to calculate their target Q-value as defined in \eqnref{eq:target_Q_TD3_and_SAC}.

On MDPs, it is known that multi-step bootstrapping can propagate the learning signal, i.e., reward, faster \cite{sutton2018reinforcement} leading to faster learning speed, and in DRL multi-bootstrapping \cite{meng2021effect} also has the effect of alleviating the overestimation problem. However, neither of these explain the \textbf{Unexpected Result }. Concretely, considering the dense reward signal used in the task, even without the faster learning signal propagation endowed by the multi-step bootstrapping, TD3 and SAC should also improve gradually, which is not observed in Fig. \ref{fig:demo_walker2d_modified_task}. In addition, both TD3 and SAC utilize the minimum of the two critics (\eqnref{eq:target_Q_TD3_and_SAC}) to calculate the bootstrapped Q-value so that they are less likely to be affected by the overestimation problem.

\textbf{Information-Theoretic Foundation}

We establish a formal theoretical basis for how rewards encode state information and how multi-step accumulation recovers temporal dynamics under partial observability, building on information-theoretic approaches to POMDPs \cite{shani2013survey} and multi-step return analysis \cite{sutton2018reinforcement,de2018multi}.

\textbf{POMDP Framework:} Consider a POMDP with state space $S$, observation space $O$, and observation function $h: S \rightarrow O$. While the agent observes $o_t = h(s_t)$ (possibly with noise), the reward $r_t = R(s_t, a_t, s_{t+1})$ depends on the true underlying state transitions. The reward sequence therefore contains information about latent state dynamics that is not directly accessible through observations alone.

\textbf{Definition 1 (Reward Information Content).} For a POMDP, the \emph{reward information content} is defined as $\mathcal{I}_r(s_t) = I(s_t; r_t)$, where $I(s_t; r_t)$ denotes the mutual information between the true state $s_t$ and the reward $r_t$. Similarly, the \emph{observation information content} is $\mathcal{I}_o(s_t) = I(s_t; o_t)$.

\textbf{Definition 2 (Informative Rewards).} Rewards are \emph{informative} if $I(s_t; r_t) > 0$, meaning that observing the reward reduces uncertainty about the true state. This occurs when: (1) different states (or state-action pairs) yield different expected rewards, and (2) the reward function exhibits sufficient variability to distinguish between states. Non-informative rewards (e.g., constant rewards or rewards independent of state) provide $I(s_t; r_t) = 0$ and cannot assist in state information recovery.

\textbf{Fundamental Information-Theoretic Relationships:} Following standard information-theoretic analysis \cite{cover2012elements}, we establish the key relationships under partial observability with informative rewards:
\begin{equation}
I(s_t; r_t) > 0 \text{ and } I(s_t; o_t) < \log|S|
\label{eq:reward_information_content}
\end{equation}
The first condition ensures rewards contain state information, while the second establishes the partial observability constraint. This forms the foundation for multi-step reward aggregation as a state information recovery mechanism.

\textbf{Theorem 1 (Multi-step Information Recovery).} Under partial observability, the $n$-step reward sequence $\{r_t, r_{t+1}, \ldots, r_{t+n-1}\}$ contains strictly more information about the latent state trajectory $\{s_t, s_{t+1}, \ldots, s_{t+n}\}$ than any single reward $r_k$.

\textit{Proof Sketch:} Let $I(X;Y)$ denote mutual information between random variables $X$ and $Y$ \cite{cover2012elements}. For the latent state sequence $S_t^{t+n} = \{s_t, \ldots, s_{t+n}\}$ and reward sequences:
\begin{align}
I(S_t^{t+n}; r_t) &\leq I(S_t^{t+n}; \{r_t, r_{t+1}, \ldots, r_{t+n-1}\})
\label{eq:info_inequality}
\end{align}

The inequality is strict when rewards exhibit temporal correlation and the observation function $h(\cdot)$ loses information, since each additional reward $r_{t+k}$ provides incremental information about state transitions that cannot be recovered from $r_t$ alone under partial observability. This follows from the data processing inequality and the fact that conditioning on additional correlated variables can only increase mutual information \cite{cover2012elements}. 

\textbf{Corollary 1 (Temporal Aggregation Reduces Observability Gap).} The observability gap $\mathcal{G}(s_t, o_t) = I(s_t; o_t)$ measures state uncertainty remaining after observation. Multi-step reward aggregation reduces this gap because the discounted sum $G_t^{(n)} = \sum_{k=0}^{n-1} \gamma^k r_{t+k}$ provides additional discriminative information about state sequences. Formally:
\begin{equation}
\mathcal{G}(S_t^{t+n}, O_t^{t+n}) < \mathcal{G}(s_t, o_t)
\label{eq:gap_reduction}
\end{equation}
where the gap reduction occurs because algorithms using $G_t^{(n)}$ have access to $I(S_t^{t+n}; O_t^{t+n}, G_t^{(n)}) > I(s_t; o_t)$ when rewards exhibit sufficient temporal correlation to distinguish between different state trajectories.

\textbf{Theorem 2 (Bias-Variance Trade-off in Multi-step Returns).} Under partial observability, the $n$-step return $G_t^{(n)} = \sum_{k=0}^{n-1} \gamma^k r_{t+k} + \gamma^n V(o_{t+n})$ exhibits a bias-variance trade-off \cite{sutton2018reinforcement,van2016effective}:

\textit{Bias Component:} $\text{Bias}[G_t^{(n)}] = \mathbb{E}[G_t^{(n)}] - V^*(s_t)$ decreases with increasing $n$ due to improved state information recovery, following classical temporal difference analysis \cite{sutton2018reinforcement}.

\textit{Variance Component:} $\text{Var}[G_t^{(n)}] = \mathbb{E}[(G_t^{(n)} - \mathbb{E}[G_t^{(n)}])^2]$ initially decreases then increases with $n$, achieving minimum at optimal horizon $n^*$, consistent with multi-step return variance analysis \cite{de2018multi,van2016effective}.

\textit{Derivation of Optimal Horizon:} The mean squared error (MSE) of the $n$-step estimator is:
\begin{equation}
\text{MSE}[G_t^{(n)}] = \mathbb{E}[(G_t^{(n)} - V^*(s_t))^2] = \text{Bias}[G_t^{(n)}]^2 + \text{Var}[G_t^{(n)}]
\label{eq:mse_decomposition}
\end{equation}
This is the standard bias-variance decomposition \cite{hastie2009elements}. The optimal horizon minimizes this MSE:
\begin{equation}
n^* = \arg\min_n \left\{ \text{Bias}[G_t^{(n)}]^2 + \text{Var}[G_t^{(n)}] \right\}
\label{eq:optimal_horizon}
\end{equation}

\textbf{Proposition 1 (Robustness Mechanism).} Algorithms using multi-step bootstrapping ($n > 1$) exhibit greater robustness to partial observability than single-step methods ($n = 1$) through three complementary mechanisms:

1. \textit{Information Recovery (from Theorem 1):} Multi-step returns implicitly reconstruct latent state information through temporal reward aggregation. Specifically, Theorem 1 establishes that $I(S_t^{t+n}; \{r_t, \ldots, r_{t+n-1}\}) > I(s_t; r_t)$, meaning the $n$-step reward sequence contains strictly more information about the state trajectory than any single reward. This additional information enables better state estimation under partial observability.

2. \textit{Optimal Information-Variance Trade-off (from Theorem 2):} The bias-variance analysis in Theorem 2 shows that there exists an optimal horizon $n^*$ (Eq. \ref{eq:optimal_horizon}) that balances information recovery with variance control. Multi-step methods can operate near this optimum, while single-step methods ($n=1$) are constrained to a suboptimal point with higher bias due to insufficient information recovery.

3. \textit{Gap Reduction Mechanism (from Corollary 1):} The observability gap reduction $\mathcal{G}(S_t^{t+n}, O_t^{t+n}) < \mathcal{G}(s_t, o_t)$ (Eq. \ref{eq:gap_reduction}) occurs because algorithms using $G_t^{(n)}$ have access to additional discriminative information about state sequences through temporal reward aggregation.

\textbf{Theoretical Synthesis:} This theoretical framework explains why PPO (which uses $\lambda$-returns combining multiple time scales) outperforms TD3/SAC (single-step) under partial observability, and predicts that multi-step extensions (MTD3, MSAC) should recover robustness.

\subsubsection{Experimental Hypotheses for Multi-step Bootstrapping}
\label{subsec:hypotheses_multistep_bootstrapping}

Based on the theoretical foundations established above, we formulate testable hypotheses to validate the information recovery mechanism empirically.

\begin{hypothesis}
\label{hypothesis_compare_n_step_TD3_and_SAC_to_vanilla}
The \lamreturn{} (\eqnref{eq:ppo_lambda_return}) and Monte-Carlo return (\eqnref{eq:ppo_monte_carlo_return}) based on multi-step bootstrapping\footnote{\lamreturn{} is a weighted mixture of various $n$-step bootstrappings, and Monte-Carlo return is a T-step bootstrapping} employed in PPO leads to robustness to POMDP settings, therefore (1) multi-step/$n$-step versions of TD3 and SAC with $n>1$ should also improve robustness to POMDPs when compared to their vanilla versions, and (2) replacing the \lamreturn{} and Monte-Carlo return with 1-step bootstrapping should cause PPO's performance to decrease when moving from a MDP to a POMDP.
\end{hypothesis}

To empirically verify (1) of the \textbf{Hypothesis \ref{hypothesis_compare_n_step_TD3_and_SAC_to_vanilla}}, we investigate the performance of the multi-step (also known as $n$-step) variants of vanilla TD3 and SAC, namely Multi-step TD3 (MTD3) \cite{tang2020self} and Multi-step SAC (MSAC) \cite{bai2021empirical}, on various tasks. Specifically, instead of sampling a mini-batch \\$\left \{ (o_t,a_t,r_t,o_{t+1})^{(k)} \right \}_{k=1}^{K}$ of $K$ 1-step experiences, we sample a mini-batch \\ $\left \{ (o_t,a_t,r_t,o_{t+1}, \cdots , o_{t+n-1},a_{t+n-1},r_{t+n-1},o_{t+n})^{(k)} \right \}_{k=1}^{K}$ of $K$ $n$-step experiences. Specifically, each sample in the mini-batch used in MTD3 and MSAC is composed of $n$ subsequent experiences. This is realized by storing all experiences, collected by the agent, sequentially in the experience replay buffer, then randomly selecting the $i$-th experience and its subsequent $\min(n-1, T-i)$ experiences ($T$ indicates the end of the episode of the $i$-experience) where if the episode of the $i$-th experience ends before the $(i+n-1)$-th experience, only the experiences before the end of the episode, i.e., $\{i, \cdots, T\}$, are included in that sample. Then, we replace the target Q-value calculation defined in \eqnref{eq:target_Q_TD3_and_SAC} with the $n$-step bootstrapping defined as 
\begin{equation}
    \begin{aligned}
     \hat{Q}^{(n)}(o_t,a_t) = \sum_{i=t}^{t+n-1}\gamma^{i-t} r_t + \gamma^{n} \left [ \min_{i=1,2} Q_{\theta_{i}^{-}}(o_{t+n}, a) + \alpha H(\pi(\cdot | o_{t+n})) \right ]
    \end{aligned}
\end{equation}
where $a=\mu_{\phi^{-}}\left ( o_{t+n} \right )$ and $a\sim\pi_{\psi^{-}}\left ( a| o_{t+n} \right )$ for MTD3 and MSAC, respectively, $\alpha=0$ for MTD3, and the policy update is the same as that for TD3 and SAC. For (2) of the \textbf{Hypothesis \ref{hypothesis_compare_n_step_TD3_and_SAC_to_vanilla}}, we replace both $\lambda$-return $G_t(\lambda)$ and Monte-Carlo return $R_t$ with $n$-step return $R_t^{(n)}$, i.e., $G_t(\lambda)=R_t^{(n)}$ and $R_t=R_t^{(n)}$ where when $n=1$ only 1-step bootstrapping is used. Note that only modifying $\lambda$ cannot reduce PPO to 1-step bootstrapping, because as indicated in \eqnref{eq:ppo_lambda_return}, $\lambda$ only controls the weights of multi-step returns $R_t^{(n)}$, while $G_t(\lambda)$ is a weighted summation of multi-step returns. Similarly, for state-value function updates, in order to control the step size in bootstrapping, $R_t^{(n)}$ is used to calculate the target state-value instead of Monte-Carlo return.

In \textbf{Hypothesis \ref{hypothesis_compare_n_step_TD3_and_SAC_to_vanilla}}, we assume multi-step bootstrapping can improve DRL algorithms' robustness to POMDP, inspired by the observation that the concatenation of multi-step observations significantly improves DRL algorithms' performance on POMDPs and the intuition that reward can be seen as an approximation to the state-transition. Given that, it is natural to analogously ask if replacing the original one-step reward $r_t$ with an accumulated reward $r_t^{avg(n)}$ or $r_t^{sum(n)}$ can achieve similar performance improvement, since this can also incorporate multi-step rewards that may possibly capture some temporal information. In other words, \textbf{Hypothesis \ref{hypothesis_compare_n_step_TD3_and_SAC_to_vanilla}} expects performance improvement when replacing \eqnref{eq:one_step_bootstrapping} with \eqnref{eq:multi_step_bootstrapping}, and similarly we can expect performance improvement when replacing \eqnref{eq:one_step_bootstrapping} with \eqnref{eq:accumulated_reward_bootstrapping} and comparable performance between \eqnref{eq:multi_step_bootstrapping} and \eqnref{eq:accumulated_reward_bootstrapping}.
\begin{equation}
    \hat{Q}(o_t,a_t) = r_t + \gamma \text{max}_a Q(o_{t+1}, a)
    \label{eq:one_step_bootstrapping}
\end{equation}
\begin{equation}
    \hat{Q}(o_t,a_t) = \sum_{k=0}^{n-1} \gamma^{k}r_{t+k} + \gamma^n \text{max}_a Q(o_{t+1}, a)
    \label{eq:multi_step_bootstrapping}
\end{equation}
\begin{equation}
    \begin{split}
    \hat{Q}(o_t,a_t) &= r_t^{avg(n)} + \gamma \text{max}_a Q(o_{t+1}, a)\\ 
            \text{               or } & \\
    \hat{Q}(o_t,a_t) &= r_t^{sum(n)} + \gamma \text{max}_a Q(o_{t+1}, a)\\ 
    \end{split}
    \label{eq:accumulated_reward_bootstrapping}
\end{equation}
Following this idea, we make a hypothesis as follows:

\begin{hypothesis}
\label{hypothesis_compre_accumulated_reward_task_to_one_step_reward_task}
If multi-step rewards can capture temporal information, the performance on tasks with accumulated reward over a few steps should be better than that on environments with one-step reward, i.e., the original reward, and achieve comparable performance to algorithms using multi-step bootstrapping on environments with one-step reward.
\end{hypothesis}

To validate \textbf{Hypothesis \ref{hypothesis_compre_accumulated_reward_task_to_one_step_reward_task}}, in \eqnref{eq:accumulated_reward_bootstrapping} we modified the original task reward ${r}_t$ in time step $t\in \left [ 1, \text{max\_step} \right ]$, where $\text{max\_step}$ indicates the maximum episode steps, as an accumulated reward $r_t^{avg(n)}$ or $r_t^{sum(n)}$ of $n$-step ($n\in \left [ 1, \text{max\_step} \right ]$) rewards as follows:
\begin{equation}
    r_t^{avg(n)} = \frac{1}{\min(n, t)} \sum_{k=0}^{\min(n, t)} {r}_{t-k}
    \label{eq:accumulated_reward_env_avg}
\end{equation}
and
\begin{equation}
    r_t^{sum(n)} = \sum_{k=0}^{\min(n, t)} {r}_{t-k},
    \label{eq:accumulated_reward_env_sum}
\end{equation}
where \eqnref{eq:accumulated_reward_env_avg} and \eqnref{eq:accumulated_reward_env_sum} correspond to the average and summation of the rewards. The modified reward $r_t^{avg(n)}$ or $r_t^{sum(n)}$ will be used for policy learning, but for performance comparison the original reward ${r}_t$ will be used.

\subsubsection{Theoretical Foundation: Exploration Under Partial Observability}
\label{subsec:theory_exploration_strategies}

Under partial observability, the mapping from latent states to observations becomes degraded through various mechanisms: state aliasing ($h(s_1) = h(s_2) = o$ for distinct states), noisy observations ($o_t = s_t + \epsilon_t$), or incomplete state representations \cite{shani2013survey,cassandra1998survey}. This fundamentally alters the exploration-exploitation trade-off and requires theoretical analysis of how different exploration strategies interact with degraded state information, building on classical POMDP theory \cite{kaelbling1998planning} and modern exploration analysis \cite{osband2016deep}.

\textbf{Definition 3 (Partial Observability-Induced Exploration Risk).} Under partial observability, the agent receives degraded observations rather than perfect state information. Let $h(s)$ be the ideal observation function mapping state $s$ to its complete observation $\tilde{o}$. Observation degradation $\mathcal{D}$ corrupts this mapping: $o_t = \mathcal{D}(\tilde{o}_t)$ where $\tilde{o}_t = h(s_t)$, encompassing information removal, noise corruption, flickering, or sensor dropout. The \emph{exploration risk} for action $a$ at degraded observation $o$ is:
\begin{equation}
\mathcal{R}(a|o) = \sum_{s} P(s|o) \cdot \mathcal{L}(a, a_s^*)
\label{eq:exploration_risk}
\end{equation}
where $P(s|o)$ is the agent's belief distribution over true states $s$ given degraded observation $o$, and $\mathcal{L}(a, a_s^*)$ is the loss from taking action $a$ instead of optimal action $a_s^*$ in state $s$. Note that the agent typically does not know the degradation mechanism $\mathcal{D}$, making the belief estimation challenging and increasing exploration risk under uncertainty.

\textbf{Theorem 3 (Conservative Exploration Under Partial Observability).} Adapting regret analysis techniques from bandit theory \cite{russo2018tutorial,lattimore2020bandit}, under partial observability, conservative exploration strategies exhibit lower expected regret than aggressive exploration:
\begin{equation}
\mathbb{E}[\text{Regret}_{\text{conservative}}] \leq \mathbb{E}[\text{Regret}_{\text{aggressive}}]
\label{eq:conservative_advantage}
\end{equation}
when the exploration risk $\mathcal{R}(a|o) > \tau$ for some threshold $\tau > 0$, where $\mathcal{R}(a|o)$ is defined in Definition 3 and captures the expected loss from degraded observations across all four degradation mechanisms (information removal, noise corruption, flickering, sensor dropout).

\textit{Proof Sketch:} Following regret decomposition techniques from \cite{lattimore2020bandit}, aggressive exploration increases the probability of selecting actions that may be optimal under one belief state $P(s|o)$ but suboptimal for the true underlying state. Since the agent cannot observe the degradation mechanism $\mathcal{D}$, conservative exploration maintains higher probability mass on actions that perform reasonably well across the belief uncertainty, reducing the exploration risk $\mathcal{R}(a|o)$ from Definition 3. This applies across all degradation types: information removal, noise corruption, flickering, and sensor dropout.

\textbf{Corollary 2 (Policy Update Conservatism).} Following from Theorem 3, conservative policy updates implement the conservative exploration strategy by constraining policy changes. Specifically, PPO's clipping mechanism \cite{schulman2017proximal} limits policy deviations, which bounds the exploration risk $\mathcal{R}(a|o)$ by preventing large action distribution changes that could be suboptimal under degraded observations.

\textbf{Proposition 2 (Algorithm-Specific Exploration Risk).} Following Theorem 3 and Corollary 2, the three algorithms exhibit different exploration risk $\mathcal{R}(a|o)$ profiles:
1. \textbf{PPO}: Conservative clipping $\Rightarrow$ low $\mathcal{R}(a|o)$;
2. \textbf{SAC}: Entropy maximization $\Rightarrow$ high $\mathcal{R}(a|o)$; 
3. \textbf{TD3}: Gaussian noise $\Rightarrow$ high $\mathcal{R}(a|o)$.

This directly instantiates Theorem 3's conservative vs. aggressive exploration prediction under partial observability. The proposition establishes the theoretical foundation for the empirical performance differences observed between these algorithms in POMDP settings.

\subsubsection{Experimental Hypotheses for Exploration Strategies}
\label{subsec:hypotheses_exploration_strategies}

Based on the theoretical analysis of exploration under partial observability, we formulate testable hypotheses to validate the conservative exploration advantage empirically.

In terms of \textbf{Diff \ref{diff_exploration}}, PPO updates its policy conservatively, controlled by hyperparameter $\epsilon$ in \eqnref{eq:ppo_policy_loss}, where smaller values of $\epsilon$ result in more conservative policy updates. SAC optimizes its policy to encourage exploration by maximizing policy entropy, controlled by hyperparameter $\alpha$ as shown in \eqnref{eq:sac_policy_loss}, where higher values of $\alpha$ lead to more exploratory policies. TD3 uses a deterministic policy optimized to maximize the estimated Q-value as in \eqnref{eq:td3_policy_loss}. For exploration, TD3 adds action noise to its policy following $a=\text{clip}(\mu_{\phi}(o)+\epsilon, a_{\text{low}}, a_{\text{high}})$, where the exploratory action is clipped to the value range $[a_{\text{low}}, a_{\text{high}}]$, $\epsilon \sim \mathbb{N}(0,\sigma)$, and $\sigma$ controls the degree of exploration.

This theoretical framework predicts that PPO's conservative exploration should minimize the exploration risk $\mathcal{R}(a|o)$ under all degradation mechanisms $\mathcal{D}$ (information removal, noise corruption, flickering, sensor dropout), while SAC's entropy maximization and TD3's action noise should increase exploration risk when observability is degraded.

\begin{hypothesis}
\label{hypothesis_exploration_strategy}
If the less exploratory policy of PPO leads to robustness in POMDP settings, then (1) reducing the exploration in TD3 and SAC should also make these algorithms more robust to POMDPs, while (2) increasing the exploration of PPO should make it less effective in POMDPs.
\end{hypothesis}
To validate \textbf{Hypothesis \ref{hypothesis_exploration_strategy}}, we adjust the action noise hyperparameter $\sigma$ of TD3 and the entropy coefficient hyperparameter $\alpha$ of SAC to examine how this affects their performance when moving from MDP to POMDP. Specifically, larger values of action noise $\sigma$ or entropy coefficient $\alpha$ result in more exploration. For PPO, we increase the clip ratio $\epsilon$ in \eqnref{eq:ppo_policy_loss} to allow the policy to update further from the current policy. As shown in \eqnref{eq:ppo_policy_loss}, if $\epsilon=0$, the policy is not permitted to change at all. To investigate various exploration levels, we evaluate a wide range of choices for these exploration-related hyperparameters, which are elaborated in the experiments.

\subsubsection{Theoretical Integration and Practical Considerations}
\label{subsubsec:Theoretical_Integration_and_Practical_Considerations}

Having established the theoretical foundations for both multi-step bootstrapping (Theorems 1-2, Definitions 1-2) and exploration under partial observability (Theorem 3, Definition 3, Proposition 2), we now synthesize these complementary mechanisms into a unified framework explaining the \textbf{Unexpected Result}.

\textbf{Synergistic Mechanism.} The two theoretical components work synergistically rather than independently. Multi-step bootstrapping (Theorem 1) enhances state information recovery through temporal aggregation, while conservative exploration (Theorem 3) prevents information loss through cautious action selection. In PPO, these mechanisms reinforce each other bidirectionally: multi-step targets from GAE provide better value estimates even under partial observability, enabling the clipping mechanism (Corollary 2) to make more informed conservative updates. Simultaneously, conservative policy updates reduce the exploration risk $\mathcal{R}(a|o)$, creating a stable learning environment that allows multi-step information integration to operate more effectively. In contrast, TD3 and SAC lack this synergy: both algorithms use single-step targets that cannot leverage temporal information recovery, while their aggressive exploration strategies (entropy maximization for SAC, Gaussian noise for TD3) actively increase exploration risk $\mathcal{R}(a|o)$, creating an unstable learning environment that compounds the information loss from single-step bootstrapping.

\textbf{Performance Inversion Prediction.} This synergy predicts performance inversion when partial observability exceeds a critical threshold. Under full observability, TD3 and SAC's aggressive exploration and single-step efficiency yield the expected ordering: $\mathbb{E}[J(\text{SAC})] \ge \mathbb{E}[J(\text{TD3})] \ge \mathbb{E}[J(\text{PPO})]$ \cite{fujimoto2018addressing,haarnoja2018soft}. Let $\Delta_{alg}(\mathcal{G}) = J_{MDP} - J_{POMDP}(\mathcal{G})$ denote the performance degradation from MDP to POMDP as a function of observability gap $\mathcal{G}$. As $\mathcal{G}$ increases, PPO's dual advantages—temporal integration and conservative updates—yield lower performance sensitivity: $\partial \Delta_{\text{PPO}}/\partial \mathcal{G} < \partial \Delta_{\text{SAC}}/\partial \mathcal{G}$ and $\partial \Delta_{\text{PPO}}/\partial \mathcal{G} < \partial \Delta_{\text{TD3}}/\partial \mathcal{G}$. At a critical threshold $\mathcal{G}^*$, the ordering inverts, providing a practical diagnostic for latent partial observability in continuous control tasks.

\textbf{Scope and Limitations.} While the integrated framework provides strong theoretical predictions, several limitations constrain its applicability: (i) \textbf{Dense reward assumption}: Multi-step information recovery requires informative rewards; sparse reward settings may not benefit from temporal aggregation; (ii) \textbf{Moderate stochasticity}: Highly stochastic transitions weaken reward correlations, reducing variance benefits; (iii) \textbf{Horizon trade-offs}: Excessively large multi-step horizons may inflate bias terms through two mechanisms: discount factor decay ($\gamma^n \to 0$) and compounding errors from bootstrapped Q-values, as consecutive rewards may be suboptimal when the policy is still learning. This can potentially reverse the variance reduction benefits of multi-step methods \cite{de2018multi}; (iv) \textbf{Memory alternatives}: Recurrent architectures with explicit memory may dominate multi-step benefits, altering the optimal balance \cite{hausknecht2015deep}. These constraints suggest the framework applies most directly to moderately stochastic continuous control with dense rewards and limited explicit memory.
\color{black}

\section{Experimental Setup}
\label{sec:experimental_setup}

\subsection{Experimental Design}
\label{sec:experimental_design}

We conduct comprehensive empirical validation of the theoretical predictions from Section \ref{sec:Analysing_and_Improving_Robustness_to_Partial_Observability}, systematically testing our hypotheses about multi-step bootstrapping and exploration strategies under partial observability.

\textbf{Scope and Rationale:} This work focuses on continuous control tasks, as they provide smooth value functions that maximize multi-step bootstrapping benefits and realistic sensor degradation scenarios that naturally motivate our POMDP variants.

\textbf{Research Questions:} Our experiments test four targeted hypotheses:
\begin{enumerate}
\item \textbf{Cross-Task Generalization (\textbf{Hypothesis \ref{hypothesis_generalization}})}: Does PPO's robustness advantage generalize across diverse continuous control tasks and observability degradation types?

\item \textbf{Multi-step Bootstrapping Efficacy (\textbf{Hypothesis \ref{hypothesis_compare_n_step_TD3_and_SAC_to_vanilla}})}: Do multi-step variants MTD3 and MSAC improve robustness compared to their single-step counterparts?

\item \textbf{Temporal vs Reward Aggregation (\textbf{Hypothesis \ref{hypothesis_compre_accumulated_reward_task_to_one_step_reward_task}})}: Does reward accumulation capture similar benefits to multi-step bootstrapping for temporal information integration?

\item \textbf{Exploration Strategy Impact (\textbf{Hypothesis \ref{hypothesis_exploration_strategy}})}: Do conservative exploration strategies improve POMDP robustness while aggressive exploration degrades performance?
\end{enumerate}

\textbf{Factorial Design:} We employ a $4 \times 5$ factorial design crossing 4 benchmark tasks (Ant, HalfCheetah, Hopper, Walker2D) with 5 observability conditions (1 MDP + 4 POMDP variants), systematically evaluating 6 algorithm variants (PPO, TD3, SAC, MTD3, MSAC, LSTM-TD3) per configuration, yielding 120 total experimental conditions.

\subsection{Benchmark Tasks}
\label{sec:benchmark_tasks}

We evaluate on four MuJoCo continuous control tasks providing diverse locomotion challenges with varying complexity and dimensionality, ensuring broad generalization across robotic control scenarios.

\begin{figure}[htp!]
    \centering
     \begin{subfigure}[b]{0.23\linewidth}
         \centering
         \includegraphics[width=2cm, height=2cm]{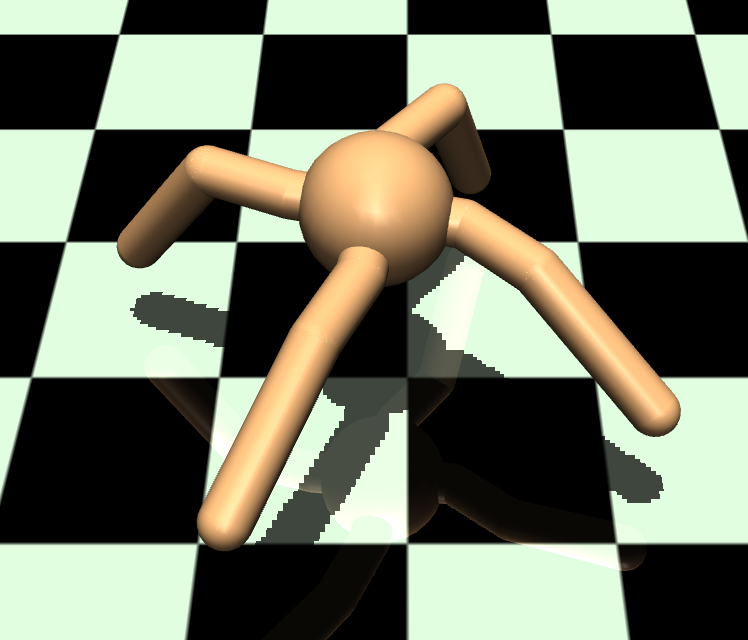}
         \caption{Ant-v2}
         \label{fig:task_Ant-v2}
     \end{subfigure}\hfill
     \begin{subfigure}[b]{0.23\linewidth}
         \centering
         \includegraphics[width=2cm, height=2cm]{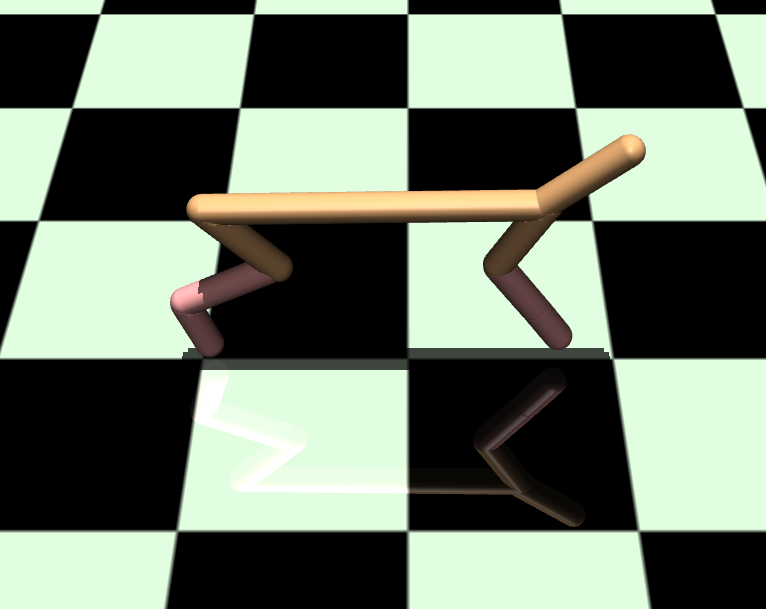}
         \caption{HalfCheetah-v2}
         \label{fig:task_HalfCheetah-v2}
     \end{subfigure}\hfill
     \begin{subfigure}[b]{0.23\linewidth}
         \centering
         \includegraphics[width=2cm, height=2cm]{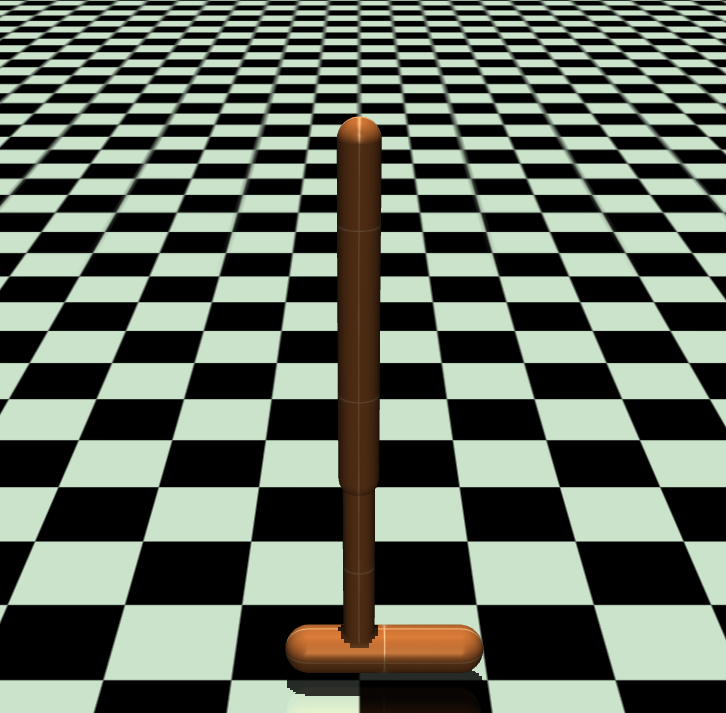}
         \caption{Hopper-v2}
         \label{fig:task_Hopper-v2}
     \end{subfigure}\hfill
      \begin{subfigure}[b]{0.23\linewidth}
         \centering
         \includegraphics[width=2cm, height=2cm]{figures/toy_tasks/Walker2d-v2.jpg}
         \caption{Walker2D-v2}
         \label{fig:task_Walker2d-v2}
     \end{subfigure}
    \caption{Benchmark continuous control tasks used for evaluating multi-step methods under partial observability.}
    \label{fig:Toy_MuJoCo_tasks_from_Gymnasium}
\end{figure}

\begin{table}[htp!]
    \centering
    \caption{Benchmark Task Specifications}
    \label{tab:benchmark_task_specs}
    \small
    \begin{tabular}{lcccc}
        \toprule
         & \textbf{Ant} & \textbf{HalfCheetah} & \textbf{Hopper} & \textbf{Walker2D} \\
        \midrule
        \textbf{Obs. Dim} & 29 & 17 & 11 & 17 \\
        \textbf{Act. Dim} & 8 & 6 & 3 & 6 \\
        \bottomrule
    \end{tabular}
\end{table}

\subsection{Partial Observability Variants}
\label{sec:pomdp_variants}

We construct four POMDP variants representing realistic failure modes in robotic systems:

\textbf{POMDP-RV (Remove Velocity):} Removes all velocity entries from observations, simulating encoder failures or IMU drift common in robotic locomotion. This challenges algorithms to reconstruct motion dynamics through temporal integration of position changes, directly testing multi-step bootstrapping's ability to recover missing kinematic information through history aggregation.

\textbf{POMDP-FLK (Flickering):} Zeros entire observations with probability $p_{flk} = 0.2$, modeling communication dropouts and sensor blackouts in wireless robotic networks. This tests robustness to severe temporal correlation disruption, validating whether algorithms can maintain policy continuity during complete information gaps through memory mechanisms.

\textbf{POMDP-RN (Random Noise):} Adds Gaussian noise $\epsilon \sim \mathcal{N}(0, \sigma_{rn}^2)$ with $\sigma_{rn} = 0.1$ to all entries, modeling ubiquitous sensor measurement errors. Unlike complete information loss, this creates continuous signal degradation that tests algorithmic noise filtering capabilities and robustness to measurement uncertainty through value function approximation.

\textbf{POMDP-RSM (Random Sensor Missing):} Zeros individual entries with probability $p_{rsm} = 0.1$, simulating gradual component failures affecting sensor subsets independently. This creates structured partial information loss that tests graceful degradation mechanisms and the ability to compensate using remaining reliable sensors under heterogeneous hardware reliability.

\begin{table}[htp!]
    \captionof{table}{Partial Observability Transformations}
    \small
    \centering
    \begin{tabular}{lll}
        \toprule
        \textbf{Variant}  & \textbf{Description}     & \textbf{Parameters} \\
        \midrule
        MDP            & Original fully observable task & $-$     \\\hline
        POMDP-RV       & \makecell[l]{Remove velocity entries from observation} & $-$      \\\hline
        POMDP-FLK      & \makecell[l]{Zero entire observation with probability $p_{flk}$} & $p_{flk}=0.2$  \\\hline
        POMDP-RN       & \makecell[l]{Add Gaussian noise $\epsilon \sim \mathcal{N}(0, \sigma_{rn}^2)$}   &  $\sigma_{rn}=0.1$            \\\hline
        POMDP-RSM      & \makecell[l]{Zero individual entries with probability $p_{rsm}$}     & $p_{rsm}=0.1$              \\
        \bottomrule
        \multicolumn{3}{p{12cm}}{\footnotesize{Note: All variants preserve original reward functions, isolating observation degradation effects from task modification.}}
    \end{tabular}
    \label{tab:MDP_and_POMDP_version_of_Tasks}
\end{table}

\subsection{Algorithm Implementation}
\label{sec:algorithm_implementation}

\textbf{Core Algorithms:} PPO (policy gradient with multi-step advantage estimation), TD3 (deterministic policy gradient with 1-step Q-learning), SAC (maximum entropy with 1-step Q-learning).

\textbf{Multi-step Variants:} MTD3($n$) and MSAC($n$) implement $n$-step bootstrapping with horizon sweep $n \in \{1,2,3,4,5\}$ to test temporal information recovery mechanisms. PPO with 1-step bootstrapping (replacing $\lambda$-return and Monte-Carlo return with 1-step returns) tests whether multi-step mechanisms are essential for PPO's robustness.

\textbf{Accumulated Reward Variants:} Algorithms tested on environments with accumulated rewards $r_t^{avg(n)}$ (average) and $r_t^{sum(n)}$ (summation) over $n$ consecutive time steps to validate whether reward aggregation can replicate multi-step bootstrapping benefits without temporal credit assignment changes.

\textbf{Exploration Strategy Variants:} TD3 with varied action noise $\sigma$, SAC with varied entropy coefficient $\alpha$, and PPO with varied clip ratio $\epsilon$ to test exploration impact on POMDP robustness as predicted by our theoretical analysis.

\textbf{Recurrent Baselines:} LSTM-TD3 \cite{meng2021memory} provides comparison with explicit memory mechanisms for distinguishing multi-step temporal integration from recurrent architectures. We focus solely on LSTM-TD3 rather than recurrent variants of PPO and SAC for three key reasons: (1) \textit{Research focus}: Our primary objective is investigating multi-step bootstrapping effects, not comprehensive recurrent architecture evaluation; (2) \textit{Baseline sufficiency}: LSTM-TD3 serves as a representative recurrent baseline to demonstrate whether simple temporal aggregation (multi-step) can match explicit memory mechanisms; (3) \textit{Methodological clarity}: PPO already incorporates multi-step mechanisms through $\lambda$-returns, making LSTM-PPO comparisons confound recurrent memory with multi-step effects, while SAC's entropy-regularized framework would introduce additional complexity orthogonal to our core investigation.

\textbf{Implementation Details:} All algorithms use identical network architectures (256×256 hidden layers), learning rates, and replay buffer sizes following OpenAI Spinning Up standards, ensuring differences stem purely from bootstrapping depth and exploration strategies. However, there is an exception that LSTM-TD3 follows the neural network structure specified in \url{https://github.com/LinghengMeng/LSTM-TD3} to maintain consistency with its original implementation and enable fair comparison with established recurrent baselines. These hyperparameters are chosen for several principled reasons: (1) \textit{Network architecture}: 256×256 hidden layers provide sufficient representational capacity for continuous control without overfitting—validated across MuJoCo benchmarks and consistent with established baselines \cite{fujimoto2018addressing,haarnoja2018soft}; (2) \textit{Standardization rationale}: OpenAI Spinning Up hyperparameters represent community-validated configurations that balance stability and performance across diverse continuous control tasks; (3) \textit{Controlled comparison}: Identical base configurations isolate the effects of multi-step bootstrapping and exploration strategies, eliminating confounding factors from hyperparameter variations; (4) \textit{Reproducibility}: Standard configurations enable direct comparison with existing literature and facilitate replication.

\subsection{Experimental Protocol}
\label{sec:experimental_protocol}

\textbf{Experimental Conditions:} Three complementary experimental sets validate our hypotheses:
\begin{enumerate}
    \item \textbf{Multi-step Bootstrapping}: Standard vs multi-step algorithms across all 20 task-observability configurations (4 tasks × 5 observability conditions)
    \item \textbf{Accumulated Reward}: Algorithms tested on original reward vs accumulated reward variants ($r_t^{avg(n)}$, $r_t^{sum(n)}$) across 16 POMDP conditions
    \item \textbf{Exploration Strategy}: Hyperparameter sweeps for exploration parameters (TD3 action noise $\sigma$, SAC entropy $\alpha$, PPO clip ratio $\epsilon$) on selected POMDP variants
\end{enumerate}

\textbf{Evaluation Metrics:}
\begin{itemize}
    \item \textbf{Primary Performance}: Maximum average return over 5 evaluation episodes within 2.5 million training steps based on 3 random seeds
    \item \textbf{Robustness}: Performance degradation ratio $(R_{MDP} - R_{POMDP})/R_{MDP}$ measuring observability sensitivity
    \item \textbf{Statistical Significance}: Two-sided t-tests with $p<0.05$ threshold for hypothesis validation, format $(t(df)=t\text{-value}, p=p\text{-value})$
    \item \textbf{Coverage Analysis}: Observation and action space coverage using dimensionality reduction for policy comparison
\end{itemize}

\textbf{Statistical Analysis:} Results aggregate over three random seeds with 95\% confidence intervals. Performance comparisons use proportion-based analysis across task sets. Learning curves use 1-D Gaussian smoothing ($\sigma=20$) for visualization while preserving raw data for analysis.

\textbf{Reproducibility:} Complete implementations and hyperparameters available at \url{https://github.com/LinghengMeng/m_rl_pomdp} with detailed protocols at \url{https://arxiv.org/abs/2209.04999}. 

\color{origColor}

\section{Experimental Results}
\label{sec:experimental_results}

This section presents experimental validation of our theoretical framework across multiple continuous control tasks and partial observability conditions. We systematically evaluate algorithm performance, multi-step bootstrapping effectiveness, accumulated reward variants, and exploration strategy impacts on POMDP robustness.

\subsection{Results on the Generalization of the Performance Inversion} on Other Tasks
\label{subsec:Results_on_the_Generalization_of_the_Unexpected_Result_on_Other_Tasks}

\begin{figure}[htp!]
    \centering
    \includegraphics[width=.95\linewidth]{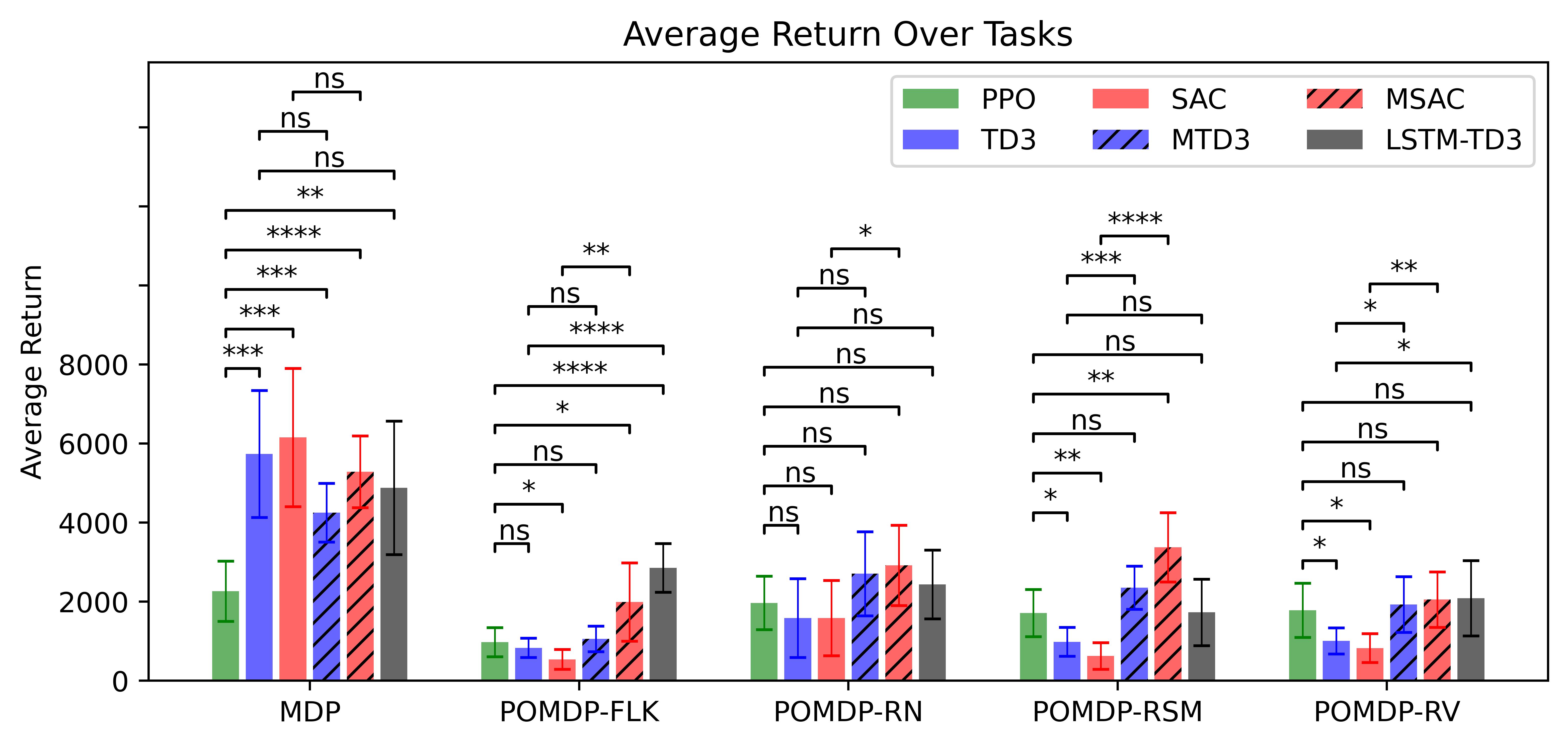}
    \caption{Bar Chart of Average Return Over Tasks in Table \ref{tab:Maximum_of_Average_Return}. The error bars indicate the 95\% confidence intervals. Two-sample two-sided T-Test is conducted between two algorithms under the end of each bracket. The labels above the bracket are as follows: ns: $p>0.05$, $\ast$: $p\leq0.05$, $\ast\ast$: $p\leq0.01$, $\ast\ast\ast$: $p\leq0.001$, $\ast\ast\ast\ast$: $p\leq0.0001$.}
    \label{fig:Average_Return_Over_Tasks}
\end{figure}

To address \textbf{Research Questions 3 and 4}, investigate the generalization of the performance inversion introduced in the exemplar robot control problem in Section \ref{sec:Exemplar_Robot_Control_Problem} and validate the \textbf{Hypothesis \ref{hypothesis_generalization}} made in in Section \ref{sec:Generalization_of_the_Unexpected_Result_on_Other_Tasks}, we run PPO, TD3 and SAC on both the MDP- and POMDP-versions of Ant-v2, HalfCheetah-v2, Hopper-v2, and Walker2D-v2. The results on each task are shown in Table \ref{tab:Maximum_of_Average_Return}
, and for better visualization, the average return over tasks listed in Table \ref{tab:Maximum_of_Average_Return} are summarized in Fig. \ref{fig:Average_Return_Over_Tasks}. For all comparisons, two-sided T-tests are conducted and the corresponding results are included in Table \ref{tab:Maximum_of_Average_Return} and Fig. \ref{fig:Average_Return_Over_Tasks}. It can be seen from Table \ref{tab:Maximum_of_Average_Return} that on most MDP-versions of the 4 tasks, TD3 and SAC achieve significantly better performance than PPO. However, on the 4 POMDP-versions of the 4 tasks, PPO outperforms TD3 and SAC on 11/16 tasks as highlighted in gray in Table \ref{tab:Maximum_of_Average_Return}, where most cases show significant statistical difference. This directly answers \textbf{Research Question 3} by demonstrating that conservative exploration strategies (PPO's clipped policy updates) consistently outperform aggressive exploration (SAC's entropy-driven and TD3's Gaussian noise exploration) under partial observability conditions. In addition, when moving from the MDP to the POMDP, PPO experiences mild performance decreases for most cases with two exceptions, i.e., Ant POMDP-RV and Hopper POMDP-RSM, where PPO even has improved performance as highlighted by $\bigstar$ in Table \ref{tab:Maximum_of_Average_Return}. While in case of Cheetah the difference is small and it could be argued that it is purely statistical noise and is thus insignificant, in Ant the difference is quite large. Following the hypothesis that PPO could incorporate temporal information, a potential explanation is that velocity is indirectly incorporated and probably additionally adding it to the observation can only make the observation space larger and introduce difficulty in finding an optimal policy.  However, remarkably TD3 and SAC all encounter significant performance decrease with no exception. It is even more clear as summarized in Fig. \ref{fig:Average_Return_Over_Tasks} that on MDP, PPO significantly under-performs TD3 and SAC and their corresponding variants, while at the same time there is no significant difference between vanilla TD3 or SAC and their variants MTD3/LSTM-TD3 or MSAC. 

To summarize, these findings are consistent with  \textbf{Hypothesis \ref{hypothesis_generalization}}, confirming the generalization of the add{performance inversion} on other tasks and directly answering \textbf{Research Question 4} by demonstrating that robustness patterns hold consistently across different robotic control tasks with varying state-action complexities (17-29 state dimensions, 3-8 action dimensions). This alerts researchers that applying state-of-the-art DRL algorithms directly to a task with partial observation may be problematic.


\setlength{\tabcolsep}{4pt}
\setlength{\extrarowheight}{0pt}
{\scriptsize
\begin{xltabular}{\linewidth}{ll|l|r|c|c|c|c|c|c}
    \caption{The Maximum of Average Return over 5 evaluation episodes within 2.5 million steps based on 3 random seeds. If TD3 or SAC perform worse than PPO, they are gray-colored. If MTD3(5) or MSAC(5) outperform the corresponding TD3 or SAC, they are red-colored. The maximum value of all evaluated algorithms for each task is in boldface. For each row, if an algorithm's performance is better than PPO, it is indicated by \cmark. Otherwise, it is indicated by \xmark. $\bigstar$ indicates the performance of an RL algorithm on the POMDP-version of a task outperforms its performance on the corresponding MDP-version of the task. Two-sided T-test results in format $(t(df)=t-value, p=p-value$ are included under average return with all significantly different cases highlighted in green. Specifically, ``PPO vs" indicates the T-test between the PPO and the agent corresponding to the column, where PPO is not compared with itself as indicated with $-$. ``Orig vs" indicates the T-test between the variant and vanilla version of the TD3 or SAC, e.g., MTD3 vs TD3, MSAC vs SAC, or LSTM-TD3 vs TD3, where the not applicable comparisons are indicated with $-$. }\label{tab:Maximum_of_Average_Return}\\
    \hline
    \multicolumn{4}{c|}{\textbf{Task}} & \multicolumn{6}{c}{\textbf{Algorithms}} \\\hline
    & \multicolumn{3}{c|}{Version}   & \textbf{PPO} & \textbf{TD3} & \textbf{SAC} & \textbf{MTD3(5)} & \textbf{MSAC(5)} & \makecell{\textbf{LSTM-}\\\textbf{TD3(5)}} \\ \hline
    \endfirsthead

    \multicolumn{10}{l}{Continuation of Table \ref{tab:Maximum_of_Average_Return}}\\
    \hline
    \multicolumn{4}{c|}{\textbf{Task}} & \multicolumn{6}{c}{\textbf{Algorithms}} \\\hline
    & \multicolumn{3}{c|}{Version}   & \textbf{PPO} & \textbf{TD3} & \textbf{SAC} & \textbf{MTD3(5)} & \textbf{MSAC(5)} & \makecell{\textbf{LSTM-}\\\textbf{TD3(5)}} \\ \hline
    \endhead
    
    \multicolumn{10}{r}{{Continued on next page}}
    \endfoot
    
    \endlastfoot
    
    \multirow{15}{*}{\rotatebox[origin=c]{90}{Ant}}   & \multirow{3}{*}{\rotatebox[origin=c]{90}{MDP}}  & \multicolumn{2}{r|}{\scalebox{.75}{Return}}      &  476.99   &  \cmark4773.70  &  \cmark\textbf{5614.10}      &    \cmark3236.93     &    \cmark5000.62     &   \cmark4070.03          \\\cdashline{3-10}[1pt/1pt]
    & & \multirow{2}{*}{\rotatebox[origin=c]{90}{t-test}}  & \scalebox{.75}{\makecell{PPO vs}}     &  $-$ &\cellcolor{green!25}{\scalebox{.75}{\makecell{(t(4)=-4.91; \\p=8.0e-03)}}} & \cellcolor{green!25}{\scalebox{.75}{\makecell{(t(4)=-26.87;\\ p=1.1e-05)}}}& \cellcolor{green!25}{\scalebox{.75}{\makecell{(t(4)=-10.94;\\p=4.0e-04)}}} & \cellcolor{green!25}{\scalebox{.75}{\makecell{(t(4)=-14.78;\\p=1.2e-04)}}} & \cellcolor{green!25}{\scalebox{.75}{\makecell{(t(4)=-7.85;\\p=1.4e-03)}}}\\\cdashline{4-10}[1pt/1pt]
    & & & \scalebox{.75}{\makecell{Orig vs}} & $-$  & $-$ & $-$ &\scalebox{.75}{\makecell{(t(4)=1.69;\\p=1.7e-01)}} & \scalebox{.75}{\makecell{(t(4)=1.74;\\p=1.6e-01)}}  & \scalebox{.75}{\makecell{(t(4)=0.71;\\p=5.1e-01)}} \\\cline{2-10}
            & \multirow{3}{*}{\rotatebox[origin=c]{90}{P-FLK}} & \multicolumn{2}{r|}{\scalebox{.75}{Return}} & 396.67   &  \cmark1087.10   &   \cmark936.85   &    \cellcolor{red!25}{\cmark1693.12}     &     \cellcolor{red!25}{\textbf{\cmark2922.37}}    &    \cmark2658.81         \\\cdashline{3-10}[1pt/1pt]
    & & \multirow{2}{*}{\rotatebox[origin=c]{90}{t-test}}& \scalebox{.75}{\makecell{PPO vs}}  & $-$ & \cellcolor{green!25}{\scalebox{.75}{\makecell{(t(4)=-9.85;\\p=6.0e-04)}}} & \cellcolor{green!25}{\scalebox{.75}{\makecell{(t(4)=-24.53;\\p=1.6e-05)}}}& \cellcolor{green!25}{\scalebox{.75}{\makecell{(t(4)=-6.04;\\p=3.8e-03)}}} & \cellcolor{green!25}{\scalebox{.75}{\makecell{(t(4)=-4.21;\\p=1.4e-02)}}} & \cellcolor{green!25}{\scalebox{.75}{\makecell{(t(4)=-7.18;\\p=2.0e-03)}}} \\\cdashline{4-10}[1pt/1pt]
    & & & \scalebox{.75}{\makecell{Orig vs}} & $-$  & $-$ & $-$ &\scalebox{.75}{\makecell{(t(4)=-2.69;\\p=5.5e-02)}} & \cellcolor{green!25}{\scalebox{.75}{\makecell{(t(4)=-3.30;\\p=3.0e-02)}}} & \cellcolor{green!25}{\scalebox{.75}{\makecell{(t(4)=-4.87;\\p=8.2e-03)}}} \\\cline{2-10}
                                                     & \multirow{3}{*}{\rotatebox[origin=c]{90}{P-RN}} & \multicolumn{2}{r|}{\scalebox{.75}{Return}} &  331.50    &   \cmark1051.60  &  \cmark1150.13   &    \cellcolor{red!25}{\cmark1078.68}     &     \cellcolor{red!25}{\textbf{\cmark2679.30}}    &   \cmark1095.33         \\\cdashline{3-10}[1pt/1pt]
    & & \multirow{2}{*}{\rotatebox[origin=c]{90}{t-test}} & \scalebox{.75}{\makecell{PPO vs}} & $-$ &\scalebox{.75}{\makecell{(t(4)=-2.63;\\p=5.8e-02)}} & \cellcolor{green!25}{\scalebox{.75}{\makecell{(t(4)=-5.48;\\p=5.4e-03)}}}& \cellcolor{green!25}{\scalebox{.75}{\makecell{(t(4)=-4.54;\\p=1.0e-02)}}} & \cellcolor{green!25}{\scalebox{.75}{\makecell{(t(4)=-10.48;\\p=4.7e-04)}}} & \cellcolor{green!25}{\scalebox{.75}{\makecell{(t(4)=-3.86;\\p=1.8e-02)}}} \\\cdashline{4-10}[1pt/1pt]
    & & & \scalebox{.75}{\makecell{Orig vs}} & $-$ & $-$ & $-$ &\scalebox{.75}{\makecell{(t(4)=-0.09;\\p=9.3e-01)}} & \cellcolor{green!25}{\scalebox{.75}{\makecell{(t(4)=-6.28;\\p=3.3e-03)}}} & \scalebox{.75}{\makecell{(t(4)=-0.14;\\p=9.0e-01)}} \\\cline{2-10}
                                                     & \multirow[c]{3}{*}{\rotatebox[origin=c]{90}{P-RSM}} & \multicolumn{2}{r|}{\scalebox{.75}{Return}} &  383.02    &   \cmark1045.62  &  \cmark885.80   &     \cellcolor{red!25}{\cmark2048.93}    &      \cellcolor{red!25}{\textbf{\cmark3450.29}}   &     \cmark1164.97        \\\cdashline{3-10}[1pt/1pt]
    & & \multirow{2}{*}{\rotatebox[origin=c]{90}{t-test}} & \scalebox{.75}{\makecell{PPO vs}} & $-$ &\cellcolor{green!25}{\scalebox{.75}{\makecell{(t(4)=-6.85;\\p=2.4e-03)}}} & \cellcolor{green!25}{\scalebox{.75}{\makecell{(t(4)=-7.28;\\p=1.9e-03)}}}& \cellcolor{green!25}{\scalebox{.75}{\makecell{(t(4)=-5.66;\\p=4.8e-03)}}} & \cellcolor{green!25}{\scalebox{.75}{\makecell{(t(4)=-5.19;\\p=6.6e-03)}}} & \cellcolor{green!25}{\scalebox{.75}{\makecell{(t(4)=-4.69;\\p=9.4e-03)}}} \\\cdashline{4-10}[1pt/1pt]
    & & & \scalebox{.75}{\makecell{Orig vs}} &  $-$ & $-$ & $-$ &\cellcolor{green!25}{\scalebox{.75}{\makecell{(t(4)=-3.39;\\p=2.8e-02)}}} & \cellcolor{green!25}{\scalebox{.75}{\makecell{(t(4)=-4.36;\\p=1.2e-02)}}} & \scalebox{.75}{\makecell{(t(4)=-0.70;\\p=5.2e-01)}} \\\cline{2-10}
                                                     & \multirow[c]{3}{*}{\rotatebox[origin=c]{90}{P-RV}} & \multicolumn{2}{r|}{\scalebox{.75}{Return}} &  $\bigstar$1997.48   &  \cellcolor{gray!25}\xmark1321.32   &  \cellcolor{gray!25}\xmark949.46   &   \cellcolor{red!25}{\cmark2557.28}      &    \cellcolor{red!25}{\cmark\textbf{2949.87}}     &   \xmark1325.87          \\\cdashline{3-10}[1pt/1pt] 
     & & \multirow{2}{*}{\rotatebox[origin=c]{90}{t-test}} & \scalebox{.75}{\makecell{PPO vs}} & $-$ &\scalebox{.75}{\makecell{(t(4)=2.11;\\p=1.0e-01)}} & \cellcolor{green!25}{\scalebox{.75}{\makecell{(t(4)=4.79;\\p=8.7e-03)}}}& \scalebox{.75}{\makecell{(t(4)=-1.89;\\p=1.3e-01)}} & \scalebox{.75}{\makecell{(t(4)=-2.02;\\p=1.1e-01)}} & \scalebox{.75}{\makecell{(t(4)=2.76;\\p=5.1e-02)}} \\\cdashline{4-10}[1pt/1pt]
     & & & \scalebox{.75}{\makecell{Orig vs}} &  $-$ & $-$ & $-$ &\cellcolor{green!25}{\scalebox{.75}{\makecell{(t(4)=-4.01;\\p=1.6e-02)}}} & \cellcolor{green!25}{\scalebox{.75}{\makecell{(t(4)=-4.78;\\p=8.8e-03)}}} & \scalebox{.75}{\makecell{(t(4)=-0.02;\\p=9.9e-01)}}\\\hline
     \multirow{15}{*}{\rotatebox[origin=c]{90}{HalfCheetah}}  & \multirow{3}{*}{\rotatebox[origin=c]{90}{MDP}} & \multicolumn{2}{r|}{\scalebox{.75}{Return}} &  2897.89   &  \cmark9587.33   &  \scalebox{.95}{\cmark\textbf{10419.59}}   &  \cmark5442.85      &   \cmark7270.04      &   \cmark8494.11          \\\cdashline{3-10}[1pt/1pt]
    & & \multirow{2}{*}{\rotatebox[origin=c]{90}{t-test}} & \scalebox{.75}{\makecell{PPO vs}} & $-$&\cellcolor{green!25}{\scalebox{.75}{\makecell{(t(4)=-9.22;\\p=7.7e-04)}}} & \cellcolor{green!25}{\scalebox{.75}{\makecell{(t(4)=-9.80;\\p=6.1e-04)}}}& \scalebox{.75}{\makecell{(t(4)=-2.75;\\p=5.2e-02)}} & \cellcolor{green!25}{\scalebox{.75}{\makecell{(t(4)=-8.19;\\p=1.2e-03)}}} & \cellcolor{green!25}{\scalebox{.75}{\makecell{(t(4)=-4.79;\\p=8.7e-03)}}} \\\cdashline{4-10}[1pt/1pt]
    & & & \scalebox{.75}{\makecell{Orig vs}} &  $-$ & $-$ & $-$ &\cellcolor{green!25}{\scalebox{.75}{\makecell{(t(4)=3.92;\\p=1.7e-02)}}} & \cellcolor{green!25}{\scalebox{.75}{\makecell{(t(4)=4.04;\\p=1.6e-02)}}} & \scalebox{.75}{\makecell{(t(4)=0.86;\\p=4.4e-01)}} \\\cline{2-10}
                                                 & \multirow{3}{*}{\rotatebox[origin=c]{90}{P-FLK}} & \multicolumn{2}{r|}{\scalebox{.75}{Return}}  &  1045.15   &  \cmark1141.83   & \cellcolor{gray!25}\xmark103.27    &   \cmark1118.23      &   \cellcolor{red!25}{\cmark\textbf{3397.47}}     &    \cmark1515.97         \\\cdashline{3-10}[1pt/1pt]
& & \multirow{2}{*}{\rotatebox[origin=c]{90}{t-test}} & \scalebox{.75}{\makecell{PPO vs}} & $-$ &\scalebox{.75}{\makecell{(t(4)=-0.30;\\p=7.8e-01)}} & \cellcolor{green!25}{\scalebox{.75}{\makecell{(t(4)=2.93;\\p=4.3e-02)}}} & \scalebox{.75}{\makecell{(t(4)=-0.22;\\p=8.3e-01)}} & \scalebox{.75}{\makecell{(t(4)=-2.11;\\p=1.0e-01)}} & \scalebox{.75}{\makecell{(t(4)=-1.46;\\p=2.2e-01)}} \\\cdashline{4-10}[1pt/1pt]
& & & \scalebox{.75}{\makecell{Orig vs}} & $-$ & $-$ & $-$ &\scalebox{.75}{\makecell{(t(4)=0.31;\\p=7.7e-01)}} & \cellcolor{green!25}{\scalebox{.75}{\makecell{(t(4)=-3.09;\\p=3.7e-02)}}} & \cellcolor{green!25}{\scalebox{.75}{\makecell{(t(4)=-6.09;\\p=3.7e-03)}}}\\\cline{2-10}
                                                 & \multirow{3}{*}{\rotatebox[origin=c]{90}{P-RN}} & \multicolumn{2}{r|}{\scalebox{.75}{Return}}   &  2876.18   &  \cmark4009.17   &  \cmark3950.56   & \cellcolor{red!25}{\cmark\textbf{4961.27}}       &  \cellcolor{red!25}{\cmark4931.28}      &   \cmark3108.72          \\\cdashline{3-10}[1pt/1pt]
& & \multirow{2}{*}{\rotatebox[origin=c]{90}{t-test}} & \scalebox{.75}{\makecell{PPO vs}} & $-$ &\scalebox{.75}{\makecell{(t(4)=-1.91;\\p=1.3e-01)}} & \scalebox{.75}{\makecell{(t(4)=-2.35;\\p=7.8e-02)}}& \cellcolor{green!25}{\scalebox{.75}{\makecell{(t(4)=-6.45;\\p=3.0e-03)}}} & \cellcolor{green!25}{\scalebox{.75}{\makecell{(t(4)=-4.98;\\p=7.6e-03)}}} & \scalebox{.75}{\makecell{(t(4)=-0.25;\\p=8.1e-01)}} \\\cdashline{4-10}[1pt/1pt]
& & & \scalebox{.75}{\makecell{Orig vs}} & $-$ & $-$ & $-$ &\scalebox{.75}{\makecell{(t(4)=-1.61;\\p=1.8e-01)}} & \scalebox{.75}{\makecell{(t(4)=-1.87;\\p=1.3e-01)}} & \scalebox{.75}{\makecell{(t(4)=0.86;\\p=4.4e-01)}} \\\cline{2-10}
                                                 & \multirow{3}{*}{\rotatebox[origin=c]{90}{P-RSM}} & \multicolumn{2}{r|}{\scalebox{.75}{Return}}  &  2162.40   &  \cellcolor{gray!25}\xmark1149.96   &  \cellcolor{gray!25}\xmark50.62   &  \cellcolor{red!25}{\xmark1892.19}      &   \cellcolor{red!25}{\cmark\textbf{4603.68}}     &    \xmark1331.18         \\\cdashline{3-10}[1pt/1pt]
& & \multirow{2}{*}{\rotatebox[origin=c]{90}{t-test}} & \scalebox{.75}{\makecell{PPO vs}} & $-$ &\cellcolor{green!25}{\scalebox{.75}{\makecell{(t(4)=5.98;\\p=3.9e-03)}}} & \cellcolor{green!25}{\scalebox{.75}{\makecell{(t(4)=12.37;\\p=2.5e-04)}}}& \scalebox{.75}{\makecell{(t(4)=0.37;\\p=7.3e-01)}} & \cellcolor{green!25}{\scalebox{.75}{\makecell{(t(4)=-3.24;\\p=3.2e-02)}}} &  \cellcolor{green!25}{\scalebox{.75}{\makecell{(t(4)=4.61;\\p=1.0e-02)}}} \\\cdashline{4-10}[1pt/1pt]
& & & \scalebox{.75}{\makecell{Orig vs}} & $-$  & $-$ & $-$ &\scalebox{.75}{\makecell{(t(4)=-1.05;\\p=3.5e-01)}} & \cellcolor{green!25}{\scalebox{.75}{\makecell{(t(4)=-6.20;\\p=3.4e-03)}}} & \scalebox{.75}{\makecell{(t(4)=-2.39;\\p=7.5e-02)}} \\\cline{2-10}
                                                 & \multirow{3}{*}{\rotatebox[origin=c]{90}{P-RV}} & \multicolumn{2}{r|}{\scalebox{.75}{Return}} &  2659.48   &  \cellcolor{gray!25}\xmark1424.24  &  \cellcolor{gray!25}\xmark157.48   &  \cellcolor{red!25}{\cmark2839.08}      &   \cellcolor{red!25}{\xmark2393.62}     &   \cmark\textbf{3432.32}         \\ \cdashline{3-10}[1pt/1pt]
& & \multirow{2}{*}{\rotatebox[origin=c]{90}{t-test}} & \scalebox{.75}{\makecell{PPO vs}} & $-$ &\scalebox{.75}{\makecell{(t(4)=1.40;\\p=2.4e-01)}} & \cellcolor{green!25}{\scalebox{.75}{\makecell{(t(4)=2.93;\\p=4.3e-02)}}}& \scalebox{.75}{\makecell{(t(4)=-0.19;\\p=8.6e-01)}} & \scalebox{.75}{\makecell{(t(4)=0.25;\\p=8.1e-01)}} & \scalebox{.75}{\makecell{(t(4)=-0.83;\\p=4.5e-01)}} \\\cdashline{4-10}[1pt/1pt]
 & & & \scalebox{.75}{\makecell{Orig vs}} & $-$  & $-$ & $-$ &\scalebox{.75}{\makecell{(t(4)=-2.52;\\p=6.6e-02)}} & \cellcolor{green!25}{\scalebox{.75}{\makecell{(t(4)=-3.33;\\p=2.9e-02)}}} & \cellcolor{green!25}{\scalebox{.75}{\makecell{(t(4)=-4.00;\\p=1.6e-02)}}} \\\hline
\multirow{15}{*}{\rotatebox[origin=c]{90}{Hopper}}       & \multirow{3}{*}{\rotatebox[origin=c]{90}{MDP}}   & \multicolumn{2}{r|}{\scalebox{.75}{Return}}     &  2602.18   &  \cmark\textbf{3674.87}   &  \cmark3575.86   &    \cmark3474.84     &   \cellcolor{red!25}{\cmark3588.95}     &  \xmark2353.90           \\\cdashline{3-10}[1pt/1pt]
& & \multirow{2}{*}{\rotatebox[origin=c]{90}{t-test}} & \scalebox{.75}{\makecell{PPO vs}} & $-$ &\cellcolor{green!25}{\scalebox{.75}{\makecell{(t(4)=-3.24;\\p=3.2e-02)}}} & \cellcolor{green!25}{\scalebox{.75}{\makecell{(t(4)=-2.95;\\p=4.2e-02)}}}& \scalebox{.75}{\makecell{(t(4)=-2.54;\\p=6.4e-02)}} & \cellcolor{green!25}{\scalebox{.75}{\makecell{(t(4)=-2.96;\\p=4.2e-02)}}} & \scalebox{.75}{\makecell{(t(4)=0.20;\\p=8.5e-01)}} \\\cdashline{4-10}[1pt/1pt]
& & & \scalebox{.75}{\makecell{Orig vs}} &  $-$ & $-$ & $-$ &\scalebox{.75}{\makecell{(t(4)=1.98;\\p=1.2e-01)}} & \scalebox{.75}{\makecell{(t(4)=-0.26;\\p=8.1e-01)}} & \scalebox{.75}{\makecell{(t(4)=1.12;\\p=3.3e-01)}} \\\cline{2-10}
                                                 & \multirow{3}{*}{\rotatebox[origin=c]{90}{P-FLK}} & \multicolumn{2}{r|}{\scalebox{.75}{Return}}   &  1204.44 & \cellcolor{gray!25}\xmark519.98   &  \cellcolor{gray!25}\xmark530.76   &   \cellcolor{red!25}{\xmark759.73}     &  \cellcolor{red!25}{\xmark804.94}      &  \scalebox{.85}{$ \bigstar$\cmark\textbf{3400.00}}           \\\cdashline{3-10}[1pt/1pt]
& & \multirow{2}{*}{\rotatebox[origin=c]{90}{t-test}} & \scalebox{.75}{\makecell{PPO vs}} & $-$ &\scalebox{.75}{\makecell{(t(4)=1.31;\\p=2.6e-01)}} & \scalebox{.75}{\makecell{(t(4)=1.27;\\p=2.7e-01)}}& \scalebox{.75}{\makecell{(t(4)=0.94;\\p=4.0e-01)}} & \scalebox{.75}{\makecell{(t(4)=0.78;\\p=4.8e-01)}} & \cellcolor{green!25}{\scalebox{.75}{\makecell{(t(4)=-4.66;\\p=9.6e-03)}}} \\\cdashline{4-10}[1pt/1pt]
& & & \scalebox{.75}{\makecell{Orig vs}} & $-$  & $-$ & $-$ &\scalebox{.75}{\makecell{(t(4)=-0.90;\\p=4.2e-01)}} & \scalebox{.75}{\makecell{(t(4)=-0.80;\\p=4.7e-01)}} & \cellcolor{green!25}{\scalebox{.75}{\makecell{(t(4)=-10.92;\\p=4.0e-04)}}}\\\cline{2-10}
                                                 & \multirow{3}{*}{\rotatebox[origin=c]{90}{P-RN}} & \multicolumn{2}{r|}{\scalebox{.75}{Return}}  &  \textbf{2360.31}   &  \cellcolor{gray!25}\xmark597.15   &  \cellcolor{gray!25}\xmark584.08   &   \cellcolor{red!25}{\xmark1579.21}     &  \cellcolor{red!25}{\xmark1569.69}      &  \cmark2912.29           \\\cdashline{3-10}[1pt/1pt]
& & \multirow{2}{*}{\rotatebox[origin=c]{90}{t-test}} & \scalebox{.75}{\makecell{PPO vs}} & $-$ & \cellcolor{green!25}{\scalebox{.75}{\makecell{(t(4)=4.30;\\p=1.3e-02)}}} & \cellcolor{green!25}{\scalebox{.75}{\makecell{(t(4)=4.70;\\p=9.3e-03)}}}& \scalebox{.75}{\makecell{(t(4)=1.15;\\p=3.1e-01)}} & \scalebox{.75}{\makecell{(t(4)=0.94;\\p=4.0e-01)}} & \scalebox{.75}{\makecell{(t(4)=-1.65;\\p=1.7e-01)}}\\\cdashline{4-10}[1pt/1pt]
& & & \scalebox{.75}{\makecell{Orig vs}} & $-$  & $-$ & $-$ &\scalebox{.75}{\makecell{(t(4)=-1.48;\\p=2.1e-01)}} & \scalebox{.75}{\makecell{(t(4)=-1.21;\\p=2.9e-01)}} & \cellcolor{green!25}{\scalebox{.75}{\makecell{(t(4)=-7.57;\\p=1.6e-03)}}}\\\cline{2-10}
                                                 & \multirow{3}{*}{\rotatebox[origin=c]{90}{P-RSM}} & \multicolumn{2}{r|}{\scalebox{.75}{Return}}  &  $\bigstar$\textbf{2733.00}   &  \cellcolor{gray!25}\xmark1057.91   &  \cellcolor{gray!25}\xmark648.71   &   \cellcolor{red!25}{\xmark2237.73}     &  \cellcolor{red!25}{\xmark2018.09}      &  \xmark585.72           \\\cdashline{3-10}[1pt/1pt]
& & \multirow{2}{*}{\rotatebox[origin=c]{90}{t-test}} & \scalebox{.75}{\makecell{PPO vs}} & $-$ &\scalebox{.75}{\makecell{(t(4)=2.29;\\p=8.4e-02)}} & \cellcolor{green!25}{\scalebox{.75}{\makecell{(t(4)=8.84;\\p=9.1e-04)}}}& \scalebox{.75}{\makecell{(t(4)=1.24;\\p=2.8e-01)}} & \scalebox{.75}{\makecell{(t(4)=0.98;\\p=3.8e-01)}} & \cellcolor{green!25}{\scalebox{.75}{\makecell{( t(4)=6.30;\\p=3.2e-03)}}} \\\cdashline{4-10}[1pt/1pt]
& & & \scalebox{.75}{\makecell{Orig vs}} & $-$  & $-$ & $-$ & \scalebox{.75}{\makecell{(t(4)=-1.44;\\p=2.2e-01)}} & \scalebox{.75}{\makecell{(t(4)=-1.81;\\p=1.4e-01)}} & \scalebox{.75}{\makecell{(t(4)=0.59;\\p=5.8e-01)}}\\\cline{2-10}
                                                 & \multirow{3}{*}{\rotatebox[origin=c]{90}{P-RV}} & \multicolumn{2}{r|}{\scalebox{.75}{Return}} &  848.52   &  \cellcolor{gray!25}\xmark532.99   &  \textbf{\cmark1029.32}   &    \xmark392.43     &  \xmark741.32      &   \xmark223.49          \\\cdashline{3-10}[1pt/1pt] 
& & \multirow{2}{*}{\rotatebox[origin=c]{90}{t-test}} & \scalebox{.75}{\makecell{PPO vs}} & $-$ &\scalebox{.75}{\makecell{(t(4)=0.82;\\p=4.6e-01)}} & \scalebox{.75}{\makecell{(t(4)=-0.60;\\p=5.8e-01)}}& \scalebox{.75}{\makecell{(t(4)=1.06;\\p=3.5e-01)}} & \scalebox{.75}{\makecell{(t(4)=0.27;\\p=8.0e-01)}} & \scalebox{.75}{\makecell{(t(4)=1.77;\\p=1.5e-01)}} \\\cdashline{4-10}[1pt/1pt]
& & & \scalebox{.75}{\makecell{Orig vs}} & $-$  & $-$ & $-$ & \scalebox{.75}{\makecell{(t(4)=0.36;\\p=7.4e-01)}} &  \scalebox{.75}{\makecell{(t(4)=1.08;\\p=3.4e-01)}} & \scalebox{.75}{\makecell{(t(4)=1.01;\\p=3.7e-01)}}\\\hline
\multirow{15}{*}{\rotatebox[origin=c]{90}{Walker2d}}     & \multirow{3}{*}{\rotatebox[origin=c]{90}{MDP}}   & \multicolumn{2}{r|}{\scalebox{.75}{Return}}     & 3067.91    &  \cmark4902.00   &  \cmark4988.70   &   \cmark4838.84     &   \cellcolor{red!25}{\cmark\textbf{5270.83}}      &   \cmark4591.12          \\\cdashline{3-10}[1pt/1pt]
& & \multirow{2}{*}{\rotatebox[origin=c]{90}{t-test}} & \scalebox{.75}{\makecell{PPO vs}} & $-$ &\cellcolor{green!25}{\scalebox{.75}{\makecell{(t(4)=-2.82;\\p=4.8e-02)}}} & \cellcolor{green!25}{\scalebox{.75}{\makecell{(t(4)=-3.35;\\p=2.9e-02)}}}& \cellcolor{green!25}{\scalebox{.75}{\makecell{(t(4)=-3.77;\\p=2.0e-02)}}} & \cellcolor{green!25}{\scalebox{.75}{\makecell{(t(4)=-4.47;\\p=1.1e-02)}}} & \cellcolor{green!25}{\scalebox{.75}{\makecell{(t(4)=-2.96;\\p=4.2e-02)}}}\\\cdashline{4-10}[1pt/1pt]
& & & \scalebox{.75}{\makecell{Orig vs}} & $-$  & $-$ & $-$ &\scalebox{.75}{\makecell{(t(4)=0.14;\\p=9.0e-01)}} & \scalebox{.75}{\makecell{(t(4)=-0.76;\\p=4.9e-01)}} & \scalebox{.75}{\makecell{(t(4)=0.62;\\p=5.7e-01)}} \\\cline{2-10}
                                                 & \multirow{3}{*}{\rotatebox[origin=c]{90}{P-FLK}} & \multicolumn{2}{r|}{\scalebox{.75}{Return}}  &  1241.49   &  \cellcolor{gray!25}\xmark569.78   &  \cellcolor{gray!25}\xmark577.98   &  \cellcolor{red!25}{\xmark639.63}      &   \cellcolor{red!25}{\xmark819.88}     &    \cmark\textbf{3829.75}         \\\cdashline{3-10}[1pt/1pt]
& & \multirow{2}{*}{\rotatebox[origin=c]{90}{t-test}} & \scalebox{.75}{\makecell{PPO vs}} & $-$ & \scalebox{.75}{\makecell{(t(4)=2.09;\\p=1.0e-01)}} & \scalebox{.75}{\makecell{(t(4)=2.04;\\p=1.1e-01)}}& \scalebox{.75}{\makecell{(t(4)=1.56;\\p=1.9e-01)}} & \scalebox{.75}{\makecell{(t(4)=1.07;\\p=3.4e-01)}} & \cellcolor{green!25}{\scalebox{.75}{\makecell{(t(4)=-7.06;\\p=2.1e-03)}}} \\\cdashline{4-10}[1pt/1pt]
& & & \scalebox{.75}{\makecell{Orig vs}} &  $-$ & $-$ & $-$ &\scalebox{.75}{\makecell{(t(4)=-0.20;\\p=8.5e-01)}} & \scalebox{.75}{\makecell{(t(4)=-0.68;\\p=5.4e-01)}} & \cellcolor{green!25}{\scalebox{.75}{\makecell{(t(4)=-10.04;\\p=5.5e-04)}}}\\\cline{2-10}
                                                 & \multirow{3}{*}{\rotatebox[origin=c]{90}{P-RN}} & \multicolumn{2}{r|}{\scalebox{.75}{Return}}  &  2290.53   &  \cellcolor{gray!25}\xmark673.22   & \cellcolor{gray!25}\xmark643.80    & \cellcolor{red!25}{\textbf{\cmark3194.54}}       &   \cellcolor{red!25}{\cmark2476.99}     &    \cmark2617.48         \\\cdashline{3-10}[1pt/1pt]
& & \multirow{2}{*}{\rotatebox[origin=c]{90}{t-test}} & \scalebox{.75}{\makecell{PPO vs}} & $-$ &\cellcolor{green!25}{\scalebox{.75}{\makecell{(t(4)=4.89;\\p=8.1e-03)}}} & \cellcolor{green!25}{\scalebox{.75}{\makecell{(t(4)=5.85;\\p=4.3e-03)}}}& \scalebox{.75}{\makecell{(t(4)=-2.57;\\p=6.2e-02)}} & \scalebox{.75}{\makecell{(t(4)=-0.20;\\p=8.5e-01)}} & \scalebox{.75}{\makecell{(t(4)=-0.28;\\p=7.9e-01)}} \\\cdashline{4-10}[1pt/1pt]
& & & \scalebox{.75}{\makecell{Orig vs}} &  $-$ & $-$ & $-$ & \cellcolor{green!25}{\scalebox{.75}{\makecell{(t(4)=-6.22;\\p=3.4e-03)}}} & \scalebox{.75}{\makecell{(t(4)=-1.96;\\p=1.2e-01)}} & \scalebox{.75}{\makecell{(t(4)=-1.65;\\p=1.7e-01)}} \\\cline{2-10}
                                                 & \multirow{3}{*}{\rotatebox[origin=c]{90}{P-RSM}} & \multicolumn{2}{r|}{\scalebox{.75}{Return}}  &  1559.80   &  \cellcolor{gray!25}\xmark669.06   & \cellcolor{gray!25}\xmark915.44    & \cellcolor{red!25}{\cmark3228.56}       &  \cellcolor{red!25}{\cmark3420.19}      &   \cmark\textbf{3820.27}         \\\cdashline{3-10}[1pt/1pt]
& & \multirow{2}{*}{\rotatebox[origin=c]{90}{t-test}} & \scalebox{.75}{\makecell{PPO vs}} & $-$ &\cellcolor{green!25}{\scalebox{.75}{\makecell{(t(4)=3.42;\\p=2.7e-02)}}} & \scalebox{.75}{\makecell{(t(4)=1.21;\\p=2.9e-01)}}& \cellcolor{green!25}{\scalebox{.75}{\makecell{(t(4)=-4.40;\\p=1.2e-02)}}} & \scalebox{.75}{\makecell{(t(4)=-2.73;\\p=5.2e-02)}} & \cellcolor{green!25}{\scalebox{.75}{\makecell{(t(4)=-8.13;\\p=1.2e-03)}}} \\\cdashline{4-10}[1pt/1pt]
& & & \scalebox{.75}{\makecell{Orig vs}} &  $-$ & $-$ & $-$ &\cellcolor{green!25}{\scalebox{.75}{\makecell{(t(4)=-8.64;\\p=9.9e-04)}}} & \cellcolor{green!25}{\scalebox{.75}{\makecell{(t(4)=-3.17;\\p=3.4e-02)}}} & \cellcolor{green!25}{\scalebox{.75}{\makecell{(t(4)=-21.76;\\p=2.6e-05)}}} \\\cline{2-10}
                                                 & \multirow{3}{*}{\rotatebox[origin=c]{90}{P-RV}} & \multicolumn{2}{r|}{\scalebox{.75}{Return}}  &  1603.86   &  \cellcolor{gray!25}\xmark735.30   & \cellcolor{gray!25}\xmark1151.56    & \cellcolor{red!25}{\cmark1908.67}       &  \cellcolor{red!25}{\cmark2109.97}      &   \cmark\textbf{3360.62}          \\\cdashline{3-10}[1pt/1pt] 
& & \multirow{2}{*}{\rotatebox[origin=c]{90}{t-test}} & \scalebox{.75}{\makecell{PPO vs}} & $-$ &\scalebox{.75}{\makecell{(t(4)=1.30;\\p=2.6e-01)}} & \scalebox{.75}{\makecell{(t(4)=0.53;\\p=6.2e-01)}}& \scalebox{.75}{\makecell{(t(4)=-0.42;\\p=7.0e-01)}} & \scalebox{.75}{\makecell{(t(4)=-0.61;\\p=5.8e-01)}} & \scalebox{.75}{\makecell{(t(4)=-2.25;\\p=8.7e-02)}} \\\cdashline{4-10}[1pt/1pt]
& & & \scalebox{.75}{\makecell{Orig vs}} &  $-$ & $-$ & $-$ &\cellcolor{green!25}{\scalebox{.75}{\makecell{(t(4)=-3.79;\\p=1.9e-02)}}} & \scalebox{.75}{\makecell{(t(4)=-1.32;\\p=2.6e-01)}} & \cellcolor{green!25}{\scalebox{.75}{\makecell{(t(4)=-6.40;\\p=3.1e-03)}}} \\
\bottomrule
\multicolumn{10}{p{13cm}}{\scriptsize{Note: P- stands for the POMDP-version of the tasks.}}
\end{xltabular}
}

\color{black}

\begin{figure*}[htp!]
    \centering
    \begin{minipage}[m]{.7\textwidth}
    \includegraphics[width=\textwidth]{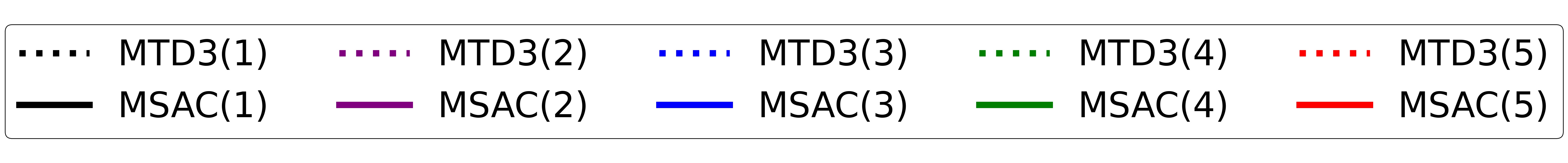}
    \end{minipage}
    \includegraphics[width=.9\textwidth]{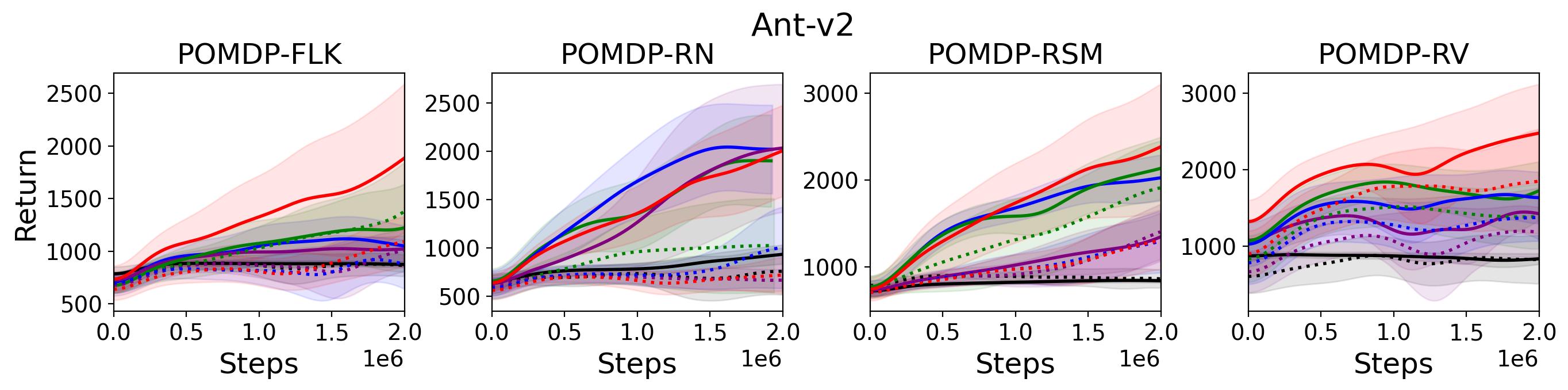}
    \includegraphics[width=.9\textwidth]{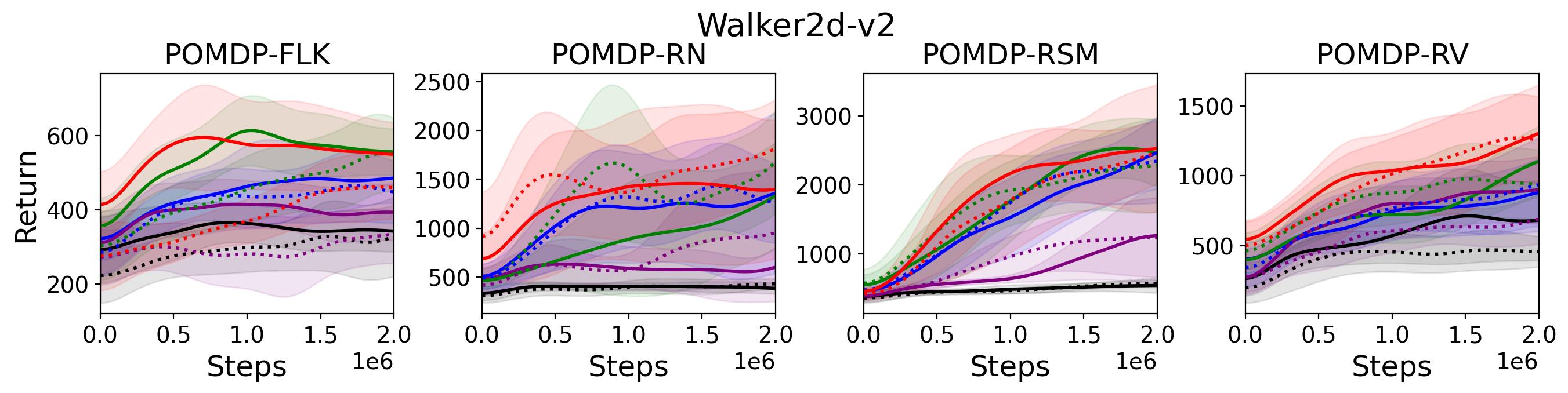}
    \caption{Effect of Multi-step Size on The Performance of MTD3 and MSAC, where the average learning curves correspond to MTD3($n$) and MSAC($n$) with different multi-step sizes $n$ and the shaded area shows half of standard deviation of the average accumulated return over 3 random seeds.}
    \label{fig:Effect_of_Multi_step_Size_on_Performance}
\end{figure*}

\subsection{Exploring the Effect of Multi-step Bootstrapping}
\label{subsec:Results_on_Understand_the_Effect_of_Multi-step_Bootstrapping}

To address \textbf{Research Questions 1 and 2} and test point (1) of \textbf{Hypothesis \ref{hypothesis_compare_n_step_TD3_and_SAC_to_vanilla}}, we compare the performance of MTD3(n) with TD3 and that of MSAC(n) with SAC in Table \ref{tab:Maximum_of_Average_Return}. As highlighted in red in Table \ref{tab:Maximum_of_Average_Return}, MTD3(5) significantly outperforms TD3 on 14/16 POMDP tasks, and similarly MSAC(5) significantly outperforms SAC on 15/16 POMDP tasks, directly answering \textbf{Research Question 1} about multi-step efficacy across different partial observability mechanisms. More obvious performance improvement can be seen in Fig. \ref{fig:Average_Return_Over_Tasks} that for most tasks when using multi-step bootstrapping TD3 and SAC achieve double performance on POMDPs compared to their vanilla version using 1-step bootstrapping, where for most cases the performance differences are statistically significant as shown in Fig. \ref{fig:Average_Return_Over_Tasks}.

\begin{figure}[htp!]
    \centering
    \includegraphics[width=.65\linewidth]{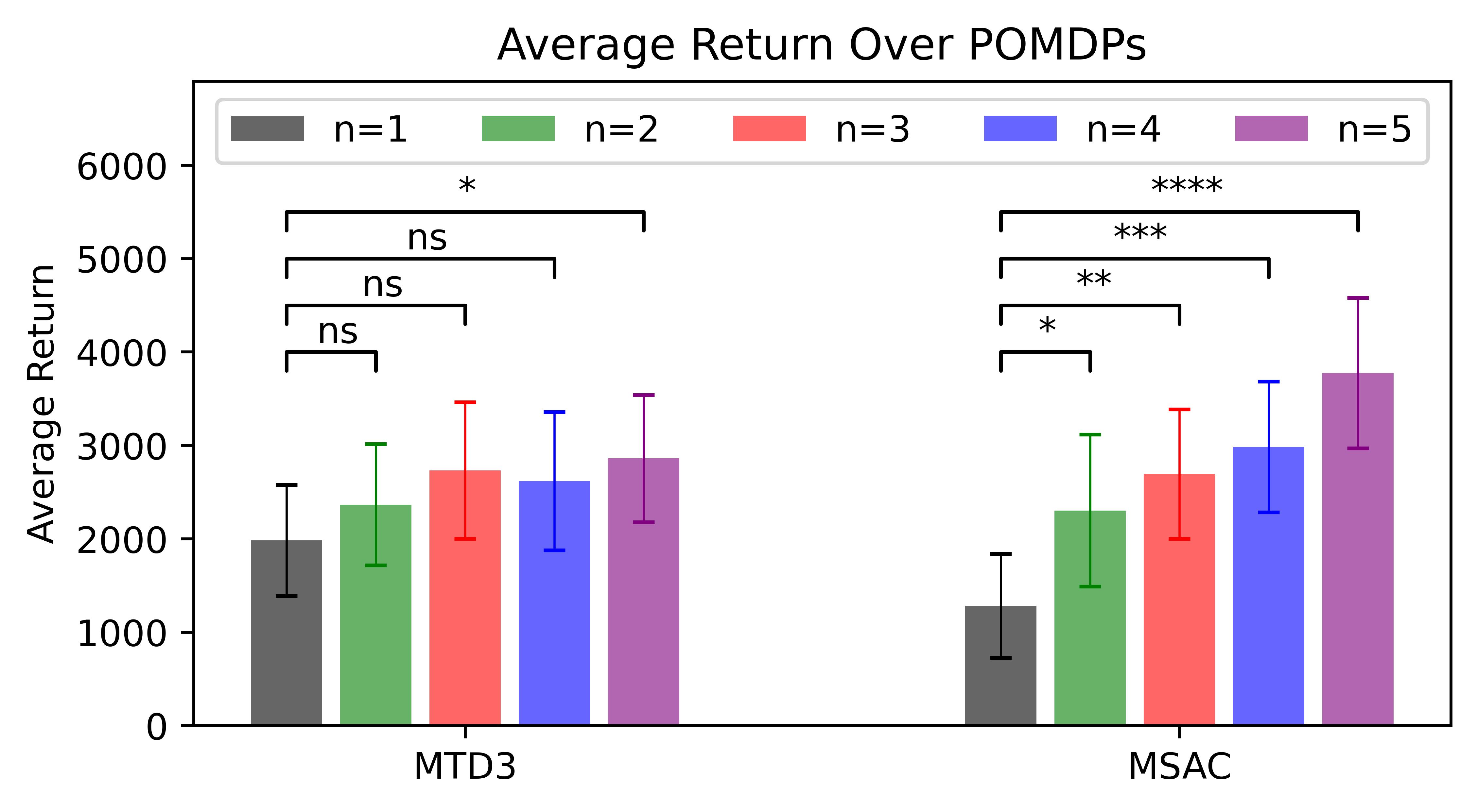}
    \caption{Bar Chart of Average Return Over 16 POMDPs for Different Multi-step Sizes shown in Table \ref{tab:Maximum_of_Average_Return}. The error bars indicate the 95\% confidence intervals. Two-sample two-sided T-Test is conducted between two algorithms under the end of each bracket. The labels above the bracket are as follows: ns: $p>0.05$, $\ast$: $p\leq0.05$, $\ast\ast$: $p\leq0.01$, $\ast\ast\ast$: $p\leq0.001$, $\ast\ast\ast\ast$: $p\leq0.0001$.}
    \label{fig:different_multistep_size_average_return_over_pomdps_bar_chart}
\end{figure}

Regarding \textbf{Research Question 2} about multi-step vs single-step performance, Table \ref{tab:Proportion_of_POMDPs_That_n_greater_than_1_Outperforming_n_equal_to_1} demonstrates that multi-step methods consistently outperform single-step approaches: MTD3 with n>1 outperforms n=1 on up to 14/16 tasks, while MSAC achieves perfect 16/16 task improvement with n=5. The choice of n=5 as our primary comparison point is justified by Fig. \ref{fig:different_multistep_size_average_return_over_pomdps_bar_chart}, which shows that n=5 represents a practical sweet spot where performance gains plateau, making it sufficient for robust improvements without excessive computational overhead. More interestingly, simply using multi-step bootstrapping with step size equal to 5 MTD3(5) and MSAC(5) achieves performance comparable to or even better than LSTM-TD3(5), e.g., POMDP-RN and POMDP-RSM, which is specifically designed to deal with POMDPs. Nevertheless, for some other cases, e.g., POMDP-FLK and POMDP-RV, directly learning a good representation of the underlying state from a short experience trajectory as that in LSTM-TD3(5) is more effective than relying on multi-step bootstrapping to pass some temporal information. This observation completes our answer to \textbf{Research Question 3}, showing that recurrent methods become necessary when information degradation is severe (as in FLK and RV variants) and simple temporal aggregation through multi-step bootstrapping is insufficient.

\begin{table}[htp!]
    \centering
    \caption{Proportion of POMDPs That $n>1$ Outperforming $n=1$ for MTD3(n) and MSAC(n)} 
    \label{tab:Proportion_of_POMDPs_That_n_greater_than_1_Outperforming_n_equal_to_1}
    \footnotesize
    \begin{tabular}{cccc|cccc}
        \toprule\hline
         \multicolumn{4}{c|}{MTD3(n)}&  \multicolumn{4}{c}{MSAC(n)}\\\hline
         $1<2$&  $1<3$&  $1<4$&  $1<5$&  $1<2$&  $1<3$&  $1<4$& $1<5$\\\hline
         11/16&  13/16&  13/16&  14/16&  14/16&  13/16&  14/16& 16/16\\
         \bottomrule
    \end{tabular}
\end{table}

To investigate how the multi-step size affects the performance of MTD3($n$) and MSAC($n$), Fig. \ref{fig:Effect_of_Multi_step_Size_on_Performance} shows the average learning curves of MTD3($n$) and MSAC($n$) with different multi-step sizes $n\in\left \{1,2,3,4,5\right \}$. When $n=1$ these reduce to TD3 and SAC, respectively\footnote{More results on other tasks can be found in \url{https://arxiv.org/abs/2209.04999}.}. It can be seen from Fig. \ref{fig:Effect_of_Multi_step_Size_on_Performance} that simply increasing $n$ by a few steps increases performance dramatically over the $n=1$ case with limited extra computational cost. For the $n$ we tested, $n=5$ shows the best performance on most tasks. Table \ref{tab:Proportion_of_POMDPs_That_n_greater_than_1_Outperforming_n_equal_to_1} summarizes the best performance of MTD3($n$) and MSAC($n$) with different multi-step sizes $n\in\left \{1,2,3,4,5\right \}$, where we specifically compare $n>1$ with $n=1$ for MTD3($n$) and MSAC($n$) on the 16 POMDP tasks. It can be seen that simply increasing step size from 1 to 2, results in an MTD3(2) performance improvement on 11/16 tasks, while MSAC(2) improves its performance on 14/16 tasks. With the larger step size used, both MTD3($n$) and MSAC($n$) get much better performance. Fig. \ref{fig:different_multistep_size_average_return_over_pomdps_bar_chart} compares the average performance of MTD3 and MSAC with different step size $n$ over all 16 POMDPs and shows the significance level between each comparison. From this figure, we can see that for MTD3 all $n>1$ outperform $n=1$ even though only $n=5$ shows significant difference. However, for MSAC, all $n>1$ significantly outperform $n=1$. These findings all support (1) of \textbf{Hypothesis \ref{hypothesis_compare_n_step_TD3_and_SAC_to_vanilla}} and reveal the benefit of multi-step bootstrapping when handling POMDPs. It is worth noting that further increasing $n$ will degrade performance eventually, e.g., $n=8$, depending on the task, consistent with the findings in \cite{meng2021effect}. The reason is that TD3 and SAC are both off-policy RL algorithms and the experiences sampled from the replay buffer are not optimal, due to the exploratory actions and the non-optimal stage of the policy, which means that the estimated Q-value is inaccurate and the longer the step size the farther the estimated Q-value is from the true Q-value.

\begin{figure}[htp!]
    \centering
    \includegraphics[width=.55\textwidth]{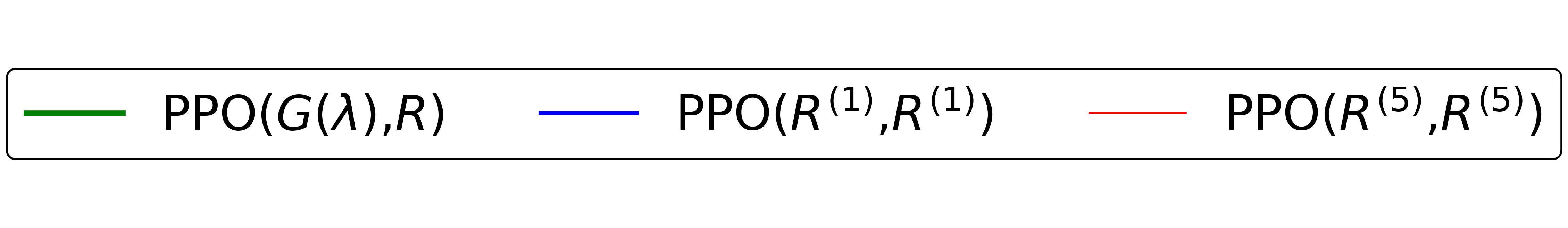}
    \includegraphics[width=.9\textwidth]{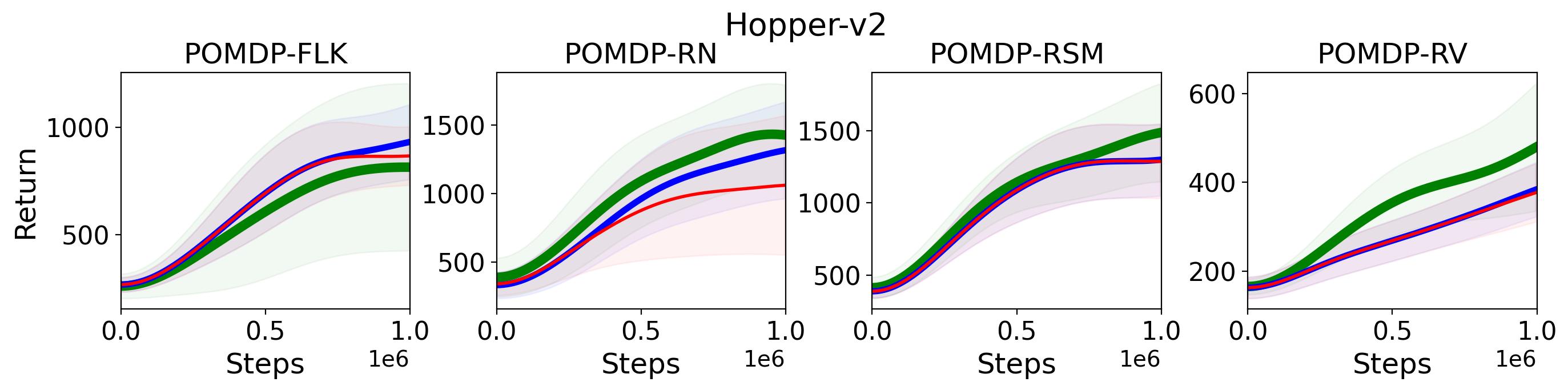}
    \includegraphics[width=.9\textwidth]{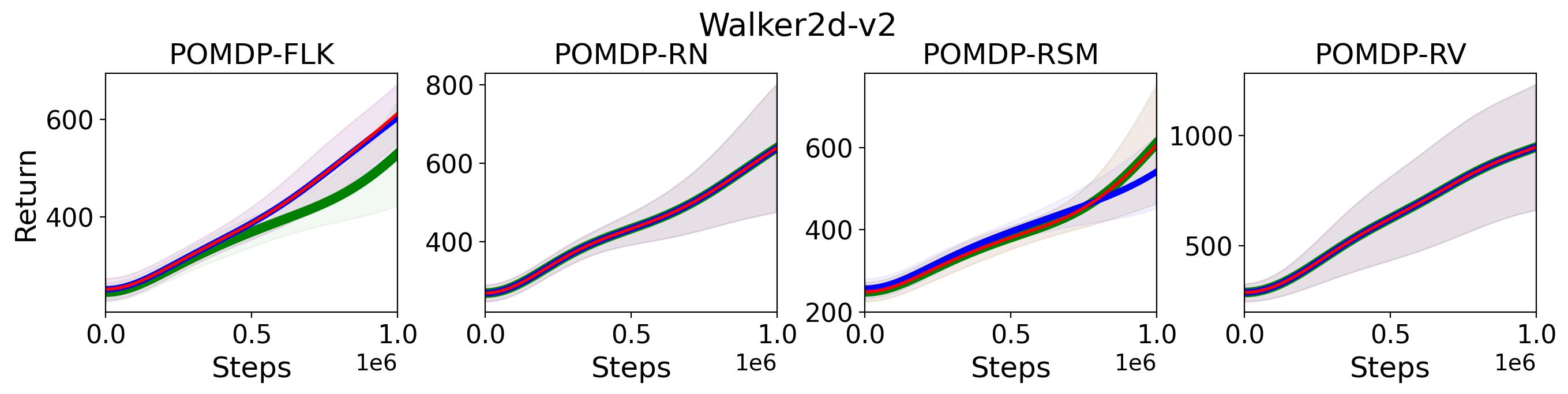}
    \caption{Effect of Multi-step Size on The Performance of PPO, where PPO($G(\lambda)$, $R$) corresponds to the original PPO using $\lambda$-return $G(\lambda)$ for advantage estimate and Monte-Carlo return $R$ for state-value function update, and PPO($R^{(n)}$, $R^{(n)}$) corresponds to the revised PPO using $n$-step bootstrapped return for both advantage estimate and state-value function update. The shaded area shows half of standard deviation of the average accumulated return over 3 random seeds.}
    \label{fig:Effect_of_Multi_step_Size_on_Performance_of_PPO}
\end{figure}

To validate (2) of the \textbf{Hypothesis \ref{hypothesis_compare_n_step_TD3_and_SAC_to_vanilla}}, we replace the $\lambda$-return $G(\lambda)$ used to calculate generalized advantage estimate and Monte-Carlo return $R$ used to update state-value function with $n$-step bootstrapped return $R^{(n)}$ for both advantage estimate and policy update, where results for PPO($R^{(n)}$,$R^{(n)}$) with $n\in\{1,5\}$ is compared with the original PPO($G(\lambda)$, $R$) in Fig. \ref{fig:Effect_of_Multi_step_Size_on_Performance_of_PPO} \footnote{More results on other tasks can be found in \url{https://arxiv.org/abs/2209.04999}.}. As shown in this figure, there is no clear difference among PPO($G(\lambda)$, $R$), PPO($R^{(1)}$,$R^{(1)}$) and PPO($R^{(5)}$,$R^{(5)}$). For example, on Walker2d-v2 POMDP-RN and POMDP-RV the performance of these three variants of PPO almost perfectly overlapped. Therefore, the results shown in Fig. \ref{fig:Effect_of_Multi_step_Size_on_Performance_of_PPO} rejects the (2) of the \textbf{Hypothesis \ref{hypothesis_compare_n_step_TD3_and_SAC_to_vanilla}}

To summarize, the results shown in this section support (1) of \textbf{Hypothesis \ref{hypothesis_compare_n_step_TD3_and_SAC_to_vanilla}} that multi-step versions of TD3 and SAC, i.e., MTD3(n) and MSAC(n), with $n>1$ significantly improve their performance compared to their vanilla versions, but (1) of \textbf{Hypothesis \ref{hypothesis_compare_n_step_TD3_and_SAC_to_vanilla}} is rejected by our results that replacing the $\lambda$-return and Monte-Carlo return with 1-step bootstrapped return does not cause the performance decrease for PPO.

\begin{figure*}[htp!]
    \centering
    \includegraphics[width=\textwidth]{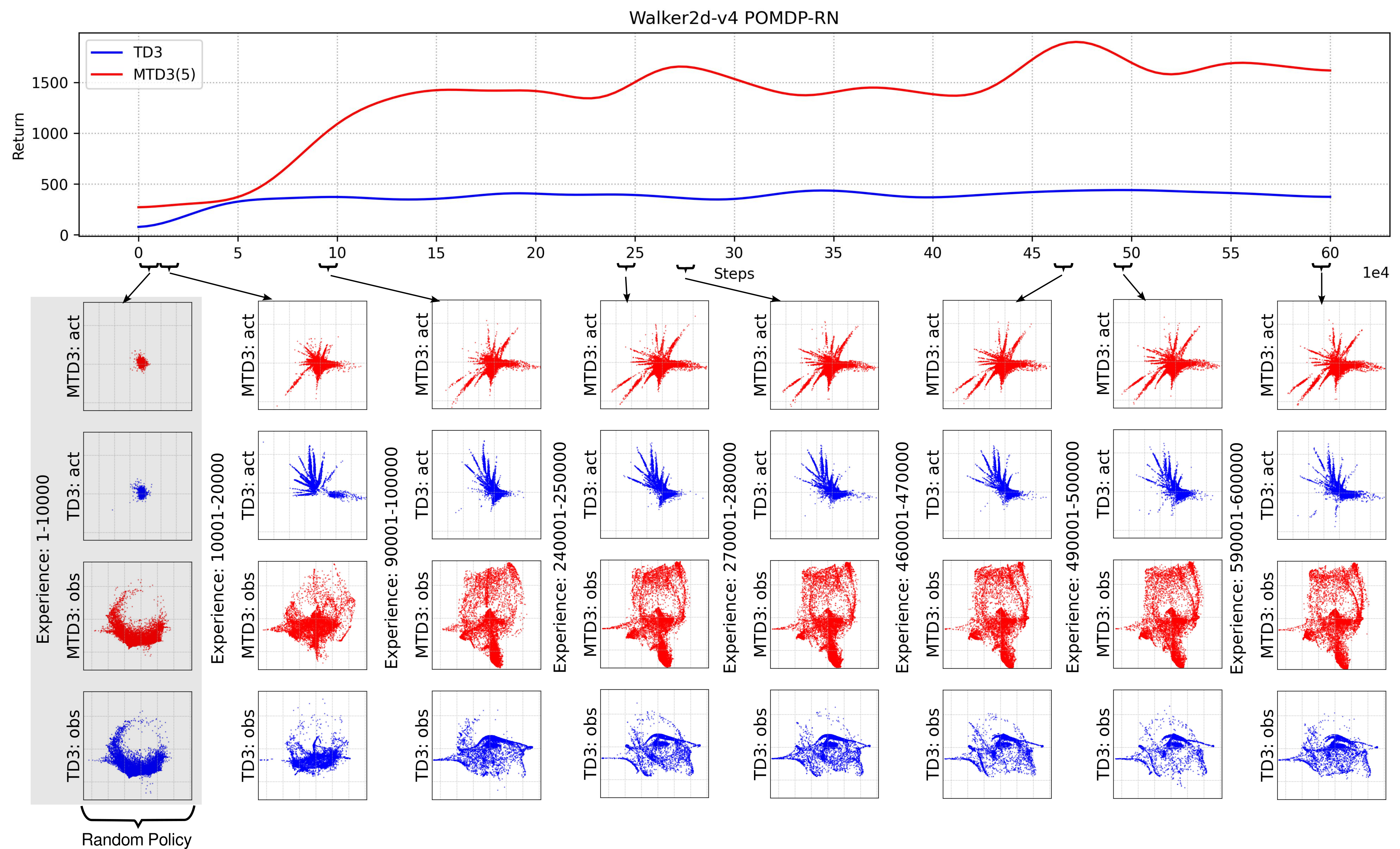}
    \caption{Observation Coverage Comparison of Policies from TD3 and MTD3(5) on Walker2d-v4 POMDP-RN, where the top panel shows the returns of the policies of TD3 and MTD3(5), while the bottom panels show the observations and actions stored in the replay buffer and embedded by TriMap\cite{2019TRIMAP}. The first 10000 experiences are collected by a random policy for both TD3 and MTD3(5), and policy training starts after 10000 steps for both TD3 and MTD3(5).}
    \label{fig:Observation_Coverrage_Comparison_TD3_MTD3}
\end{figure*}

\subsection{Observation and Action Coverage of Policy With One-step or Multi-step Bootstrapping}
\label{subsec:Observation_and_Action_Coverage_of_Policy_With_One_step_or_Multi_step_Bootstrapping}

In Section \ref{subsec:Results_on_Understand_the_Effect_of_Multi-step_Bootstrapping}, we demonstrated the effect of multi-step bootstrapping on improving TD3's performance on POMDPs in Fig. \ref{fig:Effect_of_Multi_step_Size_on_Performance}. The measurement used in that section is the accumulated reward, i.e. return. In this section, we will have a look at the difference in terms of observation and action coverage between TD3 and MTD3(5) leveraging dimensionality reduction technology to get more insights on the difference in the policies induced from one-step and multi-step bootstrapping. Specifically, we will try to embed the high-dimensional observation and action space into 2D space for visualization.

Fig. \ref{fig:Observation_Coverrage_Comparison_TD3_MTD3} compares the observation coverage of the policy learned by TD3 and MTD3(5) on Walker2d-v4 POMDP-RN, where each of the bottom panels shows 10000 observations and the actions taken by TD3 and MTD3(5) in these observations. In order to compare, we first separately combine the observations and actions collected by TD3 and MTD3(5) and then embed them into 2D using TriMap\cite{2019TRIMAP}, where TriMap is used for dimensionality reduction because it is good at maintaining the relative distances of the clusters compared to other methods, e.g., T-SNE \cite{van2008visualizing}. Because for both TD3 and MTD3(5) the first 10000 steps are driven by a random policy, the observation and action coverage of TD3 and MTD3(5) are very similar, as shown in the first column of Fig. \ref{fig:Observation_Coverrage_Comparison_TD3_MTD3}. However, once the learned policy is used to take actions, there is an immediate observation and action coverage difference between TD3 and MTD3(5) as shown in the 2nd column of the figure where the experiences from time step 10001 to 20000 are plotted. As the learning continuing, not only the TD3 and MTD3 have different observation and action coverage for time step from 90001 to 100000, they also have different coverage compared to that from time step 10001 to 20000. This indicates the policies of TD3 and MTD3(5) are evolving to different local optimal. From the last few columns of Fig. \ref{fig:Observation_Coverrage_Comparison_TD3_MTD3}, we can see from the observation and action coverage plots the policies of TD3 and MTD3(5) are relatively stable after 150000 steps, and there is a distinct difference between TD3 and MTD3(5), which also matches the difference in the performance of the policies.

\begin{figure*}[t!]
    \centering
    \includegraphics[width=.85\textwidth]{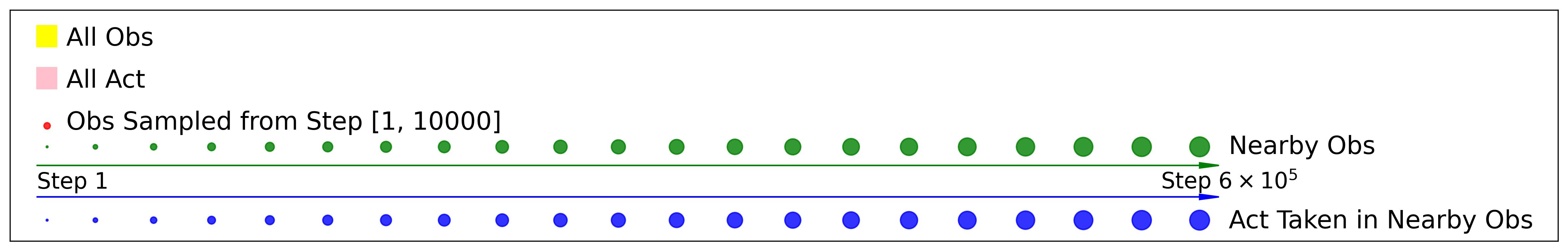}
    \includegraphics[width=.32\textwidth]{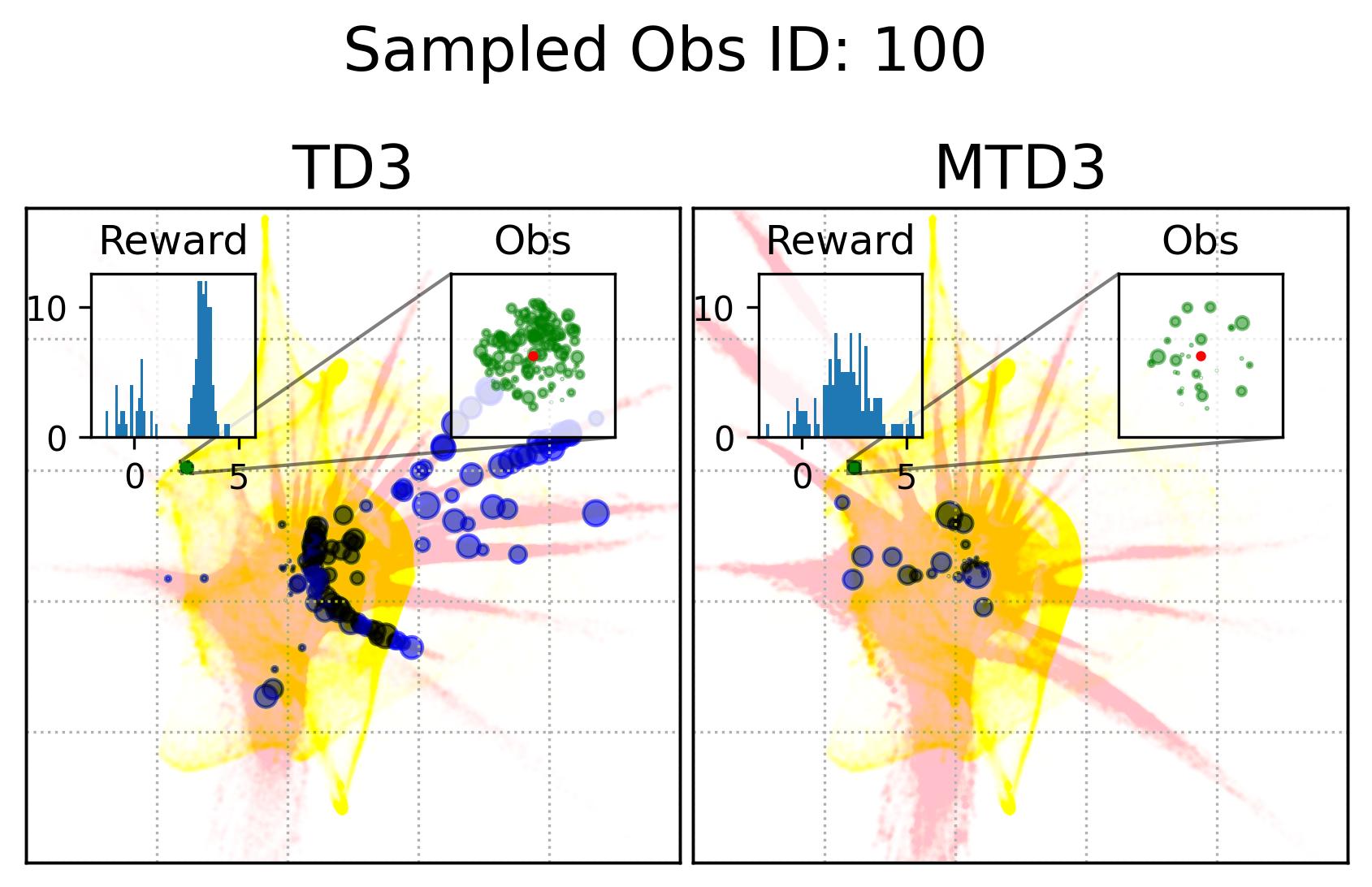}
    \includegraphics[width=.32\textwidth]{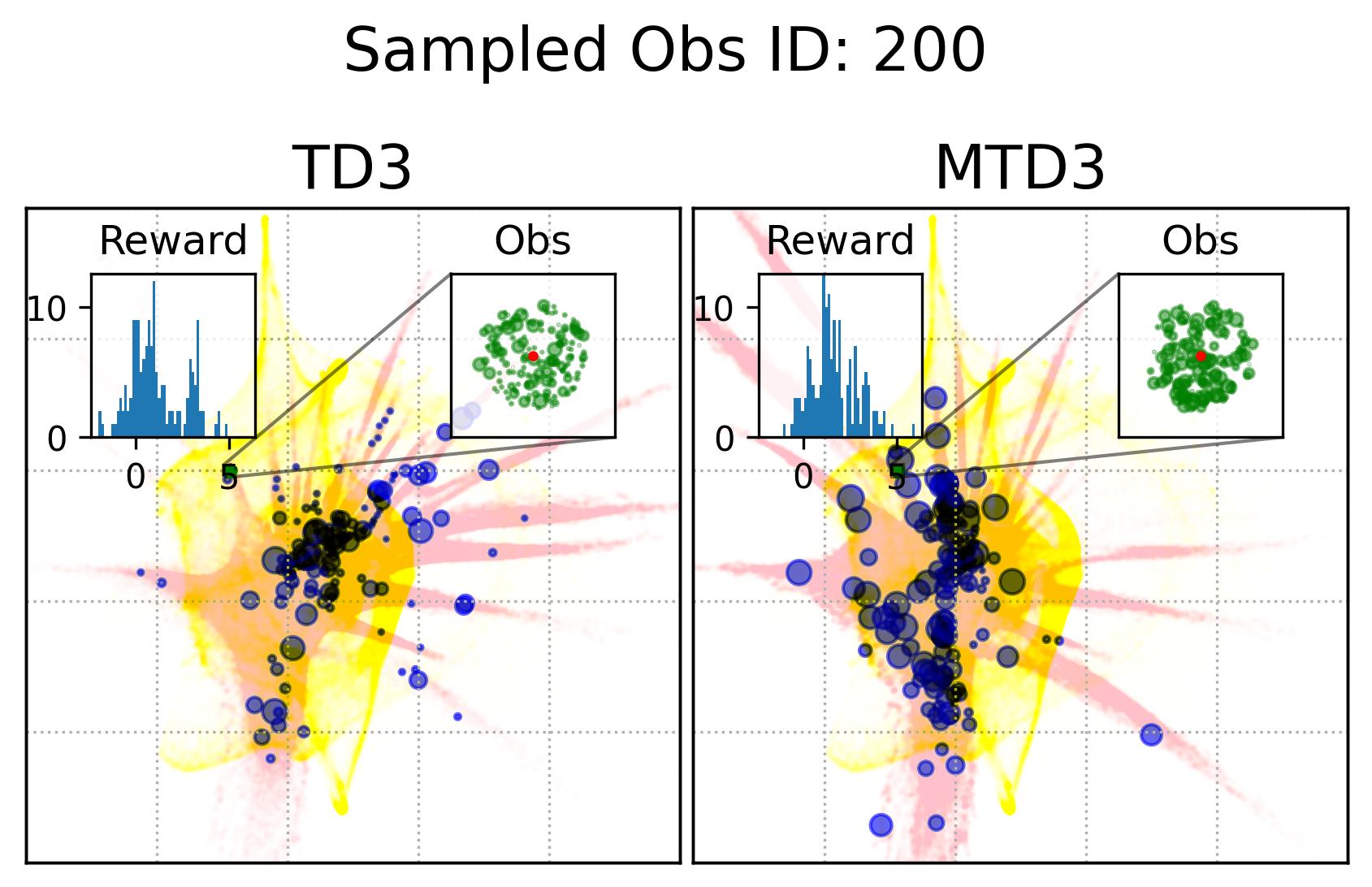}
    \includegraphics[width=.32\textwidth]{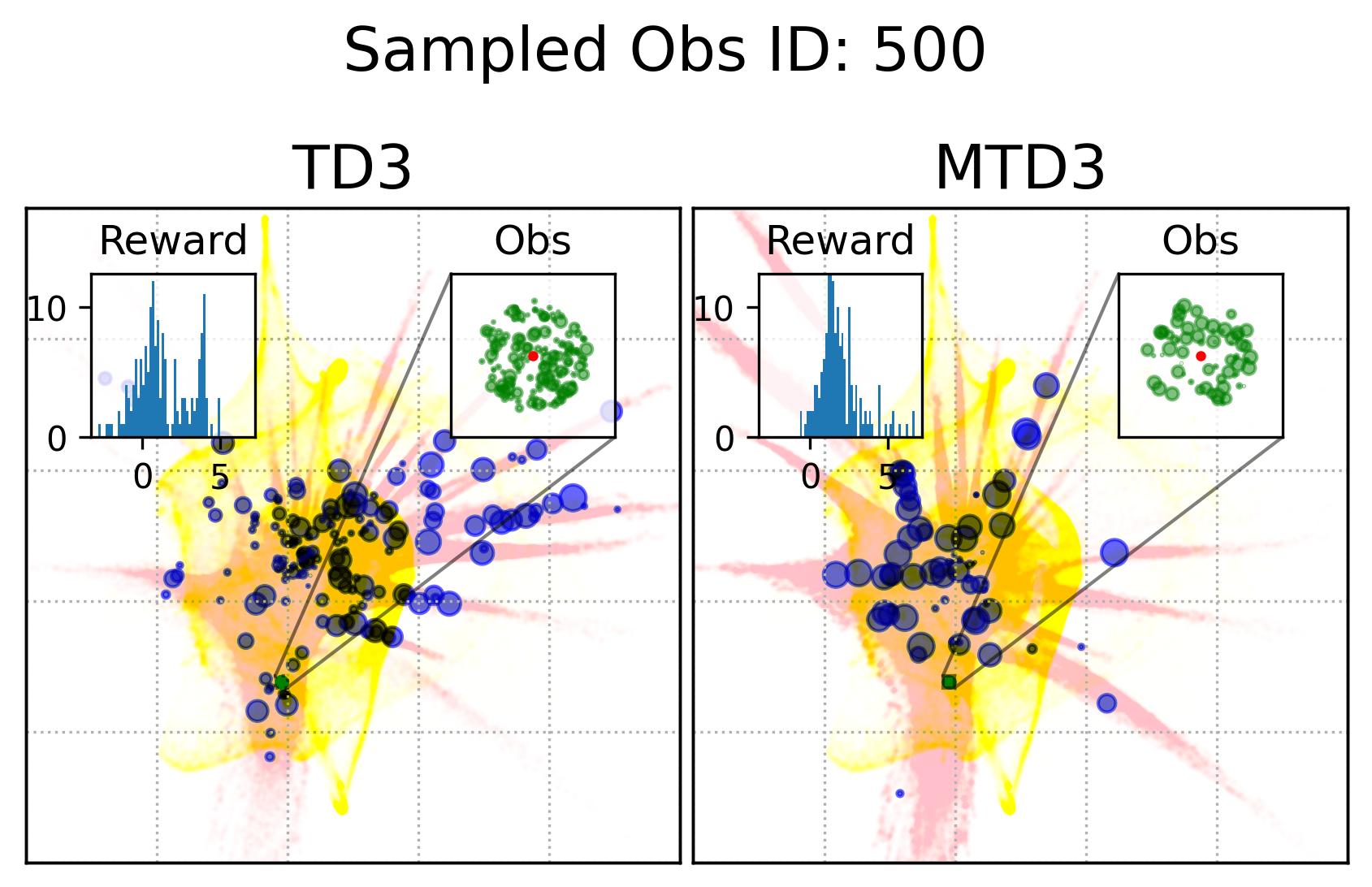}
    \includegraphics[width=.32\textwidth]{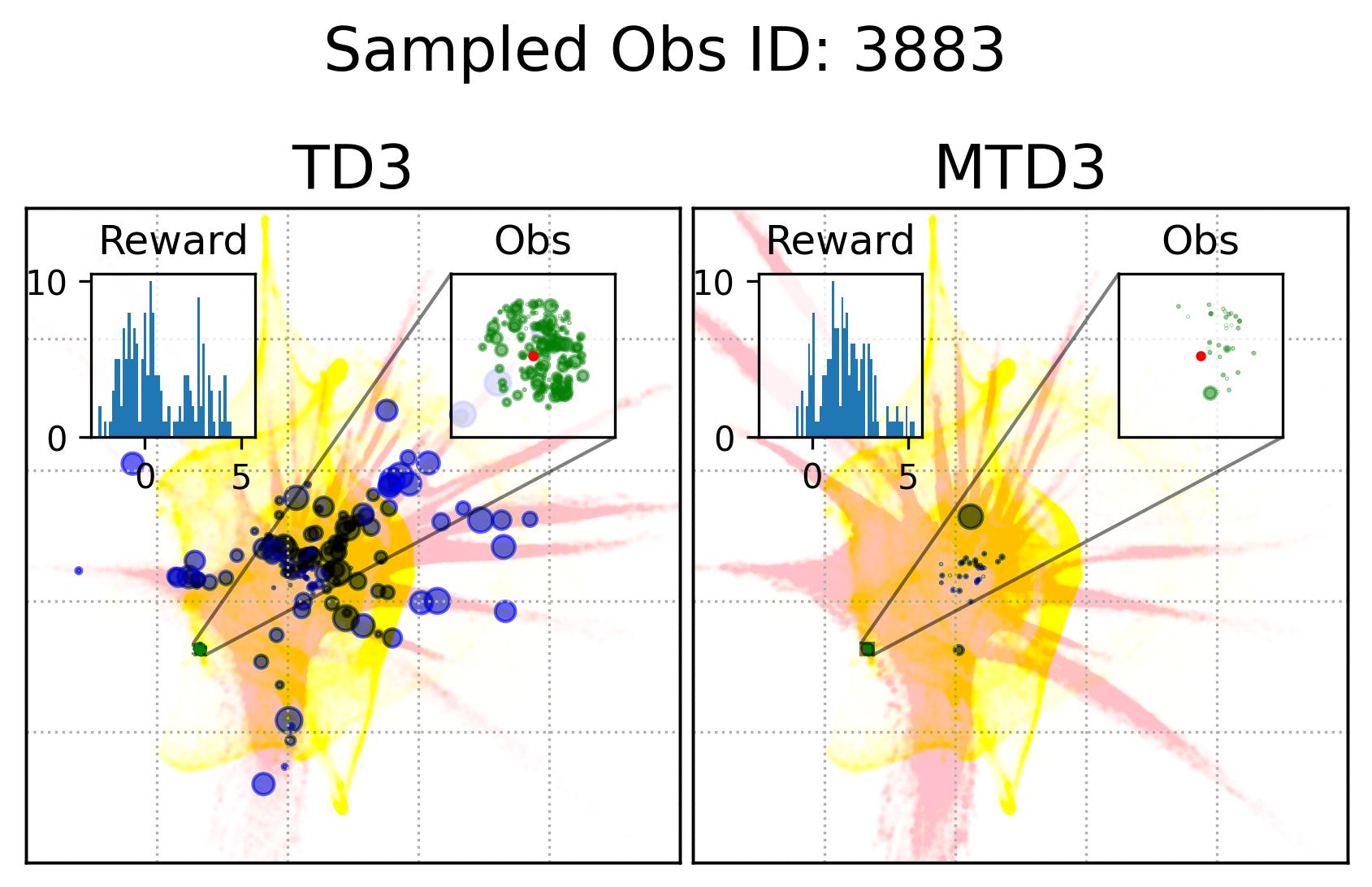}
    \includegraphics[width=.32\textwidth]{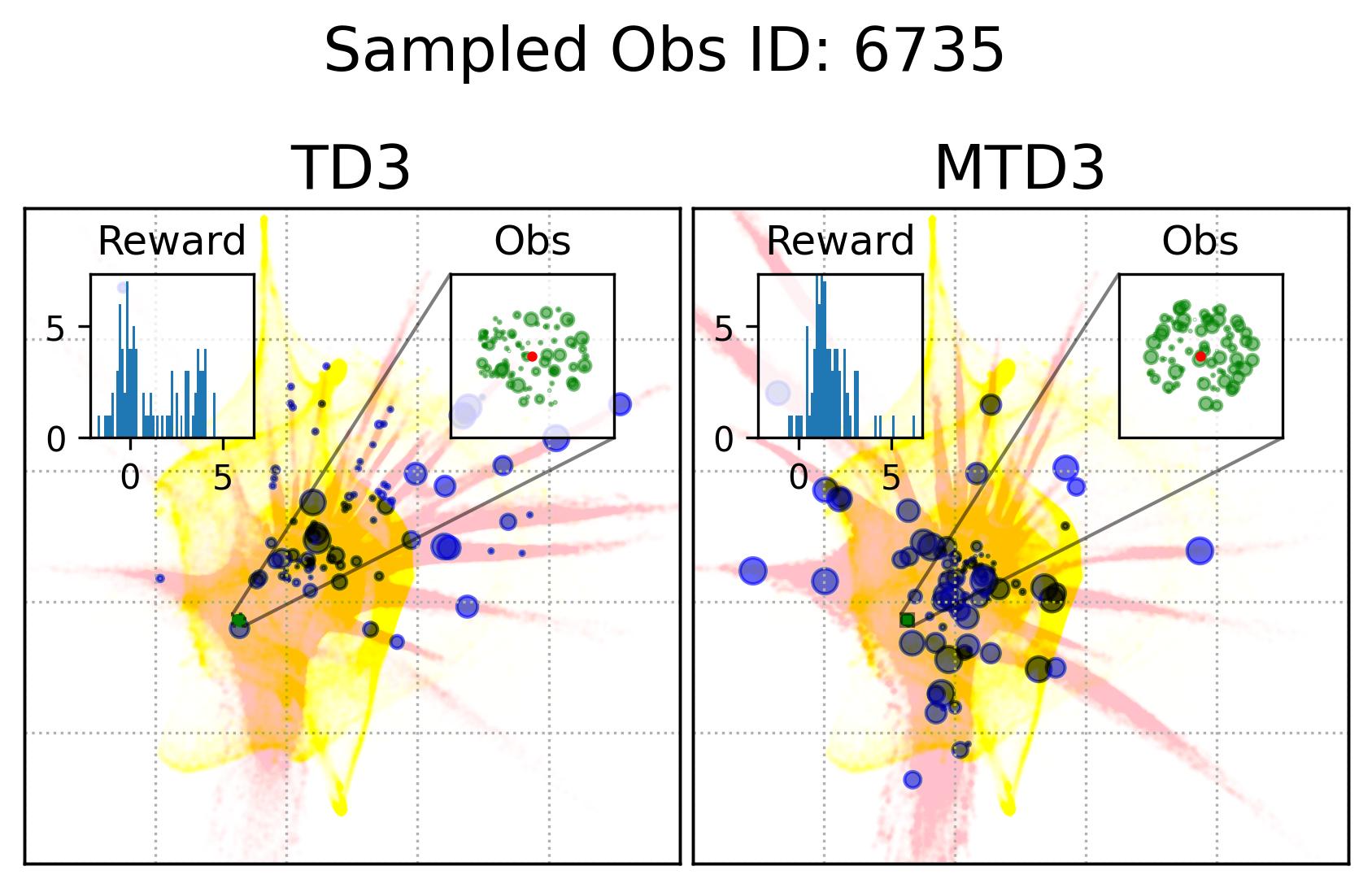}
    \includegraphics[width=.32\textwidth]{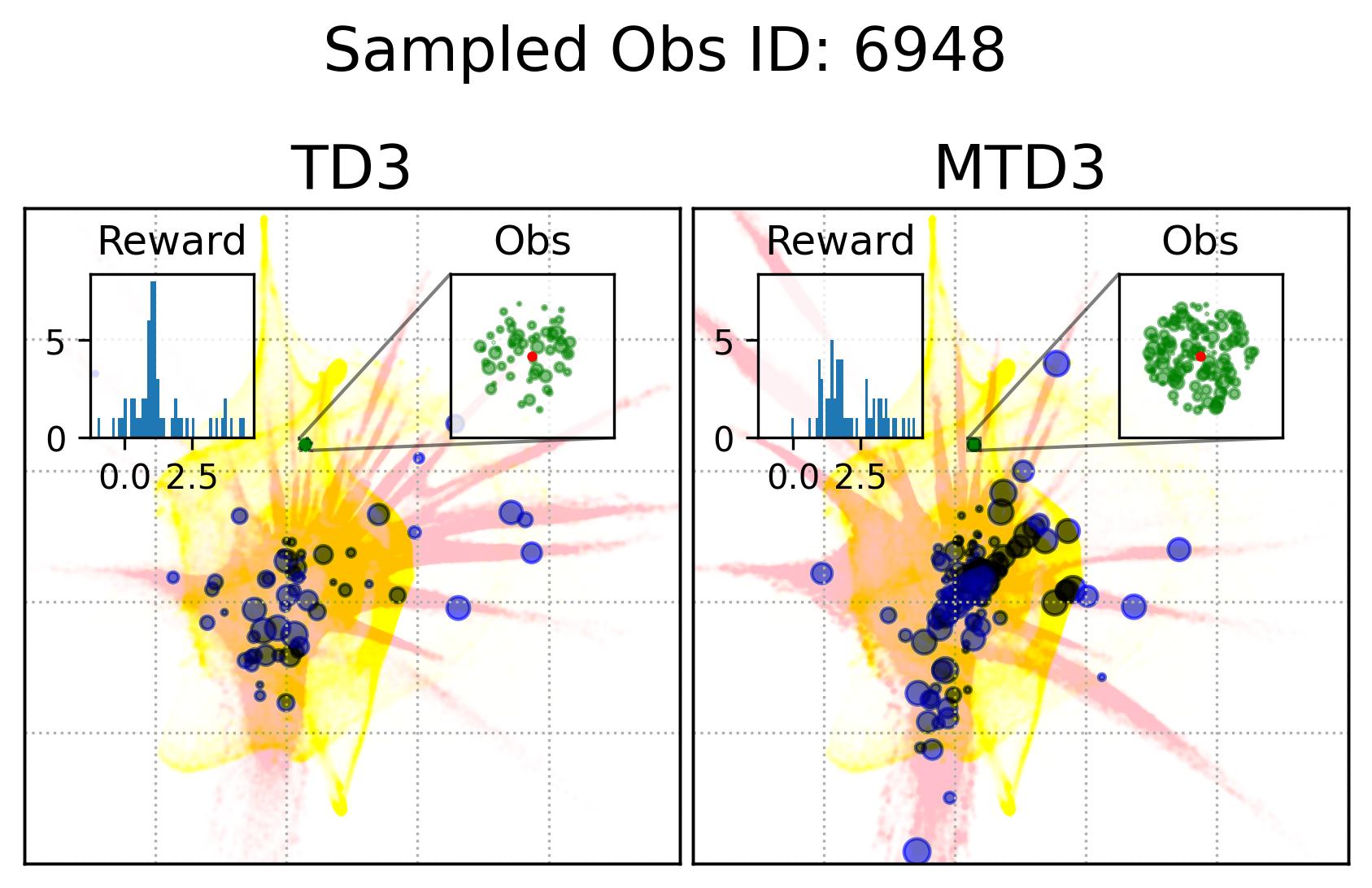}
    \caption{Actions Taken by TD3 and MTD3(5) in Nearby Observations. To visualize the observations and actions, we first combine the $6\times10^5$ observations from TD3 and MTD3(5) replay buffers, then embed the combined observations into 2D. Similar operations are done for actions. Then, we randomly sample 6 observations (red dot in each panel) from the first 10000 experiences, and find the nearby observations (green dots in each panel) of each of the 6 observations and visualize the corresponding actions (blue dots in each panel) taken in these nearby observations. As indicated in the legend, we use the size of the dots to better reflect the happening time step of the nearby observations and their corresponding actions.}
    \label{fig:Action_Taken_by_TD3_and_MTD3_on_Nearby_Observations}
\end{figure*}

Fig. \ref{fig:Observation_Coverrage_Comparison_TD3_MTD3} shows the observation and action coverage difference between TD3 and MTD3(5) from a broader perspective where the observations and actions within one period of 10000 time-steps are compared to another period. To have a closer look at their differences at a specific observation, Fig. \ref{fig:Action_Taken_by_TD3_and_MTD3_on_Nearby_Observations} shows the actions taken in observations that are nearby an observation sampled from the first 10000 experiences. The nearby observations $o_t$ of an observation $o$ is defined as $\left \{ o_t \in \left \{ o_1, \cdots , o_{6\times10^5} \right \} |  \left \| o_t - o \right \|_{2} \leq 2\right \}$ where the $o_t$ and $o$ are embedded 2D observations whose value range for each dimension is $[-155, 175]$. The neighbour observations (the green dots) of a sampled observation (the red dot) are better visualized in the inset of each panel. The corresponding actions taken in these neighbour observations are represented by the blue dots whose sizes are used to reflect the happening time of these actions where the more recent the action was taken the bigger the dot will be, as indicated by the legend of Fig. \ref{fig:Action_Taken_by_TD3_and_MTD3_on_Nearby_Observations}. In addition, to give the readers a sense of where the sampled observation, the neighbours of the sampled observation, and the action taken in these neighbour are relative to the whole embedded observations and actions, Fig. \ref{fig:Action_Taken_by_TD3_and_MTD3_on_Nearby_Observations} also show the whole embedded observations in yellow and whole embedded actions in pink. To show how good these actions are, the histogram of the reward of taking these actions in the neighbour observations is shown in each panel as well. The 1st distinct difference between TD3 and MTD3(5) that can be drawn from Fig. \ref{fig:Action_Taken_by_TD3_and_MTD3_on_Nearby_Observations} is that the actions taken by MTD3(5) in the neighbour observations are quiet different from that taken by TD3, which indicates the policy difference, and the actions taken by MTD3(5) correspond to higher rewards compared to that of TD3 as shown in the inset titled ``Reward". However, what is common between TD3 and MTD3(5) is that even though the neighbour observations are tightly close to each other, the actions taken in these observations are not so close to each other. This can be because the small difference in the observation may cause a big difference in optimal action. Another explanation could be the relative distances of actions are not well maintained by TriMap. It is also interesting to see that some observations are frequently encountered by TD3 but are rarely observed in MTD3(5), especially at the newest interactions, e.g., the 1st column of the 1st and the 2nd row, which is also an indicator of the difference between TD3 and MTD3(5) and showing that MTD3(5) learns a policy to avoid these observations whereas TD3 is stuck in a local optimal that cannot escape from it.  

As a summary, the results shown in this section demonstrate the difference in the policies derived from one-step and multi-step, specifically 5-step, bootstrapping for TD3, as a measurement to reflect the difference in learning performance, namely accumulated reward. We know that the policy of TD3 and MTD3 is derived from the action-value function, so this means essentially multi-step bootstrapping leads to a better value function than one-step bootstrapping that is robuster to partial observation of the underlying state. However, it is still unclear why simply replacing the one-step bootstrapping with a multi-step bootstrapping can make such a difference in the performance of DRL algorithms, i.e., TD3 and SAC, on POMDPs, especially given that the (2) of the \textbf{Hypothesis \ref{hypothesis_compare_n_step_TD3_and_SAC_to_vanilla}} is rejected in Section \ref{subsec:Results_on_Understand_the_Effect_of_Multi-step_Bootstrapping}.

\begin{table}[t!]
    \centering
    \caption{Performance Comparison Summary On Accumulated Reward}
    \label{tab:Performance_Comparison_Summary_On_Accumulated_Reward}
    \scriptsize
    
    \begin{tabular}{l| p{1cm} | p{3.2cm} | p{1cm} | p{3.2cm}}  \toprule\hline 
         \multirow{2}{*}{\textbf{Comparison}} & \multicolumn{2}{c|}{$r_t^{avg(5)}$} & \multicolumn{2}{c}{$r_t^{sum(5)}$}\\\cline{2-5}
          & ratio & t-test & ratio & t-test \\\hline
         TD3 $<$ TD3$+(\cdot)$ & 6/16 & (t(30)=-0.05;p=9.6e-01) & 6/16 & (t(30)=0.12;p=9.0e-01)\\ \hline 
         SAC $<$ SAC$+(\cdot)$ & 5/16 & (t(30)=0.43;p=6.7e-01)& \textbf{14/16} & (t(30)=-1.36;p=1.9e-01)\\ \hline 
         \cellcolor{gray!25}MTD3(5) $<$ \cellcolor{gray!25}TD3$+(\cdot)$ & \cellcolor{gray!25}2/16 & \cellcolor{gray!25}(t(30)=2.39;p=2.3e-02) & \cellcolor{gray!25}1/16 & \cellcolor{gray!25}(t(30)=2.67;p=1.2e-02)\\ \hline 
         \cellcolor{gray!25}MSAC(5) $<$ \cellcolor{gray!25}SAC$+(\cdot)$ & \cellcolor{gray!25}0/16 & \cellcolor{gray!25}(t(30)=4.41;p=1.2e-04) & \cellcolor{gray!25}0/16 & \cellcolor{gray!25}(t(30)=2.75;p=1.0e-02)\\ \hline
         MTD3(5) $<$ MTD3(5)$+(\cdot)$ & 5/16 & (t(30)=0.70;p=4.9e-01)&  \textbf{10/16} & (t(30)=0.23;p=8.2e-01)\\ \hline 
         MSAC(5) $<$ MSAC(5)$+(\cdot)$ & 4/16 & (t(30)=1.11;p=2.7e-01)& 5/16 & (t(30)=0.71;p=4.8e-01)\\ \hline\bottomrule
         \multicolumn{5}{p{8cm}}{\footnotesize{Note: $+(\cdot)$ indicates either $r_t^{avg(5)}$ or $r_t^{sum(5)}$ is used.}}
    \end{tabular}
\end{table}

\subsection{Effect of Accumulated Reward}
\label{subsec:Effect_of_Accumulated_Reward}

To validate \textbf{Hypothesis \ref{hypothesis_compre_accumulated_reward_task_to_one_step_reward_task}}, results on environment using average reward $r_t^{avg(n)}$ (Eq. \ref{eq:accumulated_reward_env_avg}) and summation reward $r_t^{sum(n)}$ (Eq. \ref{eq:accumulated_reward_env_sum}) over $n$ consecutive rewards are compared with that on environment using the original reward. Table \ref{tab:Performance_Comparison_Summary_On_Accumulated_Reward} shows the aggregated results on the 16 POMDPs\footnote{The results on each task can be found in \url{https://arxiv.org/abs/2209.04999}.}. It shows that the proportion of the 16 POMDPs that the performance of an algorithm (right side of $<$) on environment using accumulated reward, i.e., $r_t^{avg(n)}$ or $r_t^{sum(n)}$ defined in Eq. \ref{eq:accumulated_reward_env_avg} and Eq. \ref{eq:accumulated_reward_env_sum} respectively, is better than that of another algorithm (left side of $<$) on environment using original reward ${r}$.  Over all comparisons, only two cases, i.e., $\text{SAC}<\text{SAC}+(\cdot)$ and $\text{MTD3(5)}<\text{MTD3(5)}+(\cdot)$, experience majority (over 50\% of tasks) performance improvement, while for other cases there is a performance decrease for most tasks. However, none of them (row 1, 2, 5, 6 in Table \ref{tab:Performance_Comparison_Summary_On_Accumulated_Reward}) shows statistically significant difference. What is particularly evident is that TD3 and SAC on environments using accumulated reward cannot achieve comparable performance to MTD3(5) and MSAC(5) on environments using the original reward, with a significant difference $p<0.05$, as highlighted in gray in Table \ref{tab:Performance_Comparison_Summary_On_Accumulated_Reward}.

Overall, the results shown here reject \textbf{Hypothesis \ref{hypothesis_compre_accumulated_reward_task_to_one_step_reward_task}}, which are: (1) the performance of TD3 and SAC on environment using accumulated reward is not consistently improved compared to their performance on environment using the original reward, and (2) for cases there are performance improvement, the performance improving scale is not comparable to that of algorithms using multi-step bootstrapping. This hints the mechanism underlying the performance improvement and robustness to POMDP of algorithms using multi-step bootstrapping, e.g., MTD3 and MSAC, is more complicated and cannot be simply explained by the intuition that multi-step rewards can pass temporal information. Therefore, in the following section we will investigate if the \textbf{Unexpected Result } is related to the \textbf{Diff \ref{diff_exploration}}, i.e., exploration strategy.

\begin{figure*}[htp!]
    \centering
    \includegraphics[width=.7\textwidth]{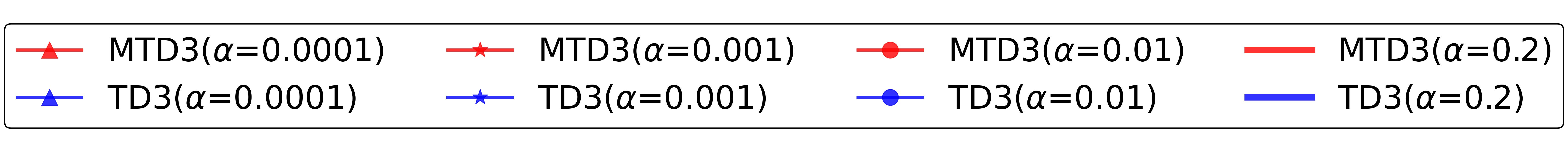}
    \includegraphics[width=.95\textwidth]{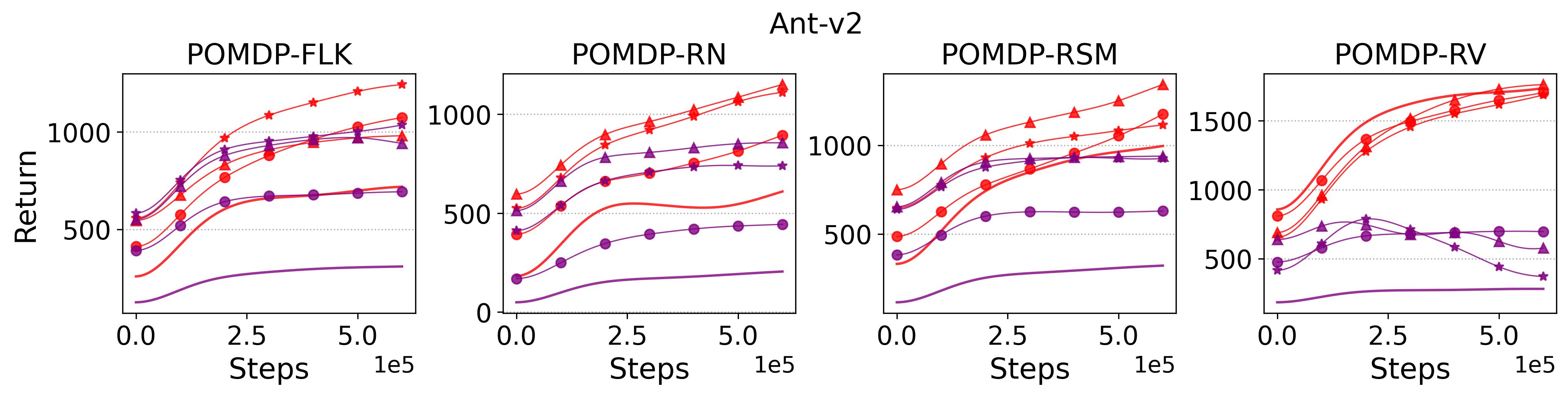}
    \includegraphics[width=.95\textwidth]{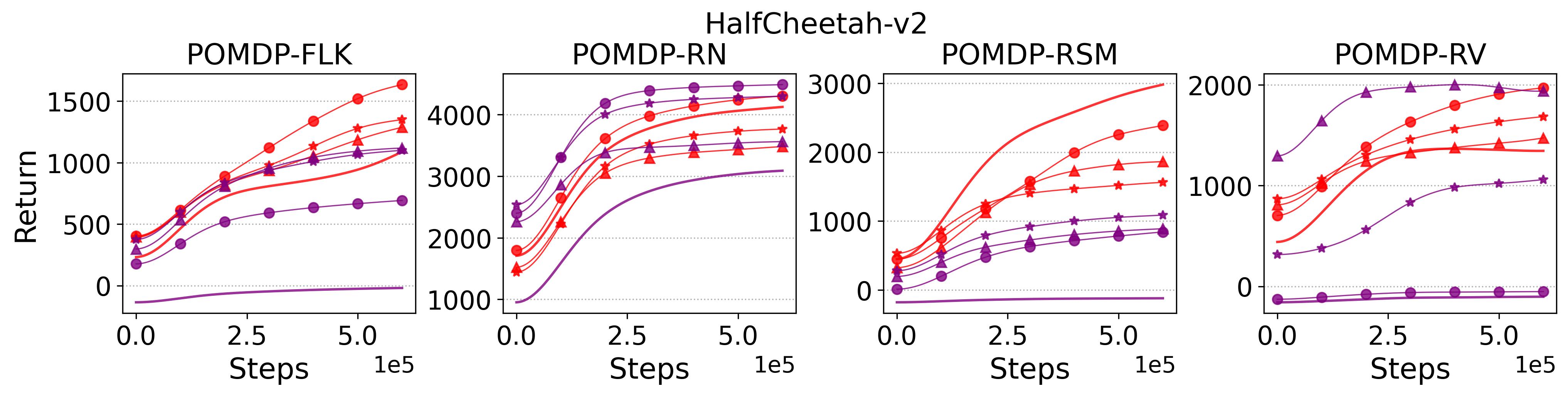}
    \caption{SAC vs MSAC(5) with Different Exploratory Strategies, where $\alpha$ is the entropy regularization coefficient. The bigger the $\alpha$ is the more exploratory the policy will be. The figure shows the average accumulated return over 4 random seeds.}
    \label{fig:SAC_and_MSAC_exploration_strategy_learning_curve}
\end{figure*}

\begin{figure*}[htp!]
    \centering
    \includegraphics[width=.7\textwidth]{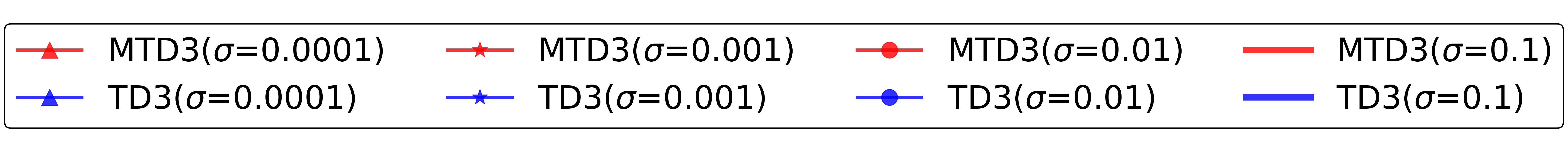}
    \includegraphics[width=.95\textwidth]{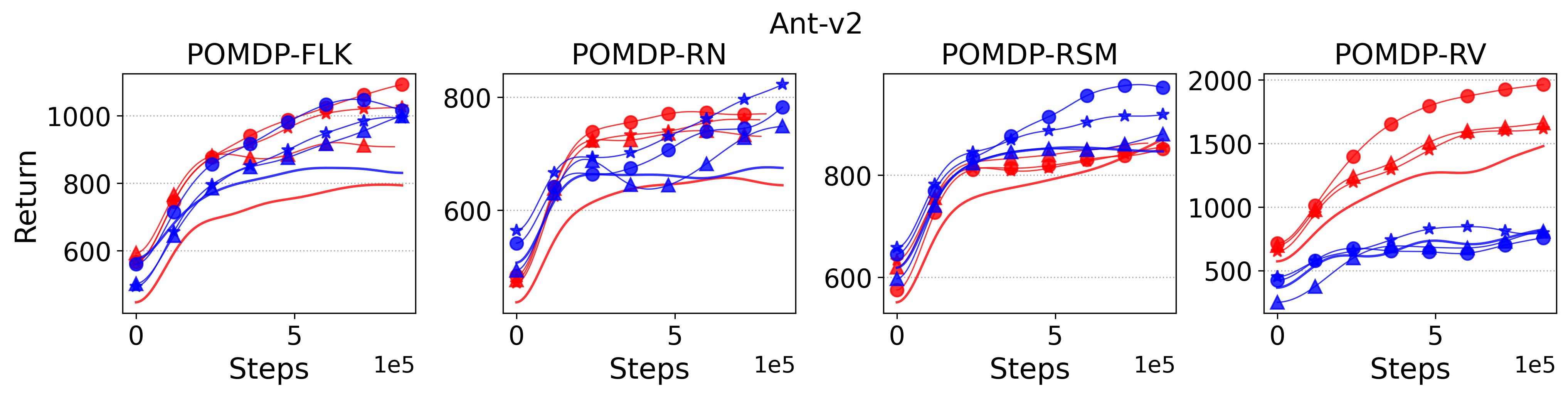}
    \includegraphics[width=.95\textwidth]{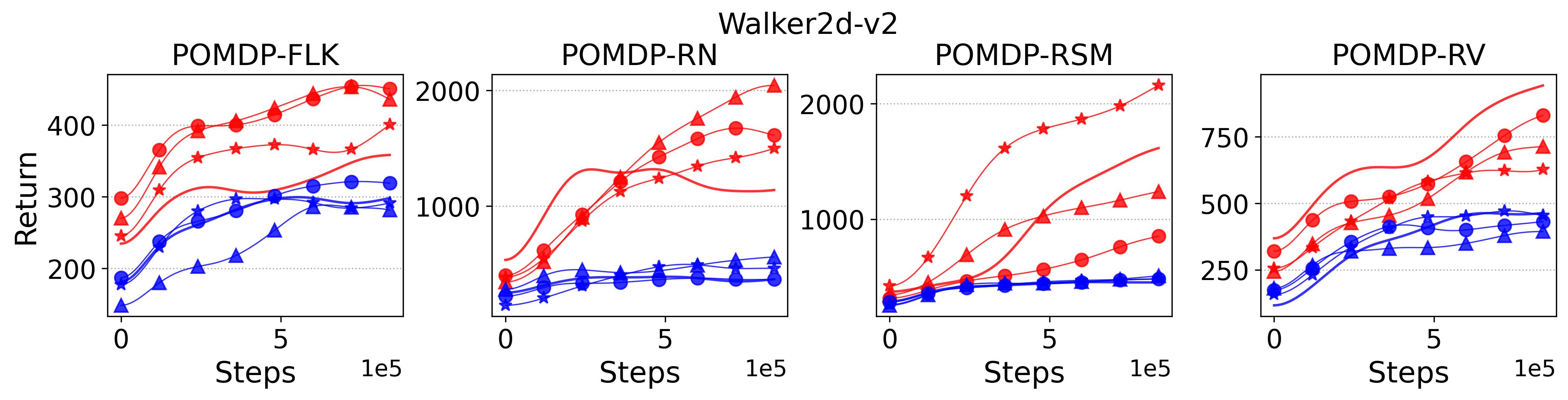}
    \caption{TD3 vs MTD3(5) with Different Exploratory Strategies, where $\sigma$ is the exploratory action noise. The bigger the $\sigma$ is the more exploratory the policy will be. The figure shows the average accumulated return over 4 random seeds. The dotted lines correspond to the performance of TD3 and MTD3(5) with $\sigma=0.1$.}
    \label{fig:TD3_and_MTD3_exploration_strategy_learning_curve}
\end{figure*}

\subsection{Results on Investigating the Effect of the Exploration Strategies}
\label{subsec:Results_on_Investigating_the_Effect_of_the_Exploration_Strategies}

In the (1) of \textbf{Hypothesis \ref{hypothesis_exploration_strategy}} we assume less exploratory strategy can make TD3 and SAC robuster to POMDP, so to validate this, we investigate SAC with $\alpha \in \{0.001, 0.01, 0.1, 0.2\}$ and TD3 with $\sigma \in \{0.0001, 0.001, 0.01, 0.1\}$. These hyperparameter ranges are chosen for principled reasons: (1) \textit{SAC entropy coefficient ($\alpha$)}: The range spans from very conservative exploration (0.001) to aggressive exploration (0.2), covering the automatic entropy tuning target range \cite{haarnoja2018soft} and enabling detection of exploration-robustness trade-offs; (2) \textit{TD3 action noise ($\sigma$)}: The logarithmic progression from 0.0001 to 0.1 captures the practical range where 0.1 represents standard exploration \cite{fujimoto2018addressing} while lower values test conservative exploration hypotheses; (3) \textit{Statistical power}: The 10-100× range variations provide sufficient signal-to-noise ratio to detect meaningful performance differences across exploration levels. For both $\alpha$ and $\sigma$, the larger the parameter is, the wilder the exploration will be. The commonly used default value for $\alpha$ and $\sigma$ are 0.2 and 0.1, respectively, as that for the results reported in Table \ref{tab:Maximum_of_Average_Return}. Fig. \ref{fig:SAC_and_MSAC_exploration_strategy_learning_curve} shows results of SAC and MSAC(5), while Fig. \ref{fig:TD3_and_MTD3_exploration_strategy_learning_curve} shows results of TD3 and MTD3(5)\footnote{For better visualization, we only show a subset of the investigated $\alpha$ and $\sigma$ values. More results can be found in \url{https://arxiv.org/abs/2209.04999}.}. Firstly, let us have a look at SAC and MSAC(5). As shown in Fig. \ref{fig:SAC_and_MSAC_exploration_strategy_learning_curve}, overall reducing the exploration can not make SAC consistently beat MSAC(5), even though there are exceptions, e.g., on HalfCheetah POMDP-RV SAC with $\alpha=0.005$ has better performance than MSAC(5) with $\alpha=0.2$. Nevertheless, for some cases, e.g., the POMDP variants of Ant-v2 and HalfCheetah-v2, reducing the exploration of SAC indeed allows it to achieve better performance, while for MSAC(5) reducing the exploration seems having less effect than that on SAC. Secondly, similar overall observations can be found for TD3 and MTD3(5) as shown in Fig. \ref{fig:TD3_and_MTD3_exploration_strategy_learning_curve} that for most tasks reducing TD3's exploration cannot achieve comparable performance of MTD3(5), especially for Walker2d-v2. However, we can see the different effect of reducing exploration for TD3 and SAC on Ant-v2, where SAC can get significant performance improvement from less exploratory policy while TD3 cannot. To summarize, it is not consistently observed that reducing the exploration in TD3 and SAC can make them robuster to POMDP, so the (1) of \textbf{Hypothesis \ref{hypothesis_exploration_strategy}} cannot be supported by our results. However, what is clear from these results is that for TD3 and SAC multi-step bootstrapping can help more than less exploratory strategy in terms of improving their performance on POMDPs.

\begin{figure*}[htp!]
    \centering
    \includegraphics[width=.75\textwidth]{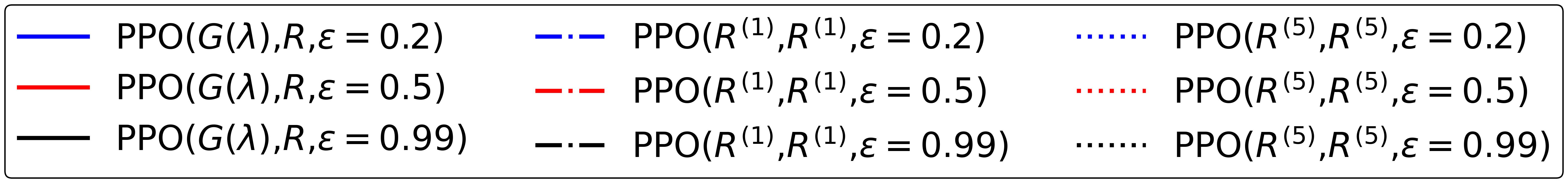}
    \includegraphics[width=.95\textwidth]{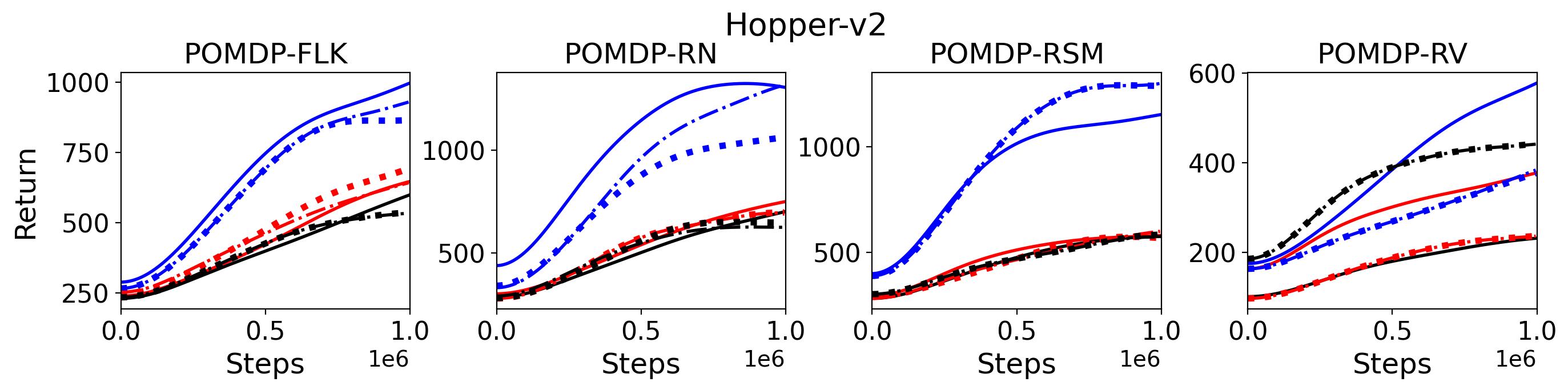}
    \includegraphics[width=.95\textwidth]{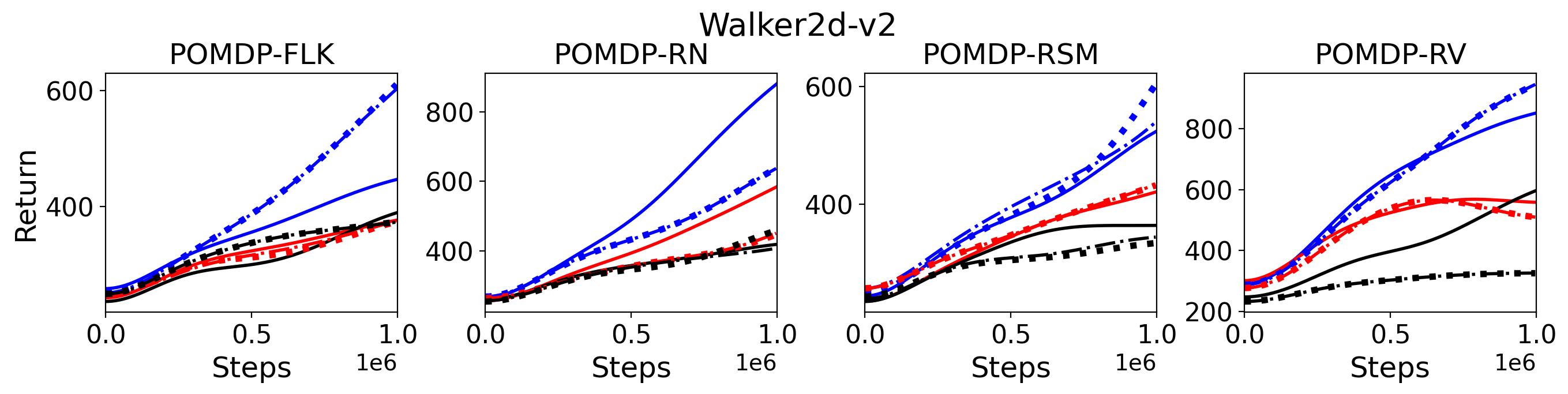}
    \caption{Effect of Exploration Strategy on The Performance of PPO, where PPO($G(\lambda)$, $R$, $\epsilon$) corresponds to the original PPO using $\lambda$-return $G(\lambda)$ for advantage estimate and Monte-Carlo return $R$ for state-value function update, and PPO($R^{(n)}$, $R^{(n)}$, $\epsilon$) corresponds to the revised PPO using $n$-step bootstrapped return for both advantage estimate and state-value function update. The investigated policy clip ratios are $\epsilon \in \{0.2,0.5,0.99\}$, where the larger the $\epsilon$ is, the wilder the exploration will be, as it allows the policy to update more far away from the current policy. For better visualization, only the average results over 3 random seeds are shown here.}
    \label{fig:Effect_of_Exploration_Strategy_on_Performance_of_PPO}
\end{figure*}

In the (2) of \textbf{Hypothesis \ref{hypothesis_exploration_strategy}} we assume wilder exploration strategy can make PPO vulnerable to POMDP, so to validate this, we investigate PPO with $\epsilon \in \{0.2, 0.5, 0.99\}$. The clip ratio range selection is theoretically motivated: (1) \textit{Conservative baseline ($\epsilon=0.2$)}: Standard PPO configuration \cite{schulman2017proximal} providing stable policy updates; (2) \textit{Moderate exploration ($\epsilon=0.5$)}: 2.5× increase allowing substantially larger policy deviations while maintaining reasonable bounds; (3) \textit{Aggressive exploration ($\epsilon=0.99$)}: Near-maximum theoretical value (1.0 would remove clipping entirely) testing extreme exploration scenarios; (4) \textit{Theoretical grounding}: The range captures the full spectrum from conservative to near-unlimited policy updates, enabling detection of exploration-induced instability in partial observability. $\epsilon=0.2$ is the commonly used default value for PPO. The results are presented in Fig. \ref{fig:Effect_of_Exploration_Strategy_on_Performance_of_PPO}, where we also show results on PPO($R^{(n)}$, $R^{(n)}$, $\epsilon$), i.e., the revised PPO using $n$-step bootstrapped return for both advantage estimate and state-value function update. In the figure, the blue-, red- and black-colored lines correspond to low, mediate, and large exploration, respectively. To save space, we only show results on Hopper-v2 and Walker2d-v2, but the results on Ant-v2 and HalfCheetah-v2 are very similar and can be found in \url{https://arxiv.org/abs/2209.04999}. Since Section \ref{subsec:Results_on_Understand_the_Effect_of_Multi-step_Bootstrapping} has discussed the effect of the multi-step bootstrapping of PPO in the Fig. \ref{fig:Effect_of_Multi_step_Size_on_Performance_of_PPO}, we will mainly consider the effect of exploration strategy here. It can be seen from the figure that for both the original PPO($G(\lambda)$, $R$, $\epsilon$) and the variants of PPO($R^{(n)}$, $R^{(n)}$, $\epsilon$), increasing the exploration leads to worse performance for most cases, which strongly supports the (2) of \textbf{Hypothesis \ref{hypothesis_exploration_strategy}}. 

As a summary for the validation of \textbf{Hypothesis \ref{hypothesis_exploration_strategy}}, the (1) of \textbf{Hypothesis \ref{hypothesis_exploration_strategy}} is not supported by our experiment results as there is no consistent trend can be found that reducing the exploration in TD3 and SAC makes them robuster to POMDPs, whereas the (2) of \textbf{Hypothesis \ref{hypothesis_exploration_strategy}} is strongly supported by our results that on most POMDPs increasing the exploration of PPO makes it vulnerable to POMDPs. These results reveal that the \textbf{Unexpected Result } cannot be explained by the \textbf{Diff \ref{diff_exploration}} too.

\section{Discussion}
\label{sec:Discussion}

\subsection{Summary of Hypothesis Validation Through Research Questions}

Our experimental results provide mixed validation of the four formal hypotheses, revealing both confirmations and rejections that offer valuable insights:

\begin{enumerate}
\item \textbf{Hypothesis \ref{hypothesis_generalization} (Confirmed)}: Performance inversion (MDP: TD3/SAC $>$ PPO; POMDP: PPO $>$ TD3/SAC) replicates systematically across diverse continuous control tasks (Ant, HalfCheetah, Hopper, Walker2D) and degradation mechanisms (RV, FLK, RSM, RN), validating the generalizability of this phenomenon.

\item \textbf{Hypothesis \ref{hypothesis_compare_n_step_TD3_and_SAC_to_vanilla} (Partially Confirmed)}: Multi-step bootstrapping significantly enhances robustness—MTD3 outperforms TD3 on 14/16 POMDP tasks, MSAC outperforms SAC on 15/16 POMDP tasks—strongly supporting part (1). However, part (2) is rejected: restricting PPO to single-step bootstrapping does not degrade POMDP performance, indicating PPO's robustness operates through different mechanisms than temporal integration.

\item \textbf{Hypothesis \ref{hypothesis_compre_accumulated_reward_task_to_one_step_reward_task} (Rejected)}: Accumulated reward modifications cannot consistently replicate multi-step bootstrapping benefits, demonstrating that the robustness mechanism is more sophisticated than simple reward smoothing and involves complex value function learning dynamics.

\item \textbf{Hypothesis \ref{hypothesis_exploration_strategy} (Partially Confirmed)}: Part (1) is rejected—reducing exploration in TD3/SAC does not consistently improve POMDP performance. Part (2) is confirmed—increasing PPO exploration (larger clip ratios) makes it vulnerable to POMDPs, supporting the role of conservative policy updates in PPO's robustness.
\end{enumerate}

These mixed results illuminate the mechanistic diversity underlying robustness to partial observability: TD3/SAC achieve improved POMDP performance through multi-step temporal credit assignment that enhances value function learning under state aliasing, while PPO's robustness derives from inherently conservative policy optimization that naturally limits destabilizing exploration in uncertain observational conditions. Critically, these mechanisms operate independently—PPO's robustness persists even when restricted to single-step returns, while TD3/SAC require explicit temporal extension to match PPO's stability.

\subsection{Computational Cost Analysis of Robustness Strategies}
We analyze computational overhead of key robustness strategies relative to baseline single-step algorithms, distinguishing between training and inference phases.

\textbf{Training Phase.} Multi-step bootstrapping (MTD3/MSAC) requires only one critic evaluation for the bootstrapped value at step $n$, while the discounted reward summation over $n$ immediate steps involves only lightweight arithmetic operations that can be pre-calculated and stored in the replay buffer. This results in negligible training overhead compared to single-step methods. History stacking increases input dimension by factor $H$, where $H$ is the history length, proportionally scaling network forward/backward pass costs. Recurrent methods (LSTM/GRU) introduce $O(d \cdot h + h^2)$ per-step costs \cite{hochreiter1997lstm,cho2014learning}, where $d$ is input dimension and $h$ is hidden size, and require sequential processing that limits parallelization, significantly increasing training time.

\textbf{Inference Phase.} Multi-step bootstrapping maintains identical inference costs to single-step methods—only the policy network forward pass is needed for action selection. History stacking requires maintaining and processing $H$ observation frames, increasing memory and computation linearly with history length. Recurrent methods must maintain hidden states and process sequential dependencies, adding $O(d \cdot h + h^2)$ operations per timestep \cite{hochreiter1997lstm,cho2014learning} with limited parallelization potential.

For early robustness assessment under suspected partial observability, multi-step bootstrapping provides the most efficient intervention with zero inference overhead, supporting the staged workflow prioritizing low-cost solutions before architectural complexity.

\subsection{Detecting and Addressing Partial Observability Through Performance Inversion}
Performance inversion—where PPO unexpectedly outperforms TD3/SAC contrary to MDP benchmarks—serves as a diagnostic signal for hidden partial observability. Missing latent variables (velocities, forces, delays) violate the Markov property, causing state aliasing that differentially impacts algorithm robustness. We propose a 4-stage workflow for systematic detection and mitigation (Fig. \ref{fig:practical_workflow_flowchart}):
\textbf{Stage 1: Detection.} Train PPO, TD3, SAC with standard settings. If PPO significantly outperforms TD3/SAC (performance inversion), flag partial observability.
\textbf{Stage 2: Multi-step Integration.} Apply MTD3/MSAC with $n=3,5,7$ steps. Select minimal $n$ achieving stable performance recovery with acceptable overhead.
\textbf{Stage 3: Observation Enhancement.} Test additional observations (velocities, forces, history) via ablation. If inversion disappears, finalize enhanced design.
\textbf{Stage 4: Advanced Methods.} Apply recurrence (LSTM/GRU) or model-based approaches only if stages 2-3 insufficient. Document final configuration and validation.
\textbf{Workflow:} Detection $\rightarrow$ Multi-step Integration $\rightarrow$ Observation Enhancement $\rightarrow$ Advanced Methods. This prioritizes low-cost interventions before complex architectures.

\begin{figure*}[htp!]
    \centering
    \begin{tikzpicture}[node distance=8mm and 12mm, scale=0.85, transform shape]
        \node[stage] (s1) {1 Detection};
        \node[decision, below=of s1] (d1) {Inversion?};
        \node[stage, below=of d1] (s2) {2 Multi-step\\Integration};
        \node[stage, below=of s2] (s3) {3 Observation\\Enhancement};
        \node[decision, below=of s3] (d2) {Sufficient?};
        \node[stage, below=of d2] (s4) {4 Advanced\\Methods};

        \node[stage, right=28mm of d1] (normal) {No inversion \\ Standard tuning};
        
        \node[stage, right=28mm of d2] (complete) {Complete \\ Solution found};
        
        \node[small, right=20mm of s2] (noteMS) {\begin{tabular}{c}MTD3/MSAC\\n=3,5,7\end{tabular}};
        \node[small, right=20mm of s3] (noteObs) {\begin{tabular}{c}Add velocities\\forces, history\end{tabular}};
        \node[small, right=20mm of s4] (noteAdv) {\begin{tabular}{c}LSTM/GRU\\if needed\end{tabular}};

        \draw[link] (s1) -- (d1);
        \draw[link] (d1) -- node[right, small]{Yes} (s2);
        \draw[link] (s2) -- (s3);
        \draw[link] (s3) -- (d2);
        \draw[link] (d2) -- node[right, small]{No} (s4);
        
        \draw[link] (d1) -- node[above, small]{No} (normal);
        \draw[link] (d2) -- node[above, small]{Yes} (complete);
        
        \draw[dashed, ->, thin] (noteMS) -- (s2);
        \draw[dashed, ->, thin] (noteObs) -- (s3);
        \draw[dashed, ->, thin] (noteAdv) -- (s4);
    \end{tikzpicture}
    \caption{Streamlined 4-stage workflow: (1) detect performance inversion (PPO unexpectedly \textgreater TD3/SAC), (2) apply multi-step integration (MTD3/MSAC), (3) enhance observation space, (4) escalate to advanced methods (LSTM/GRU). Prioritizes low-cost interventions before complex architectures.}
    \label{fig:practical_workflow_flowchart}
\end{figure*}

\begin{table}[htp!]
    \centering
    \caption{Summary of Hypotheses Validation}
    \label{tab:Summary_of_Hypotheses_Validation}
    \small
    \begin{tabular}{c|l|l|c}
        \toprule\hline
        \multirow{6}{*}{\rotatebox[origin=c]{90}{Hypothesis}} & \multicolumn{2}{l|}{\textbf{H \ref{hypothesis_generalization}} (Generalization)}    &  \cmark \\\cline{2-4}              
                    & \multirow{2}{*}{\textbf{H \ref{hypothesis_compare_n_step_TD3_and_SAC_to_vanilla}} (Multi-step Bootstrapping)}  & (1) TD3\&SAC $\Uparrow$   &  \cmark  \\\cline{3-4}
                    &                                                                    & (2) PPO  $\Downarrow$     &  \xmark  \\\cline{2-4}
                    & \multicolumn{2}{l|}{\textbf{H \ref{hypothesis_compre_accumulated_reward_task_to_one_step_reward_task}} (Accumulated-reward)} & \xmark \\\cline{2-4}
                    & \multirow{2}{*}{\textbf{H \ref{hypothesis_exploration_strategy}} (Exploration)}                   & (1) TD3\&SAC $\Downarrow$ & \xmark \\\cline{3-4}
                    &                                                                    & (2) PPO  $\Uparrow$       & \cmark \\\bottomrule
        \multicolumn{4}{p{6cm}}{\footnotesize{\cmark: support, \xmark: reject}}
    \end{tabular}
\end{table}

\subsection{Implications for Robustness Mechanisms}
Table \ref{tab:Summary_of_Hypotheses_Validation} reveals distinct robustness pathways: \textbf{TD3/SAC} achieve POMDP robustness primarily through multi-step bootstrapping (improving value function learning), while \textbf{PPO} relies on conservative policy updates rather than temporal integration. The rejection of accumulated reward and exploration hypotheses suggests that multi-step benefits arise from enhanced value estimation rather than temporal information propagation.

\subsection{Conceptual Summary}
Our contribution combines empirical validation with initial theoretical development: we provide an observability-gap framework with variance bounds and bias-variance horizon trade-offs that explain why multi-step bootstrapping and conservative exploration attenuate performance degradation under partial observability. Beyond theoretical insights, we isolate two practical levers available to practitioners—(i) temporal aggregation via multi-step returns and (ii) exploration conservatism—and demonstrate how they shift robustness profiles under observation insufficiency. The findings support a streamlined diagnostic workflow that prioritizes low-cost interventions: detect performance inversion, apply multi-step bootstrapping with exploration tuning, enhance observation space systematically, and escalate to complex architectures only when necessary. These interventions require no latent-state identification or structural environment changes, making them immediately applicable in early development phases. This theoretical foundation provides both immediate practical guidance and groundwork for future analyses explaining the divergent robustness mechanisms of conservative policy updates versus multi-step value estimation.

\subsection{Limitations and Future Directions}
While TriMap dimensionality reduction provides insights into policy behavior differences, visualization artifacts may affect interpretability. Additional robustness factors beyond multi-step bootstrapping and exploration strategies warrant investigation, though our systematic approach establishes a foundation for understanding algorithm-specific POMDP resilience mechanisms.


\section{Conclusion and Future Works}
\label{sec:conclusion}

This paper investigates the counter-intuitive \textit{performance inversion} phenomenon where PPO outperforms TD3 and SAC in partially observable environments, contrary to their typical rankings in fully observable MDPs. Through systematic experimental validation across diverse benchmarks, we confirm this inversion generalizes broadly across continuous control tasks with various partial observability mechanisms.

We provide both theoretical and empirical insights into the underlying robustness mechanisms. Our observability-gap framework with variance bounds and bias-variance trade-offs explains why multi-step bootstrapping and conservative exploration strategies mitigate performance degradation under partial observability. Experimentally, we demonstrate that multi-step extensions (MTD3, MSAC) significantly improve robustness, with MTD3(5) outperforming TD3 on 14/16 POMDP tasks and MSAC(5) outperforming SAC on 15/16 tasks.

Key findings reveal distinct robustness pathways: TD3/SAC benefit primarily from multi-step temporal aggregation improving value function learning, while PPO's robustness stems from conservative policy updates rather than temporal integration. These insights inform our proposed diagnostic workflow for systematic detection and mitigation of partial observability challenges.

\textbf{Future Research Directions:} Several promising avenues emerge from this work: (1) extending theoretical analysis to provide deeper mechanistic understanding of multi-step bootstrapping's variance-reduction effects; (2) generalizing findings to discrete control domains (MiniGrid\cite{MinigridMiniworld23}, Procgen\cite{cobbe2019procgen}) and high-dimensional observations (vision, point clouds); (3) developing the performance inversion diagnostic into quantitative partial observability detection tools for observation design; (4) integrating multi-step methods with modern sequence architectures for enhanced temporal modeling; (5) establishing standardized POMDP benchmarks for systematic algorithm evaluation and automatic MDP/POMDP classification.

\section*{Acknowledgments}
This work is supported by a SSHRC Partnership Grant in collaboration with Philip Beesley Studio, Inc. and enabled in part by support provided by the Digital Research Alliance of Canada (www.alliancecan.ca). Dana Kuli{\'c} is supported by the ARC Future Fellowship (FT200100761). We would like to thank all reviewers for taking the time and effort to review the manuscript. We sincerely appreciate all valuable comments and suggestions, which helped us to improve the quality of the manuscript.

\section*{Declaration of Generative AI and AI-assisted Technologies in the Manuscript Preparation Process}
During the preparation of this work the authors used GPT-5 and Claude Sonnet 4 in order to proofread the paper, highlight key findings in the paper, improve theoretical perspective, inspire discussion based on the experiment results, etc. After using these tools, the authors reviewed and edited the content as needed and take full responsibility for the content of the published article.

\bibliographystyle{elsarticle-num} 
\bibliography{sample}

\end{document}

\endinput